\documentclass[lettersize,journal]{IEEEtran}
\usepackage{amsmath,amsfonts}
\usepackage{array}
\usepackage{textcomp}
\usepackage{stfloats}
\usepackage{url}
\usepackage{verbatim}
\usepackage{graphicx}
\usepackage{cite}
\usepackage{bm}
\usepackage{array}
\usepackage{amssymb}
\makeatletter
\let\NAT@parse\undefined
\makeatother
\usepackage{hyperref}  
\hypersetup{hypertex=true,
            colorlinks=true,
            linkcolor=blue,
            anchorcolor=blue,
            citecolor=blue
            }
\usepackage{float}
\usepackage{subfigure}
\usepackage{color}
\usepackage{multirow}

\usepackage {amssymb}
\newtheorem{hypothesis}{Hypothesis}
\newtheorem{definition}{Definition}

\usepackage{algpseudocode}
\usepackage{footmisc}
\usepackage{threeparttable}
\usepackage{adjustbox}
\usepackage{booktabs}
\usepackage{utfsym}
\usepackage{makecell}
\definecolor{DarkRed}{rgb}{0.55, 0.0, 0.0}
\definecolor{DarkGreen}{rgb}{0.0, 0.5, 0.0}
\usepackage[ruled,vlined,linesnumbered]{algorithm2e}
\algnewcommand\algorithmicforeach{\textbf{for each}}
\algdef{S}[FOR]{ForEach}[1]{\algorithmicforeach\ #1\ \algorithmicdo}
\begin{document}
\title{\LARGE \bf
UniLGL: Learning Uniform Place Recognition for FOV-limited/Panoramic LiDAR Global Localization}
\author{
Hongming Shen$^{1,\dag}$,~\IEEEmembership{Member,~IEEE}, Xun Chen$^{1,\dag}$, Yulin Hui$^2$, Zhenyu Wu$^{1,3*}$, \\Wei Wang$^1$, Qiyang Lyu$^1$, Tianchen Deng$^4$, and Danwei Wang$^1$,~\IEEEmembership{Life Fellow,~IEEE}
\thanks{This research is supported by the National Research Foundation (NRF),
Singapore, under the NRF Medium Sized Centre scheme (CARTIN).
Any opinions, findings and conclusions or recommendations expressed in
this material are those of the author(s) and do not reflect the views of
National Research Foundation, Singapore.}
\thanks{$^1$ Authors are with the Centre for Advanced Robotics Technology Innovation, Nanyang Technological University, Singapore 639798.}
\thanks{$^2$ Author is with the School of Electrical and Information Engineering, Tianjin University, Tianjin, China 300072.}
\thanks{$^3$ Author is with the School of Automation, Hangzhou Dianzi University, Hangzhou, China 310018.}
\thanks{$^4$ Author is with the Department of Automation, Shanghai Jiao Tong University, Shanghai, China 200240.}
\thanks{$^\dag$ These authors contributed equally to this work. $^*$ Corresponding author: ZHENYU002@e.ntu.edu.sg.
}%
}
\maketitle
\begin{abstract}
LiDAR-based Global Localization (LGL) is an essential ingredient for autonomous robots.
However, existing LGL methods typically consider only partial information (e.g., geometric features) from LiDAR observations or are designed for homogeneous LiDAR sensors, overlooking the uniformity in LGL.
In this work, a uniform LGL method is proposed, termed UniLGL, which simultaneously achieves spatial and material uniformity, as well as sensor-type uniformity.
The key idea of the proposed method is to encode the complete point cloud, which contains both geometric and material information, into a pair of Bird’s Eye View (BEV) images (i.e., a spatial BEV image and an intensity BEV image), thereby transforming the LGL problem into a cascaded LiDAR Place Recognition (LPR) and pose estimation problem from the perspective of image fusion.
An end-to-end multi-BEV fusion network is designed to extract uniform features, equipping UniLGL with spatial and material uniformity.
To ensure robust LGL across heterogeneous LiDAR sensors, a viewpoint invariance hypothesis is introduced, which replaces the conventional translation equivariance assumption commonly used in existing LPR networks and supervises UniLGL to achieve sensor-type uniformity in both global descriptors and local feature representations.
Moreover, UniLGL introduces a pipeline that leverages a pre-trained single-image Vision Foundation Model (VFM) for feature extraction to enhance the multi-BEV fusion LPR network, enabling strong generalization with only a few LiDAR data for fine-tuning.
Finally, based on the mapping between local features on the 2D BEV image and the point cloud, a robust global pose estimator is derived that determines the global minimum of the global pose on $\text{SE}(3)$ without requiring additional registration.
To validate the effectiveness of the proposed uniform LGL, extensive benchmarks are conducted in real-world environments, and the results show that the proposed UniLGL is demonstratively competitive compared to other State-of-the-Art (SOTA) LGL methods.
Furthermore, UniLGL has been deployed on diverse platforms, including full-size trucks and agile Micro Aerial Vehicles (MAVs), to enable high-precision localization and mapping as well as multi-MAV collaborative exploration in port and forest environments, demonstrating the applicability of UniLGL in industrial and field scenarios.
The code will be released at \url{https://github.com/shenhm516/UniLGL}.
\end{abstract}
\section{Introduction}
\begin{figure}[!t]\centering
\subfigure[Spatial and Material Uniformity.]{
\includegraphics[width=0.9\linewidth]{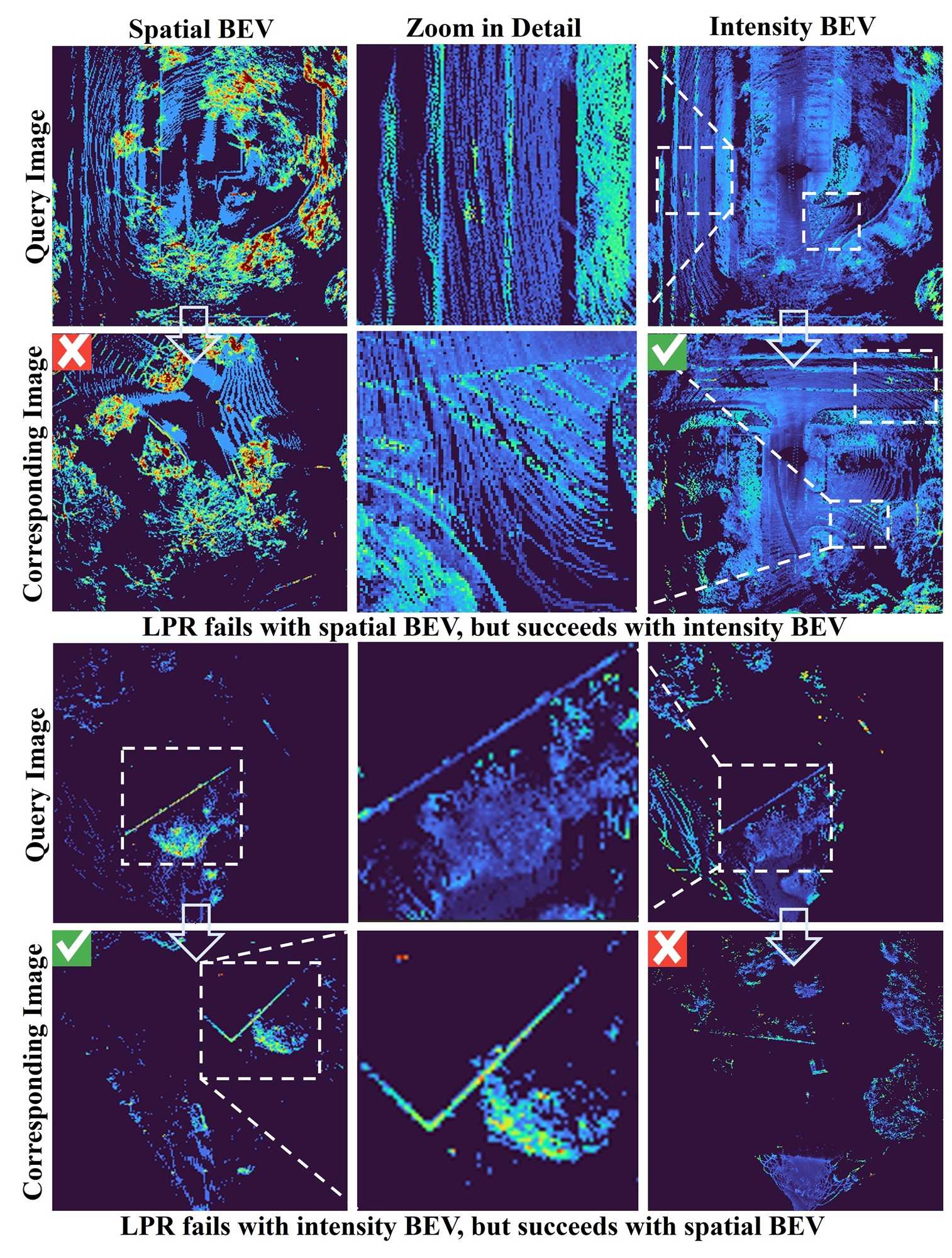}
\label{Fig: MotivationOfIntroduceIntensity}
}
\subfigure[Sensor-type Uniformity.]{
    \includegraphics[width=0.9\linewidth]{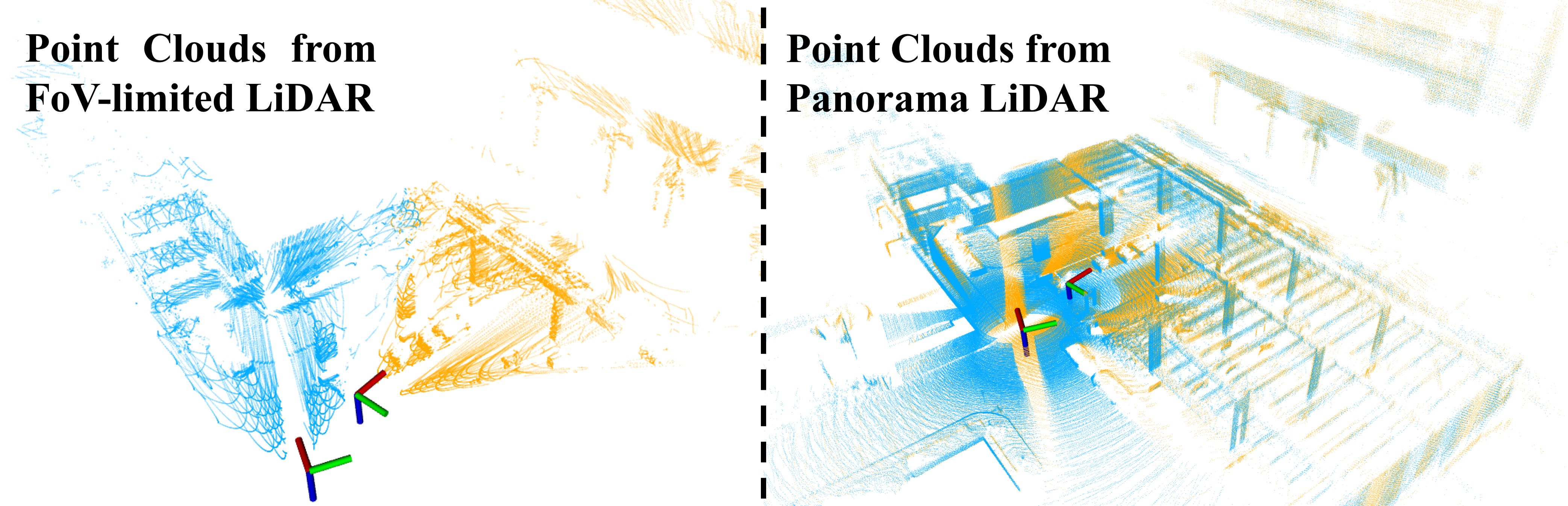} 
    \label{Fig: CorrespondingPointCloudUnderTranslationEquivariance}
}
\caption{Demonstration of the Uniformity. (a) Spatial and Material Uniformity: If only the spatial BEV of a LiDAR point cloud is used for LPR, material properties of environmental structures, such as the highly reflective painted marker, will be lost. Conversely, if intensity information is introduced to replace the height channel in the spatial BEV, as shown in the intensity BEV of the LiDAR point cloud, height-related details (such as trees and tall buildings) will be difficult to distinguish.
(b) Sensor-type Uniformity: For panoramic LiDAR, structures observed at nearby locations remain consistent regardless of the rotation. In contrast, for FoV-limited LiDAR, structures scanned at close locations can differ significantly under different rotations.}
\label{Fig: title_fig}
\vspace{-1em}
\end{figure}
\IEEEPARstart{G}{lobal} localization is a fundamental task in developing autonomous robotic systems.
During the last decade, global localization has been achieved using the Global Navigation Satellite System (GNSS)\cite{9562037}, infrastructures (e.g., Ultra-Wideband\cite{9502143}, RFIDs\cite{9509296}, and QR codes\cite{8602360}), and visual/LiDAR place recognition\cite{NETVLAD-PAMI,ScanContext++,PointNetVLAD,BEVPlace++}.
However, GNSS cannot work in an indoor or dense urban area due to the multi-path effect, and infrastructure-based global localization
\cite{9509296,8602360,9502143}
relies on installation and calibration.
Besides, visual place recognition can degrade significantly in conditions of appearance changes, i.e., the appearances of particular areas may change drastically under different illumination conditions.
Owning to the direct depth measurement of LiDAR, which is immune to scene illumination changes, in recent years, LPR has been widely adopted to enable infrastructure-free global localization in GNSS-denied environments.
However, LPR still faces several challenges, including the lack of uniformity, the absence of a foundation model, and the gap between topological and metric localization.

\textbf{Lack of Uniformity}: In this work, the term \textit{Uniform} manifests in two key aspects: Spatial and Material Uniformity, and Sensor-Type Uniformity.
\textit{Spatial and Material Uniformity}: A LPR algorithm with spatial and material uniformity should leverage both the spatial structure and the material information (i.e., intensity) captured by LiDAR for reliable place recognition.
However, most existing LPR methods either neglect the material cues encoded in the intensity channel\cite{BEVPlace++,lu2025ringSharp,ScanContext,ScanContext++,STD,yuan2024btc} or exploit only partial spatial information\cite{isc,IDBoW}, thereby considerably limiting their performance.
As a concrete example shown in Fig. \ref{Fig: MotivationOfIntroduceIntensity}, pruning material information leads to LPR failure with the spatial BEV due to missing reflective ground markings, while pruning spatial information causes failure with the intensity BEV as height cues are lost (e.g., trees and grass being difficult to distinguish).
\textit{Sensor-Type Uniformity}: A LPR method with sensor-type uniformity should be appropriate for both FoV-limited and panoramic LiDAR.
Most existing learning-based LPR methods \cite{BEVPlace++,lu2025ringSharp,lcdnet,LoGG3D-Net,jung2025imlpr} follow the paradigm of position-based place recognition \cite{SLAM2}, which determines whether a robot revisits the same geographical region solely based on its position, while ignoring the effects of viewpoint changes.
However, as illustrated in Fig. \ref{Fig: CorrespondingPointCloudUnderTranslationEquivariance}, for FoV-limited LiDARs, the observed point clouds at the same position but with different orientations can be almost disjoint, making position-based place recognition unsuitable for FoV-limited LiDARs.

\textbf{Absence of LiDAR Foundation Model}: The superior performance of most modern learning-based place recognition methods has been largely attributed to large-scale training, as corroborated in \cite{delf,delg,NETVLAD-PAMI,CosPlace}.
Over the past decade, the scale of training data has expanded dramatically, increasing from hundreds of thousands in early works (e.g., the Pittsburgh dataset \cite{Pittsburgh} powered NetVLAD \cite{NETVLAD-PAMI}) to millions (e.g., Google-Landmark datasets\cite{Google-Landmark} powered DeLF\cite{delf} and DeLG\cite{delg}) or even tens of millions (e.g., San Francisco XL
dataset powered CosPlace\cite{CosPlace}) in recent years.
However, unlike visual images that can be easily acquired at scale through map services (e.g., Google Street View), collecting large-scale point cloud data remains challenging and resource-intensive.
Consequently, most existing LPR methods either rely on classical hand-crafted features\cite{ScanContext++,yuan2024btc,ScanContext,ring++} or lightweight models\cite{BEVPlace++,lu2025ringSharp,lcdnet,LCRNet,LoGG3D-Net} trained on relatively small-scale datasets, which greatly limits the generalization ability of LPR.

\textbf{Gap between Topological and Metric Localization}: A complete LGL pipeline should encompass not only the LPR component discussed above but also the global pose estimation module.
Nonetheless, many existing approaches concentrate solely on realizing topological localization through LPR\cite{PointNetVLAD,jung2025imlpr,mwjung-2025-icra,ScanContext,isc,LoGG3D-Net}, assuming that metric localization can be achieved using point cloud registration algorithms\cite{icp, GICP, NDT}.
However, registration is prone to falling into local minima when there is a large initial pose discrepancy between point clouds, e.g., reverse-direction loop closures.
Developing a complete LGL system that performs both place recognition and global pose estimation, thereby effectively bridging the gap between topological and metric localization without requiring any initial guess information, remains a challenging problem.

\subsection{Related Works}
In this section, a literature review is presented according to the aforementioned challenges in LGL.
\subsubsection{LPR with Uniformity}
Over the past decade, several studies have achieved LPR through different paradigms, including local feature matching (e.g., SHOT\cite{tombari2010unique}, FPFH\cite{rusu2009fast}), hand-crafted global descriptors (e.g., the well-known ScanContext series\cite{ScanContext, ScanContext++}), and end-to-end neural networks (e.g., PointNetVLAD\cite{PointNetVLAD}).
Despite the extensive research on LPR, the concept of uniformity has rarely been considered.
To achieve spatial and material uniformity, a series of methods integrates the intensity information into well-established place recognition descriptors.
In \cite{ISHOT}, a LPR descriptor with spatial and material uniformity is proposed, which encodes intensity information into the original SHOT descriptor\cite{tombari2010unique} constructed from 3D geometric information.
Intensity Scan Context \cite{isc} replaces the height channel used in the original Scan Context \cite{ScanContext} with the intensity of LiDAR points, which justifies that intensity information can be distinctive for places recognition.
In addition to the aforementioned handcrafted descriptor-based LPR methods, some approaches, such as LoGG3D-Net\cite{LoGG3D-Net} and LCDNet\cite{lcdnet}, directly take 4D point clouds (containing both geometric and intensity information) as network input, attempting to learn spatial and material uniformity.
Considering the sparsity of point clouds, methods like OverlapNet\cite{chen2021overlapnet} and ImLPR\cite{jung2025imlpr} represent the 4D point cloud using multi-channel range images, including range channel, intensity channel, etc., and learn the similarity between a pair of images to achieve spatial and material uniform LPR.

Compared with spatial and material uniformity, sensor-type uniformity has received little attention in LPR research, especially for learning-based approaches.
As corroborated by Solid \cite{Solid}, designing an LPR system with sensor-type uniformity can significantly improve performance, particularly for FoV-limited LiDARs.
However, most existing learning-based LPR methods\cite{BEVPlace++,lu2025ringSharp,lcdnet,LoGG3D-Net,PointNetVLAD, jung2025imlpr} assume that LiDAR observations are rotation-invariant, which naturally holds for panoramic LiDARs but not for FoV-limited LiDARs, as illustrated in Fig.~\ref{Fig: CorrespondingPointCloudUnderTranslationEquivariance}.
To address this problem, OverlapNet \cite{chen2021overlapnet} represents point clouds in Range Image View (RIV) and employs a Siamese network to estimate the overlap between two range images, providing a more natural metric than geographical distance for point cloud similarity.
However, the RIV representation limits OverlapNet's generalization to FoV-limited LiDARs.
Specifically, to ensure that range images do not contain large regions of vacant pixels, different horizontal resolutions have to be used for panoramic and FoV-limited LiDARs, making it difficult to generalize a single network across heterogeneous LiDAR configurations.
To achieve sensor-type uniformity, HeLioS \cite{mwjung-2025-icra} replaces the Siamese network used in OverlapNet\cite{chen2021overlapnet} with a sparse U-Net\cite{U-net}, enabling it to process sparse point cloud data instead of relying on RIV images.
The extracted local features are subsequently aggregated into global LPR descriptors through GeM pooling and the SALAD attention module \cite{izquierdo2024optimal}.
\subsubsection{Foundation Model Enhanced Place Recognition}
With the emergence of self-supervised learning in recent years, which promises to eliminate the need for manual data annotation, enabling models to scale effortlessly to massive datasets and larger architectures (referred to as foundation models).
Following the success of Vision Transformers (ViTs)\cite{ViT} in demonstrating the scalability of the Transformer for computer vision, a broad range of modern downstream applications have consumed pre-trained foundation models, including 3D understanding\cite{wang2025vggt}, vision language action\cite{kim2024openvla}, and segmentation\cite{ravi2024sam}.
For place recognition, several studies (e.g., AnyLoc\cite{keetha2023anyloc}, SALAD\cite{izquierdo2024optimal}) have demonstrated that VFMs can encode rich visual representations that provide a strong substrate for building visual place recognition solutions, even without fine-tuning or with only a few epochs of adaptation.
The remarkable success of vision VFMs has inspired several efforts to develop 3D point cloud foundation models, such as Sonata\cite{wu2025sonata} and PTv3\cite{PTv3}.
However, due to the difficulty of collecting large-scale and diverse 3D point clouds, existing LiDAR foundation models still lack the expressiveness required to generalize across diverse outdoor environments, and thus remain limited to small-scale downstream tasks, diverging from LPR.
Instead of developing LiDAR foundation models, several recent studies have attempted to leverage VFMs for 3D point cloud processing tasks.
VFM-Registration\cite{vodisch2025lidar} extracts visual features from RGB images using VFMs and projects them into 3D space using the extrinsic between the camera and LiDAR, achieving high-performance point cloud registration.
ImLPR\cite{jung2025imlpr} represents LiDAR point clouds as RIV images and employs the vision transformer foundation model DINOv2\cite{oquab2023dinov2} to extract discriminative features from these RIV representations, enabling highly generalizable LPR.

\subsubsection{Complete LGL with Metric Localization}
Unlike the aforementioned pure LPR methods, which mainly focus on learning discriminative global descriptors, a complete LGL system should not only extract global descriptors of point clouds but also capture their local features.
Specifically, the global descriptors, aggregated from local features, are more robust to local noise and are typically used for LPR, whereas the local features are employed for corresponding point cloud alignment (global pose estimation).
STD \cite{STD} voxelizes the point cloud to extract keypoints as local features and constructs a triangle descriptor based on voxel distribution for place recognition.
The relative pose between matched point clouds is estimated by aligning the triangle vertices using the \textit{Umeyama} alignment \cite{icp}.
BTC \cite{yuan2024btc} improves global localization performance by introducing a binary descriptor into the STD\cite{STD}, providing a more detailed and discriminative representation of local point cloud geometry.
RING++ \cite{ring++} applies the Radon transform to the BEV image of point clouds to extract local features.
A translation-invariant global descriptor is then constructed by applying the discrete Fourier transform to these local features.
The handcrafted descriptors used in the aforementioned methods, though elegantly designed, often suffer from low information density.
To mitigate the risk of numerous false matches, handcrafted LPR is typically coupled with a geometric verification step.
For example, STD \cite{STD} and BTC\cite{yuan2024btc} typically identify multiple-loop candidate point clouds for a query scan through descriptor voting and then select the one with the best geometric verification. Similarly, RING++ \cite{ring++} directly couples rotation estimation with the LPR process by performing an exhaustive search over the orientation space.

To address this challenge, a series of methods have been proposed to learn highly discriminative descriptors.
LCDNet \cite{lcdnet} presents an end-to-end LGL framework, which employs PV-RCNN \cite{PV-RCNN} for robust local feature extraction and NetVLAD \cite{NETVLAD-PAMI} for global descriptor aggregation.
LCRNet \cite{LCRNet} introduces a novel feature extraction backbone and a pose-aware attention mechanism to jointly estimate place similarity and 6-DoF relative pose between pairs of LiDAR scans.
SpectralGV \cite{vidanapathirana2023sgv} extends Logg3D-Net\cite{LoGG3D-Net} to LGL by leveraging its local features to register point cloud pairs that are initially matched using the global descriptors produced by Logg3D-Net\cite{LoGG3D-Net}.
BEV-Place++\cite{BEVPlace++} represents point clouds using BEV images and adopts a rotation equivariant and invariant network to extract local features from the BEV images. The local features are then aggregated into a global descriptor using NetVLAD\cite{NETVLAD-PAMI}.
BEV-Place++\cite{BEVPlace++} demonstrates that representing point clouds with BEV images yields superior generalization capability compared to raw point clouds, particularly under viewpoint variations and scene changes.
Similarly, RING\# \cite{lu2025ringSharp} is proposed as an end-to-end framework that jointly learns LPR and 3-DoF global localization, leveraging the frequency-domain BEV representation introduced in RING++ \cite{ring++}.
\begin{table*}[!t] \centering
    \setlength{\tabcolsep}{0.8pt} 
    \centering
\caption{Overview of State-of-the-art LPR/LGL Approaches.}
\label{tab: related_work}
\begin{threeparttable}
    \begin{tabular}{l|| c c c c c c c c c c c c c}
    \hline\hline
    Capability &\makecell{ISC\\\cite{isc}} &\makecell{Solid\\\cite{Solid}} &\makecell{RING++\\\cite{ring++}} &\makecell{HeLioS\\\cite{mwjung-2025-icra}}&\makecell{ImLPR\\\cite{jung2025imlpr}}&\makecell{OverlapNet\\\cite{chen2021overlapnet}}&\makecell{BEVPlace++\\\cite{BEVPlace++}}&\makecell{RING\#\\\cite{lu2025ringSharp}}&\makecell{LCDNet\\\cite{lcdnet}}&\makecell{Logg3D-Net\\\cite{LoGG3D-Net,vidanapathirana2023sgv}}&\makecell{UniLGL\\(Proposed)}\\\hline
    Place Recognition &Handcrafted &Handcrafted &Handcrafted &Learning &Learning &Learning &Learning &Learning &Learning &Learning &Learning\\
    Global Localization &{\color{DarkRed}\usym{2717}} &1-DoF &3-DoF &{\color{DarkRed}\usym{2717}} &{\color{DarkRed}\usym{2717}} &1-DoF &3-DoF &3-DoF &6-DoF &6-DoF &6-DoF\\
    Spatial and Material Uniformity &{\color{DarkGreen}\usym{1F5F8}} &{\color{DarkRed}\usym{2717}} &{\color{DarkRed}\usym{2717}} &{\color{DarkRed}\usym{2717}} &{\color{DarkGreen}\usym{1F5F8}} &{\color{DarkGreen}\usym{1F5F8}} &{\color{DarkRed}\usym{2717}} &{\color{DarkRed}\usym{2717}} &{\color{DarkGreen}\usym{1F5F8}} &{\color{DarkGreen}\usym{1F5F8}} &{\color{DarkGreen}\usym{1F5F8}}\\
    Sensor-type Uniformity &{\color{DarkRed}\usym{2717}} &{\color{DarkGreen}\usym{1F5F8}} &{\color{DarkGreen}\usym{1F5F8}} &{\color{DarkGreen}\usym{1F5F8}} &{\color{DarkRed}\usym{2717}} &{\color{DarkRed}\usym{2717}}  &{\color{DarkRed}\usym{2717}} &{\color{DarkRed}\usym{2717}} &{\color{DarkRed}\usym{2717}} &{\color{DarkRed}\usym{2717}} &{\color{DarkGreen}\usym{1F5F8}}\\
    Foundation Model Enhanced &{\color{DarkRed}\usym{2717}} &{\color{DarkRed}\usym{2717}} &{\color{DarkRed}\usym{2717}} &{\color{DarkRed}\usym{2717}} &{\color{DarkGreen}\usym{1F5F8}} &{\color{DarkRed}\usym{2717}} &{\color{DarkRed}\usym{2717}} &{\color{DarkRed}\usym{2717}} &{\color{DarkRed}\usym{2717}} &{\color{DarkRed}\usym{2717}} &{\color{DarkGreen}\usym{1F5F8}}\\
    Point Cloud Representation &Point &Point &BEV &Point &RIV &RIV &BEV &BEV &Point &Point &BEV\\
    \hline\hline
    \end{tabular}
\begin{tablenotes}
    \footnotesize
    \item Logg3D-Net \cite{LoGG3D-Net} was originally proposed for LPR, and can provide 6-DoF global localization by integrating with its follow-up work, SpectralGV\cite{vidanapathirana2023sgv}.
\end{tablenotes}
\end{threeparttable}
\vspace{-1em}
\end{table*}

\subsection{Motivation and Contributions}
\begin{figure}[!t]\centering
\subfigure{
\includegraphics[width=0.28\linewidth]{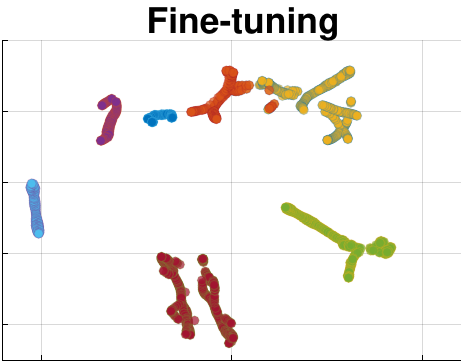}
}
\subfigure{
\includegraphics[width=0.28\linewidth]{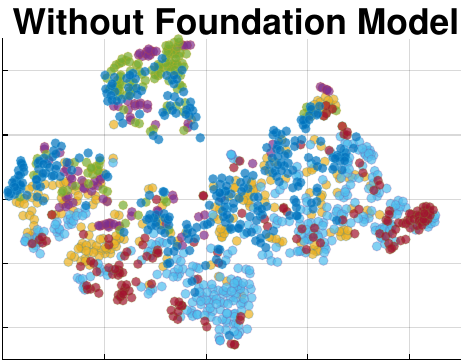}
}
\subfigure{
\includegraphics[width=0.28\linewidth]{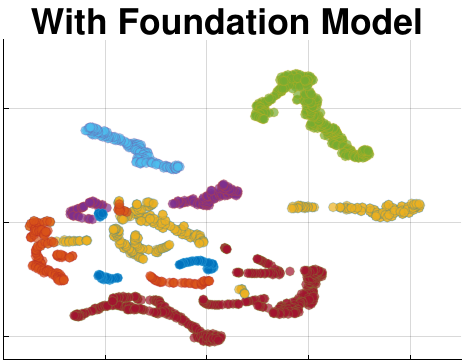}
}
\caption{t-SNE\cite{tsne} visualization of LPR. We select 7 distinct locations to visualize the discriminability of the LPR descriptors.}\label{Fig: tsne-Motivation}
\vspace{-2em}
\end{figure}
In view of the aforementioned analysis, as summarized in Table~\ref{tab: related_work}, the uniformity illustrated in Fig.~\ref{Fig: title_fig} is commonly neglected \cite{BEVPlace++,lu2025ringSharp} or only partially considered \cite{isc,Solid,chen2021overlapnet,lcdnet,LoGG3D-Net,jung2025imlpr,mwjung-2025-icra} in modern LPR methods.
Moreover, most existing methods either focus exclusively on LPR\cite{jung2025imlpr,isc,mwjung-2025-icra} or provide only partial solutions for LGL, such as 1-DoF heading alignment\cite{Solid,chen2021overlapnet} or 3-DoF pose estimation constrained to SE(2)\cite{BEVPlace++,ring++,lu2025ringSharp}.
These limitations motivate us to develop a uniform LGL system, called UniLGL, that enables fully 6-DoF global pose estimation over SE(3), while simultaneously preserving spatial and material uniformity as well as sensor-type uniformity.
A further observation from Table~\ref{tab: related_work} is that foundation-model-empowered LGL systems remain largely unexplored, due to the lack of LiDAR foundation models pre-trained on large-scale and diverse datasets. This motivates us to explore the integration of VFMs into the LGL task. As shown in Fig.~\ref{Fig: tsne-Motivation}, the strong cross-task generalization capabilities of VFMs provide the LPR network with an initial discrimination capability. By fine-tuning with a small amount of LiDAR data, the domain gap between the foundation models and the LGL task can be effectively bridged, enabling high-performance LGL.
The main contributions of this paper are listed as follows:
\begin{itemize}
    \item     
    \textbf{Uniform LPR}: An end-to-end LPR network is designed to provide a uniform place representation. UniLGL fuses the spatial BEV images and intensity BEV images of LiDAR scans through a novel feature fusion network to achieve spatial and material uniformity, and a viewpoint invariance hypothesis is introduced to supervise UniLGL with sensor-type uniformity, which hypothesis replaces the conventional translation equivariance hypothesis commonly used in conventional LPR networks.
    \item \textbf{Global Localization}: UniLGL provides a complete global localization framework that achieves both LPR and 6-DoF pose estimation on SE(3) without requiring additional point cloud registration.
    Unlike conventional image-based LGL methods, which are typically limited to 1-DoF or 3-DoF pose estimation, the proposed method enforces local feature consistency within the network. This enables robust 6-DoF global localization through local feature matching between BEV images.
    \item \textbf{Foundation Model}: Following the paradigm shift of the task-agnostic foundation model in Natural Language Processing (NLP), we explore the incorporation of foundation models into LGL, and an adaptation strategy is proposed to incorporate the foundation model originally designed for single-image feature extraction into our multi-BEV-image fusion network.
    By leveraging the strong generalization capability of foundation models, UniLGL is able to deliver effective performance using only a small amount of LiDAR data for fine-tuning.
    \item \textbf{Fully-fledged}:
    UniLGL is a full-fledged LGL framework that has been validated across multiple public datasets as well as real-world applications.
    Extensive benchmark comparisons on public datasets demonstrate that UniLGL delivers competitive place recognition and global localization performance compared with SOTA LGL methods.
    Beyond benchmark evaluations, UniLGL has been further extended to various real-world industrial and field applications, including autonomous driving trucks and collaborative exploration with multi-MAV systems, to demonstrate its performance and industrial applicability.
\end{itemize}
\section{Learning Uniform Global Descriptor for LiDAR Place Recognition} \label{Sec: Learning Uniform Global Descriptor for LiDAR Place Recognition}
Given a query point cloud $\mathbb{P}_q$ and a set of database point cloud ${\mathbb{M}} \buildrel \Delta \over = \{ {\mathbb{P}_{db,1}},{\mathbb{P}_{db,2}},\ldots,{\mathbb{P}_{db,n}}\}$, LPR is in charge of retrieving the most similar point cloud to $\mathbb{P}_q$ from $\mathbb{M}$.
\begin{equation}
\small{
\hat i = \mathop {\arg \max }\limits_{i = 1, \ldots ,n} S\left( {{{\mathbb{P}}_q},{{\mathbb{P}}_{db,i}}} \right) \label{Eq: LPR}
}
\end{equation}
where $S(\cdot)$ measures the similarity of two point clouds.
In this work, a mapping function $\Phi:\mathbb{P}\rightarrow\mathbf{D}$ is developed that represents the point cloud with a global descriptor $\mathbf{D}$ and transforms the LPR problem (\ref{Eq: LPR}) into a descriptor matching problem.
\begin{equation}
\small{
\hat i = \mathop {\arg \min }\limits_{i = 1, \ldots ,n} {\left\| {{{\bf{D}}_q} - {{\bf{D}}_{i}}} \right\|_2}\label{Eq: descriptor matching problem}
}
\end{equation}
where ${{\bf{D}}_q}$ and ${{\bf{D}}_{i}}$ are descriptors of $\mathbb{P}_{q}$ and $\mathbb{P}_{db,i}$, respectively.
The descriptor matching problem (\ref{Eq: descriptor matching problem}) can be easily solved by performing a K-Nearest Neighbor (KNN) search.
Therefore, the key to solving the LPR problem (\ref{Eq: LPR}) lies in learning a mapping function $\Phi:\mathbb{P}\rightarrow\mathbf{D}$ that ensures point clouds with a high similarity yield highly similar descriptors, and vice versa.

\subsection{Learning Spatial and Material Uniformity} \label{Sec: Learning Spatial and Material Uniformity}
LiDAR is a 4D sensor that captures point clouds $\mathbb{P} \in \mathbb{R}^{N \times 4}$, where each point $\mathbf{p} = [p_x,p_y,p_z, I]^\top \in \mathbb{R}^4$ is represented by its 3D spatial coordinates $[p_x,p_y,p_z]$ and intensity $I$.
To achieve the spatial and material uniformity,
we represent the point cloud using two complementary BEV images, called spatial BEV image $\mathbf{I}_s\in \mathbb{R}^{H\times W}$ and intensity BEV image $\mathbf{I}_I\in \mathbb{R}^{H\times W}$, and learning LPR with an elaborate feature fusion network.
\subsubsection{BEV Image Projection Model}
For spatial BEV image, each pixel value $\mathbf{I}_s(u,v)$ is computed as the normalized point density.
\begin{equation}
\small{
    \mathbf{I}_s(u,v) = \frac{{\left|{{\mathbb{P}}_{uv}}\right| - \mathop {\min }\limits_{u,v} \left( {\left|{{\mathbb{P}}_{uv}}\right|} \right)}}{{\mathop {\max }\limits_{u,v} \left( {\left|{{\mathbb{P}}_{uv}}\right|} \right) - \mathop {\min }\limits_{u,v} \left( {\left|{{\mathbb{P}}_{uv}}\right|} \right)}} \label{Eq: spatialBEV}
    }
\end{equation}
where ${{\mathbb{P}}_{uv}} = \left\{ {{\bf{p}} \in {\mathbb{P}}|\left\lfloor {\frac{{{p_x}}}{r}} \right\rfloor  = u,\left\lfloor {\frac{{{p_y}}}{r}} \right\rfloor  = v} \right\}$ denotes the point cloud corresponding to the pixel $[u, v]$, $r$ is the resolution of BEV images, $|\cdot|$ returns the cardinal number of a set, and $\left\lfloor \cdot \right\rfloor$ denotes the floor operation.

\begin{figure}
    \centering
    \includegraphics[width=\linewidth]{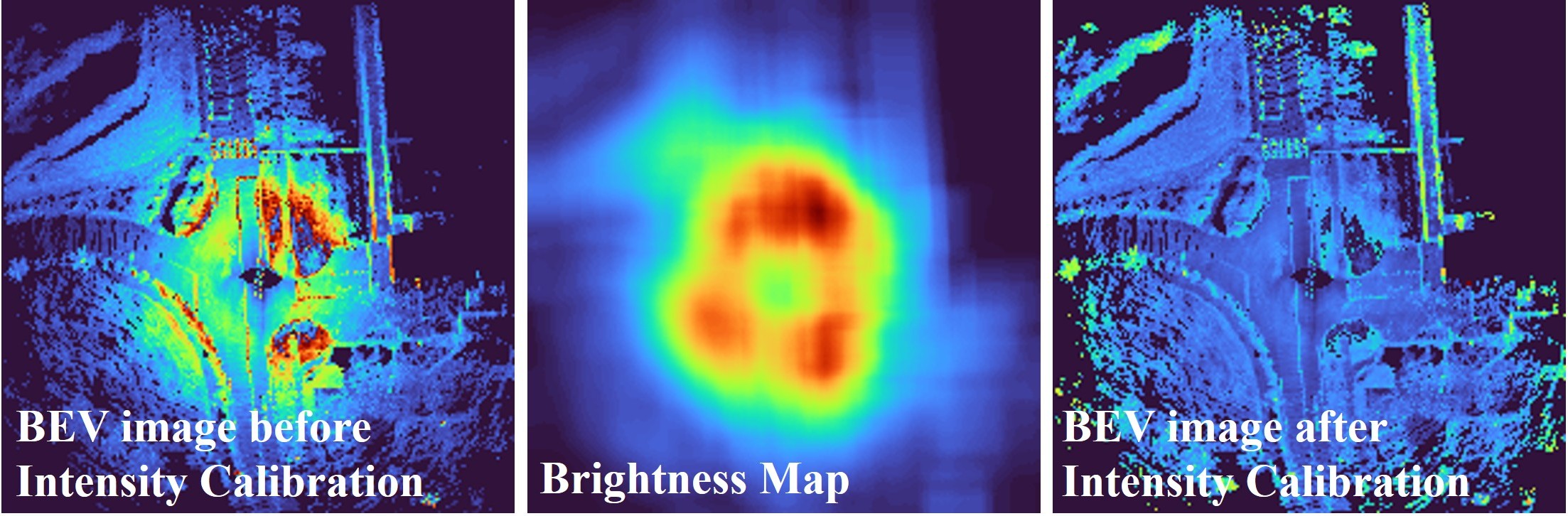}
    \vspace{-1.0em}
    \caption{Demonstration of the intensity calibration.}
    \label{Fig: intensity_calib}
\end{figure}

In contrast to geometric information, LiDAR intensity measurements are influenced not only by the material of the surface but also by factors such as distance and incidence angle.
To mitigate the influence of non-material factors, a brightness image $\mathbf{I}_b$ is introduced, which is constructed by averaging the intensity values within a neighborhood window.
\begin{equation}
\small{
\mathbf{I}'_I(u,v) = \frac{{\max \left[ {I({{\mathbb{P}}_{uv}})} \right]}}{{{\mathbf{I}_b}(u,v) + 1}} \label{Eq: intensity_calib}
}
\end{equation}
where $I(\cdot)$ returns the intensity of a point cloud.
As illustrated in Fig. \ref{Fig: intensity_calib}, the brightness image provides a simple yet effective solution for LiDAR intensity calibration.
Then, the intensity BEV image is defined as
\begin{equation}
\small{
{\mathbf{I}_I}(u,v) = \frac{{\mathbf{I}'_I(u,v) - {\min (\mathbf{I}'_I)}}}{{{\max (\mathbf{I}'_I)} - {\min (\mathbf{I}'_I)}}}\label{Eq: intensity_norm}
}
\end{equation}
which rescales each pixel value in the intensity BEV image to lie within the range $[0,1]$.

\begin{figure*}
    \centering
    \includegraphics[width=\linewidth]{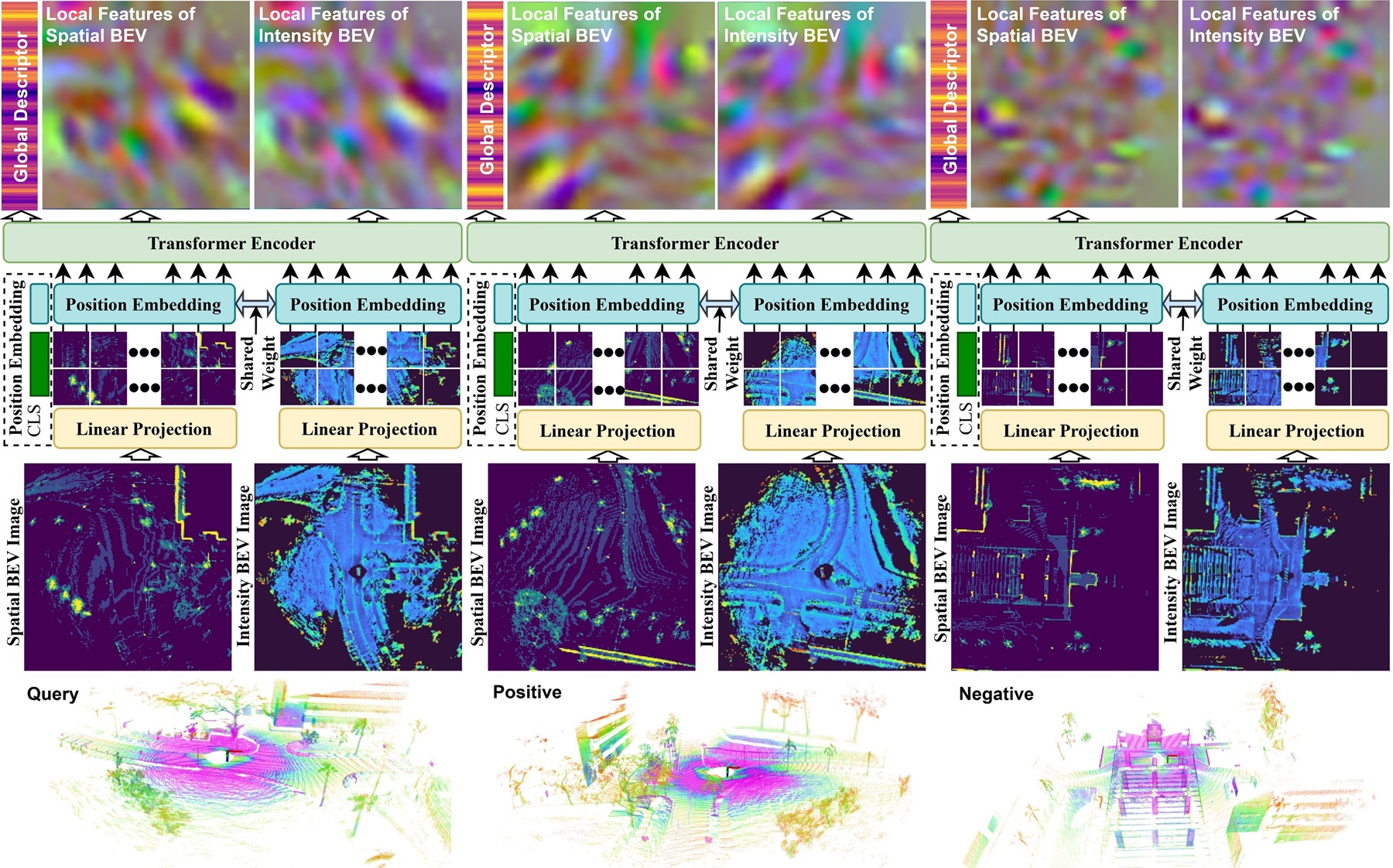}
    \caption{Network architecture of UniLGL for learning uniform place recognition.}
    \label{Fig: NetworkArchitecture}
\end{figure*}
\subsubsection{End-to-end BEV Images Fusion LPR Network}\label{Sec:End-to-end BEV images fusion LPR network}
To extract the uniform descriptor of the spatial and intensity BEV images, we extend the impressive ViT \cite{ViT} network to enable feature fusion.
As illustrated in Fig. \ref{Fig: NetworkArchitecture}, BEV images are split into a sequence of flattened 2D patches $\bm{\rho}\in\mathbb{R}^{C\times C}$, where $C$ is the resolution of each image patch.
Two independent Convolutional Neural Networks (CNNs) are employed to learn the patch embedding projection for spatial and intensity BEV images, respectively, and a learnable classification vector $\texttt{CLS}\in\mathbb{R}^D$ is augmented to the sequence of embedded patches.
\begin{equation}
\small{
\mathbf{L} = \left[ {\texttt{CLS},{\bm{\rho }}_s^1{{\bf{E}}_s}, \ldots ,{\bm{\rho }}_s^M{{\bf{E}}_s},{\bm{\rho }}_I^1{{\bf{E}}_I}, \ldots ,{\bm{\rho }}_I^M{{\bf{E}}_I}} \right]
}
\end{equation}
where ${\bm{\rho}}_s^i$ and ${\bm{\rho}}_I^i$ denote the $i$-th patch of the spatial BEV image and the intensity BEV image, respectively; $\mathbf{E}_s \in\mathbb{R}^{C\times C\times D}$ and $\mathbf{E}_I\in\mathbb{R}^{C\times C\times D}$ represent the learnable embedding projections for the spatial and intensity BEV images, and $D$ is the dimension of each embedded patch.
The resulting embedded patches are concatenated with the classification vector $\texttt{CLS}$ into $\mathbf{L} \in \mathbb{R}^{(2M + 1) \times D}$, where $M$ is the number of patches per BEV image.
A standard learnable position embedding is employed on $\mathbf{L}$ to retain positional information, the resulting sequence of embedding vectors $\mathbf{Z}$ serves as input to the transformer encoder.
\begin{equation}
\small{
{\bf{Z}} = {\bf{L}} + {{\bf{E}}_{pos}},{{\bf{E}}_{pos}} = \left[ {{\bf{E}}_{pos}^{cls},{\bf{E}}_{pos}^s,{\bf{E}}_{pos}^I} \right]
}
\end{equation}
where ${\bf{E}}_{pos}^{cls}$, ${\bf{E}}_{pos}^s$, and ${\bf{E}}_{pos}^I$ are position embeddings of classification vector $\texttt{CLS}$, embedded patches of spatial BEV image, and embedded patches of intensity BEV image, respectively.

The Transformer encoder\cite{transformer}, composed of alternating Multi-head Self-Attention (MSA) and Multi-Layer Perceptron (MLP) blocks, is employed to learn both the global descriptor (i.e., classification token) for LPR and the local features (i.e., local feature token) of each patch.
\begin{equation}
\small{
\left[ {{\bf{Z}}_T^{cls},{\bf{Z}}_T^s,{\bf{Z}}_T^I} \right] = \text{MLP}\left( {\text{MSA}({\bf{Z}})} \right),
{\bf D} = {\bf{Z}}_T^{cls}
}
\end{equation}
where ${\bf{Z}}_T^{cls}\in\mathbb{R}^{D}$ is the joint-image-level classification token, and ${\bf{Z}}_T^{s}\in\mathbb{R}^{M\times D}$ and ${\bf{Z}}_T^{I}\in\mathbb{R}^{M\times D}$ are patch-level local feature tokens, respectively.
The joint-image-level classification token ${\bf{Z}}_T^{cls}$ is treated as the unified global descriptor for LPR.
\subsection{Learning Sensor-Type Uniformity} \label{sec: Learning Sensor Type Uniformity}
In this section, we begin a discussion with the \textit{Translation Equivariance} hypothesis, which is a fundamental hypothesis in conventional learning-based LPR.
\begin{hypothesis}
\textit{
(Translation Equivariance) An effective global descriptor $\mathbf{D}$ should ensure that point clouds captured at spatially proximate states yield highly similar descriptors, and vice versa.
\begin{equation}
\small{
\left\| {{{\bf{t}}_q} - {{\bf{t}}_i}} \right\| \le \left\| {{{\bf{t}}_q} - {{\bf{t}}_j}} \right\| \Rightarrow \left\| {{{\bf{D}}_q} - {{\bf{D}}_i}} \right\| \le \left\| {{{\bf{D}}_q} - {{\bf{D}}_j}} \right\| \label{Eq: Translation Equivariance}
}
\end{equation}
where ${{\bf{t}}_q}$ denotes the translation state of the robot capturing the query point cloud, and ${{\bf{t}}_i}$ and ${{\bf{t}}_j}$ denote the translation states of the robot capturing the database point cloud.}\label{Hyp: Translation Equivariance}
\end{hypothesis}
\begin{definition}
\textit{
(Rotation Invariance) If a mapping function $\Phi(\cdot)$ is rotation invariant, it satisfies the following equation.
\begin{equation}
\small{
    \Phi(\mathbb{P}) = \Phi(\mathbb{P}_R)
    }
\end{equation}
where $\mathbb{P}_R$ denotes the point cloud $\mathbb{P}$ transformed by an arbitrary rotation $\mathbf{R} \in \mathrm{SO}(3)$.}
\end{definition}

An important implication of Hypothesis~\ref{Hyp: Translation Equivariance} is that the global descriptor extraction network should exhibit \textit{rotation invariance}, ensuring that point clouds captured at the same location but under different rotations yield identical descriptors.
\textit{However, as illustrated in Fig.~\ref{Fig: CorrespondingPointCloudUnderTranslationEquivariance}, LPR networks with built-in rotation invariance are not well-suited for FoV-limited LiDARs.}
To mitigate this limitation, a viewpoint invariance hypothesis is proposed to supervise the LPR network.
\begin{hypothesis}
\textit{(Viewpoint Invariance) An effective global descriptor extraction network should ensure that point clouds capturing similar scenes from arbitrary viewpoints yield similar descriptors, and conversely, that dissimilar scenes produce dissimilar descriptors.} \label{Hyp: Viewpoint Invariance}
\end{hypothesis}

To give a mathematical formulation of viewpoint invariance, the Intersection over Union (IoU) of the convex hulls of point clouds is employed to assess whether the two point clouds capture the same scene, even when observed from different viewpoints.
\begin{equation}
\small{
{\rm IoU}({\mathbb{P}_q},{\mathbb{P}_{i}}) =
\begin{cases}
0, \left\|{\mathbf{t}_{q,z}} - {\mathbf{t}_{i,z}}\right\|_1 > \delta,\\
\dfrac{\text{Area}\!\left[ \text{Covn}({\mathbb{P}_q}) \cap \text{Covn}({\mathbb{P}_{i}}) \right]}
      {\text{Area}\!\left[ \text{Covn}({\mathbb{P}_q}) \cup \text{Covn}({\mathbb{P}_{i}}) \right]}, \text{Otherwise}
\end{cases}
\label{Eq: IoU}
}
\end{equation}
where $\mathbf{t}_{q,z}$ and $\mathbf{t}_{i,z}$ respectively denote the height of robot capturing the query point cloud $\mathbb{P}_q$ and $i$-th database point cloud $\mathbb{P}_{i}$, $\delta$ is a threshold used to determine whether $\mathbb{P}_q$ and $\mathbb{P}_{i}$ are captured at similar heights, $\text{Covn}(\cdot)$ returns the convex hull of a point cloud,
and $\text{Area}\left(\cdot \right)$ denotes the area measurement.
The computation of the IoU between 3D convex hulls is time-consuming and can considerably degrade the training efficiency.
To address this issue, a simple yet efficient approximation is employed in this work, where the IoU is calculated between the 2D convex hulls of point clouds projected onto the horizontal plane.
Furthermore, to prevent false overlaps caused by point clouds captured at different heights, the IoU is set to zero when their height difference exceeds a predefined threshold of $\delta = 3m$ in our experiments.

\begin{figure}[!t]\centering
    \includegraphics[width=\linewidth]{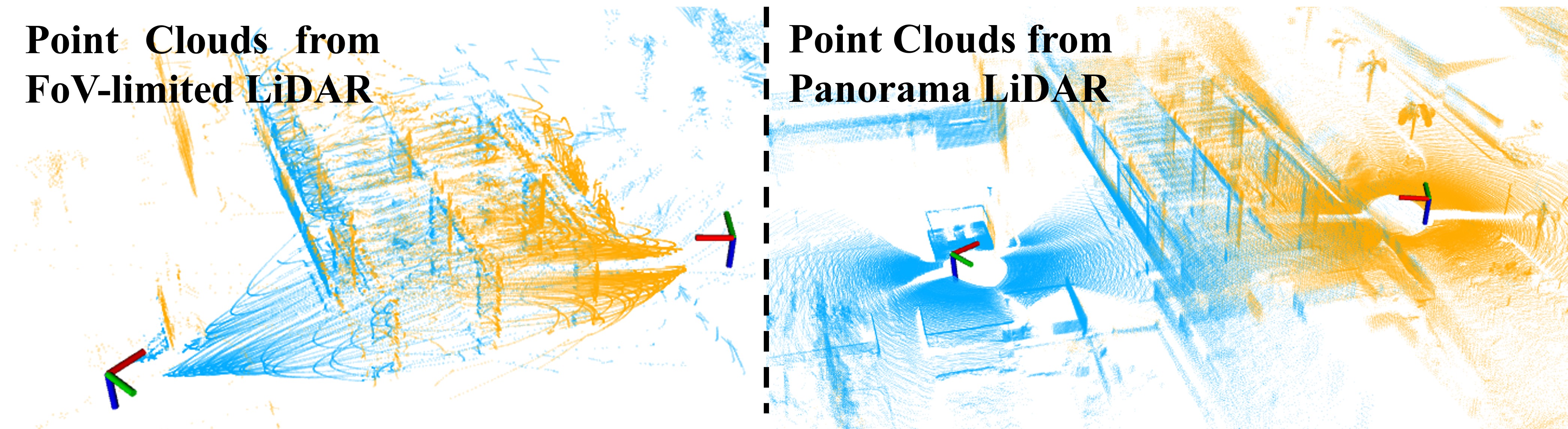}
    \vspace{-0.5em}
    \caption{Corresponding point cloud under Viewpoint Invariance Hypothesis.}\label{Fig: CorrespondingPointCloudUnderViewpointInvariance}
    \vspace{-1em}
\end{figure}

According to (\ref{Eq: IoU}), the quantitative formulation of the viewpoint invariance hypothesis can be formulated as
\begin{equation}
\small{
    {\rm{IoU}}({\mathbb{P}_q},{\mathbb{P}_{i}}) \ge {\rm{IoU}}({\mathbb{P}_q},{\mathbb{P}_{j}}) \Rightarrow \left\| {{{\bf{D}}_q} - {{\bf{D}}_{i}}} \right\| \le \left\| {{{\bf{D}}_q} - {{\bf{D}}_{j}}} \right\|
}
\end{equation}
which states that point clouds exhibiting a more considerable overlap are expected to have more similar global descriptors.
As shown in Fig.~\ref{Fig: CorrespondingPointCloudUnderViewpointInvariance}, viewpoint invariance serves as a unified hypothesis for both FoV-limited and panoramic LiDAR, which defines positive point clouds based on scene similarity rather than spatial proximity.
Moreover, for panoramic LiDARs, LPR methods based on the \textit{Viewpoint Invariance} hypothesis are expected to achieve superior performance compared to those relying on the \textit{Translation Equivariance} hypothesis.
For example, translation-equivariance-based LPR methods \cite{BEVPlace++,PointNetVLAD,lcdnet,lu2025ringSharp,LoGG3D-Net,jung2025imlpr} typically use a distance-based criterion and would treat point clouds with substantial co-visible regions as negative samples due to their spatial separation.

\subsubsection{Image-level Sensor-type Uniformity} An intuitive approach to enforcing sensor-type uniformity in LPR is to supervise the descriptor extraction network using the IoU (\ref{Eq: IoU}) between the convex hull of the query point cloud and those of the positive and negative samples, respectively.
The lazy triplet loss\cite{PointNetVLAD} is adopted to supervise the place recognition process, which aims to maximize the descriptor distance between a query and negative point clouds while minimizing the descriptor distance between a query and a positive image.
\begin{equation}
\small{
{{\cal L}_{lpr}}\! =\! \mathop {\max }\limits_{{\bf{D}}_{n,i} \in \mathbb{D}_n} \!\left[ {\max \!\left(c \!+\! {{\| {{{\bf{D}}_q} \!-\! {\bf{D}}_p} \|}_2} - {{\| {{{\bf{D}}_q} - {\bf{D}}_{n,i}}\|}_2},0\right)} \right]\label{Eq: loss_global}
}
\end{equation}
where $\mathcal{L}_{lpr}$ is the LPR loss, $c$ is a constant margin, $\mathbb{D}_n$ denotes the negative global descriptor set, and $\mathbf{D}_p$ is the positive global descriptor that corresponds to the query descriptor $\mathbf{D}_q$.
In LPR, the network is expected to retrieve point clouds with very limited overlap.
However, supervising the network with pairs that have almost no overlap can confuse the model and lead to LGL degeneration due to the insufficient local feature correspondences.
Empirically, for each query point cloud, its positive samples are point clouds with IoU $>0.25$, while negative samples are point clouds with IoU $<0.2$.
\subsubsection{Patch-level Sensor-type Uniformity}
The LPR loss defined in (\ref{Eq: loss_global}) is employed to enforce the consistency of the image-level global descriptors, which neglects the consistency at the patch level.
As corroborated by \cite{LoGG3D-Net} and \cite{simeoni2025dinov3}, ignoring patch-level consistency during training can lead to the cosine similarity between the classification token and the local feature tokens gradually increasing, which means that the network tends to overlook the local information contained in BEV images, thereby diminishing the discriminability of the global descriptor.
Moreover, UniLGL is a complete LGL framework, where the feature matching process for global pose estimation (details can be found in Section~\ref{Sec: Relative Pose Estimation on Manifolds}) also requires the local features to be explicitly supervised for viewpoint invariance.

For each image pair $\{\mathbf{I}_1,\mathbf{I}_2\}$ matched via LPR, we divide both images into multiple patches and randomly select one pixel from each patch as a keypoint.
A keypoint $(u^1_i,v^1_i)$ from $\mathbf{I}_1$ located in the overlapping region between $\mathbf{I}_1$ and $\mathbf{I}_2$ can find its corresponding keypoint $(u^2_j,v^2_j)$ in $\mathbf{I}_2$ using the ground truth trajectory.
The pixel $(u^2_j,v^2_j)$ is treated as the positive pixel of $(u^1_i,v^1_i)$ while all other keypoints on $\mathbf{I}_2$ are considered negative samples.
To enable pixel-level contrastive learning, each pixel of an image is represented by a local feature $\mathbf{f}\in\mathbb{R}^D$, obtained via interpolation from the patch-level local feature token map.
We introduce the InfoNCE loss\cite{oord2018representation} to maximize the cosine similarity between the local feature $\mathbf{f}_i^1$ and its corresponding feature $\mathbf{f}_j^2$ while minimizing the cosine similarity between $\mathbf{f}_i^1$ and its negative samples.
\begin{equation}
\small{
    {{\mathcal{L}}_l}({{\bf{I}}_1},{{\bf{I}}_2}) = -\frac{1}{\left|{\mathbb{F}}_{ol}^1\right|}\sum\limits_{{\bf{f}}_i^1 \in {\mathbb{F}}_{ol}^1} {\log \left( {\frac{{\exp \left( {{\bf{f}}_i^1 \cdot {\bf{f}}_j^2} \right)}}{{\sum\limits_{{\bf{f}}_k^2 \in {{\mathbb{F}}^2}} {\exp \left( {{\bf{f}}_i^1 \cdot {\bf{f}}_k^2} \right)} }}} \right)}  \label{Eq: loss_local}
    }
\end{equation}
where $\mathcal{L}_l(\mathbf{I}_1,\mathbf{I}_2)$ denotes the local feature loss of a matched image pair $\{\mathbf{I}_1,\mathbf{I}_2\}$, which is designed to supervise the viewpoint invariance in local feature perspective; $\mathbb{F}_{ol}^1 = \{\cdots,\mathbf{f}_i^1,\cdots\}$ is the set of local features in $\mathbf{I}_1$ located within the overlapping region between $\mathbf{I}_1$ and $\mathbf{I}_2$; and $\mathbb{F}^2 = \{\cdots,\mathbf{f}_k^2,\cdots\}$ is the full local feature set of $\mathbf{I}_2$.

\subsubsection{Joint Image- and Patch-Level Sensor-type Uniformity} To achieve sensor-type uniformity in both global descriptor and local feature perspective, the final loss function $\mathcal{L}$ is designed as a linear combination of the LPR loss (\ref{Eq: loss_global}) and local feature loss (\ref{Eq: loss_local}):
\begin{equation}
\small{
\begin{aligned}
    \mathcal{L} = \mathcal{L}_{lpr} + \alpha&\left[\mathcal{L}_l(\mathbf{I}_q^s,\mathbf{I}_{db}^s) + \mathcal{L}_l(\mathbf{I}_{db}^s,\mathbf{I}_{q}^s)\right. \\&\left.+ \mathcal{L}_l(\mathbf{I}_q^I,\mathbf{I}_{db}^I) + \mathcal{L}_l(\mathbf{I}_{db}^I,\mathbf{I}_{q}^I)\right]
\end{aligned}
}
\end{equation}
where $\alpha$ is the loss balancing coefficient, empirically set to $0.125$, and $\{\mathbf{I}_q^s,\mathbf{I}_{db}^s\}$ and $\{\mathbf{I}_q^I,\mathbf{I}_{db}^I\}$ denote the spatial BEV image pair and the intensity BEV image pair, respectively.

\subsection{LPR Meets Foundation Models}
Motivated by the remarkable success of task-agnostic pre-trained representations
in NLP, we explore the incorporation of foundation models into the proposed UniLGL, which are trained on large-scale datasets to learn general-purpose features.
Owning to the efficiency for acquiring large-scale visual data, in this paper, a self-supervised VFM, DINO\cite{DINO}, is adopted to pre-initialize the feature fusion network designed in Section~\ref{Sec: Learning Spatial and Material Uniformity}.
\begin{equation}
\small{
\begin{aligned}
&{{\bf{E}}_s} = {{\bf{E}}_I} = {{\bf{E}}_{\text{DINO}}},{\bf{E}}_{pos}^{cls} = {\bf{E}}^{pos}_{\text{DINO}}[:,1]\\
&{\bf{E}}_{pos}^s = {\bf{E}}_{pos}^I = {\bf{E}}^{pos}_{\text{DINO}}[:,2:M]\\
&\text{MLP}( \cdot ) = \text{MLP}{_{\text{DINO}}}( \cdot ),\text{MSA}( \cdot ) = \text{MSA}{_{\text{DINO}}}( \cdot )
\end{aligned} \label{Eq: DINO}
}
\end{equation}
where $\mathbf{E}_{\text{DINO}}\in\mathbb{R}^{C\times C\times D}$ and $\mathbf{E}^{\text{pos}}_{\text{DINO}}\in \mathbb{R}^{(M+1)\times D}$ represent the embedding and positional encoding layers, respectively, and $\text{MSA}_{\text{DINO}}(\cdot)$ and $\text{MLP}_{\text{DINO}}(\cdot)$ denote the MSA and MLP networks, all pre-trained as part of the DINO vision transformer\cite{DINO}.
In (\ref{Eq: DINO}), we expand the use of DINO pre-trained weights, initially developed for single-image feature extraction, to support the proposed multi-image feature fusion LPR architecture.
To further bridge the domain gap, note that the BEV images obtained through the BEV image projection model (\ref{Eq: spatialBEV})-(\ref{Eq: intensity_norm}) are grayscale, while DINO is pre-trained on RGB images, we replicate the single-channel BEV images into three channels to form pseudo-RGB inputs and perform a lightweight fine-tuning using a small amount of BEV-image-represented LiDAR data.
This simple yet effective adaptation strategy enables the VFM to seamlessly integrate with the proposed LPR network, as demonstrated in Fig.~\ref{Fig: tsne-Motivation} and experiments validated in Section~\ref{Sec: Experimental Evaluations and Validations}.
\section{Robust Global Registration on Manifolds}\label{Sec: Relative Pose Estimation on Manifolds}
Global registration is a fundamental task that aims to align a query point cloud $\mathbb{P}_q$ with a database point cloud $\mathbb{P}_{db}$, where $\mathbb{P}_{db}$ is retrieved via a KNN search over the global descriptor space (\ref{Eq: descriptor matching problem}).
\subsection{Local Feature Enabled Point-Level Matching}
As demonstrated in Section \ref{Sec: Learning Spatial and Material Uniformity}, the proposed uniform LPR network extracts not only the classification token, which serves as the global descriptor of a point cloud, but also patch-level local feature tokens, denoted as $\mathbf{Z}_T^s$ and $\mathbf{Z}_T^I$. Specifically, the $i$-th patch of the spatial and intensity BEV images is represented by $\mathbf{z}^s_i = \mathbf{Z}_T^s[:,i] \in \mathbb{R}^{D}$ and $\mathbf{z}^I_i = \mathbf{Z}_T^I[:,i] \in \mathbb{R}^{D}$, respectively, where $i = 1, \ldots, M$.
To enable pixel-level matching, a dense feature map is reconstructed by interpolating the patch-level local feature tokens, allowing each pixel in the BEV image to be associated with a local feature $\mathbf{f} \in \mathbb{R}^D$.
Let $\mathbf{f}_{q,i}^s$ and $\mathbf{f}_{q,k}^I$ denote the local features of keypoints in the query spatial BEV image and intensity BEV image, respectively, with $i, k \in \{1,\dots, H \times W\}$. We then establish correspondences by conducting a KNN search within the pixel-level local feature embedding space of the database image.
\begin{equation}
\small{
\hat j \!=\! \mathop {\arg \min }\limits_{j = 1, \ldots ,H\times W} \!{\left\| {{\bf{f}}_{q,i}^s - {\bf{f}}_{db,j}^s} \right\|_2}, 
\hat l \!=\! \mathop {\arg \min }\limits_{l = 1, \ldots ,H\times W} \!{\left\| {{\bf{f}}_{q,k}^I - {\bf{f}}_{db,l}^I} \right\|_2}
\label{Eq: feature KNN}
}
\end{equation}
where $\mathbf{f}^s_{q,i}$ and $\mathbf{f}^s_{db,j}$ denote the $i$-th and $j$-th pixel-level local features of the spatial BEV images from the query and database, respectively. Similarly, $\mathbf{f}^I_{q,k}$ and $\mathbf{f}^I_{db,l}$ represent the $k$-th and $l$-th pixel-level local features of the intensity BEV images from the query and database, respectively.

Benefiting from the BEV image representation of 4D point clouds and the local feature supervision strategy designed in Section~\ref{sec: Learning Sensor Type Uniformity}, UniLGL is able to achieve point-level global matching through pixel-level matching (\ref{Eq: feature KNN}) on BEV images.
\begin{equation}
\small{
\begin{aligned}
&{{\bf{p}}_{i}^s} = \mathop {\arg \min }\limits_{{\bf{p}} \in {\mathbb{P}}_{q}^i} {\|p_z\|_1}, {{\bf{p}}_{j}^s} = \mathop {\arg \min }\limits_{{\bf{p}} \in {\mathbb{P}}_{db}^j} {\|p_z\|_1}\\
&{{\bf{p}}_{k}^I} = \mathop {\arg \max }\limits_{{\bf{p}} \in {\mathbb{P}}_{q}^k} I,
{{\bf{p}}_{l}^I} = \mathop {\arg \max }\limits_{{\bf{p}} \in {\mathbb{P}}_{db}^l} I
\end{aligned}\label{Eq: point_matching}
}
\end{equation}
where ${\mathbb{P}}_{q}^i = \left\{ {{\bf{p}} \in {\mathbb{P}_q}|\left\lfloor {\frac{{{p_x}}}{r}} \right\rfloor  = {u_i},\left\lfloor {\frac{{{p_y}}}{r}} \right\rfloor  = {v_i}} \right\}$ and ${\mathbb{P}}_{db}^j = \left\{ {{\bf{p}} \in {\mathbb{P}_{db}}|\left\lfloor {\frac{{{p_x}}}{r}} \right\rfloor  = {u_j},\left\lfloor {\frac{{{p_y}}}{r}} \right\rfloor  = {v_j}} \right\}$ denote point clouds corresponding to the $i$-th pixel of the query spatial BEV image and the $j$-th pixel of the database spatial BEV image, respectively.
Similarly, ${\mathbb{P}}_{q}^k$ and ${\mathbb{P}}_{db}^l$ respectively denote point clouds corresponding to the $k$-th pixel query intensity BEV image and the $l$-th pixel of the database intensity BEV image.

\subsection{Robust Global Pose Recovery} \label{Sec: Robust Global Pose Recovery}
According to the point-level matching result (\ref{Eq: point_matching}), the global registration problem is defined as:
\begin{equation}
\small{
\mathop {\min }\limits_{\Delta {\bf{T}}} \sum\limits_{i = 1}^Q {{{\left\| {\Delta {\bf{Tq}}_{q,i}^{s} - {\bf{q}}_{db,j}^{s}} \right\|}_2}}  + \sum\limits_{k = 1}^R {{{\left\| {\Delta {\bf{Tq}}_{q,k}^{I} - {\bf{q}}_{db,l}^{I}} \right\|}_2}}\label{Eq: relative pose estimation}
}
\end{equation}
where ${\bf{q}}_{q,i}^s=[{{p}}_{i,x}^s,{{p}}_{i,y}^s, p_{i,z}^s]^\top$ and ${\bf{q}}_{q,k}^I=[{{p}}_{k,x}^I,{{p}}_{k,y}^I, p_{k,z}^I]^\top$ are $i$-th and $k$-th points of the query point cloud $\mathbb{P}_{q}$;
${\bf{q}}_{db,j}^s$ and ${\bf{q}}_{db,l}^I$ are their corresponding points in the database point cloud $\mathbb{P}_{db}$; $Q$ and $R$ represent the number of matched keypoints associated with the spatial BEV image and the intensity BEV image, respectively; $\Delta\mathbf{T} \in \text{SE}(3)$ is the relative pose between query and database point clouds.

With the consideration of matching outlier, we assume the noise of point matching is unknown but bounded, and write the global registration problem in a Truncated Least Squares (TLS) formulation:
\begin{equation}
\small{
\Delta {\bf{\hat T}} = \mathop {\arg \min }\limits_{\Delta {{\bf{T}}}} \sum\limits_{i = 1}^{Q + R} {\min \left( {{{\left\| {\Delta {\bf{T}}{{\bf{q}}_{q,i}} - {{\bf{q}}_{db,j}}} \right\|}_2},{\xi ^2}} \right)}  \label{Eq: TLS}
}
\end{equation}
where ${{\bf{q}}_{q,i}}$ and ${{\bf{q}}_{db,j}}$ are the $i$-th and $j$-th rows of ${{\bf{q}}_{q}} = [{{\bf{q}}_q^{s,\top}},{{\bf{q}}_q^{I,\top}}]^\top$ and ${{\bf{q}}_{db}} = [{{\bf{q}}^{s,\top}_{db}},{{\bf{q}}^{I,\top}_{db}}]^\top$, respectively.
The TLS formulation of the global registration problem (\ref{Eq: TLS}) discards measurements with large residuals (when $\left\| {\Delta {\bf{T}}{{\bf{q}}_{q,i}} - {{\bf{q}}_{db,j}}} \right\|_2 > \xi^2$ the $i$-th summand does not influence the optimization).
To solve the global optimization problem (\ref{Eq: TLS}) without an initial guess, a Graduated Non-Convexity (GNC)\cite{8957085} optimization is adopted.
According to the Black-Rangarajan duality \cite{Black1996}, the TLS problem (\ref{Eq: TLS}) is equivalent to:
\begin{equation}
\small{
\Delta \mathbf{T} = \mathop {\arg\min }\limits_{{\Delta{\bf{T}}},{w_i}} \sum\nolimits_i {{w_i}{{\left\| {{\Delta{\bf{T}}}{{\bf{q}}_{q,i}} - {{\bf{q}}_{db,j}}} \right\|}_2} + \frac{{\mu \left( {1 - {w_i}} \right)}}{{\mu  + {w_i}}}} {\xi ^2} \label{Eq: Black Rangarajan Duality}
}
\end{equation}
where $w_i\in [0,1]$ and $\mu > 0$ are slack variables.
The GNC TLS problem (\ref{Eq: Black Rangarajan Duality}) can be solved by alternating optimization:
\begin{equation}
\small{
{\Delta{{\bf{\hat T}}}} = \mathop {\arg \min }\limits_{{\Delta{\bf{T}}}} \sum\nolimits_i {{{\hat w}_i}{{\left\| {{\Delta{\bf{T}}}{{\bf{q}}_{q,i}} - {{\bf{q}}_{db,j}}} \right\|}_2}}\label{Eq: GNC-1} \\
}
\end{equation}
\begin{equation}
\small{
{\bf{\hat w}} = \mathop {\arg \min }\limits_{{w_i} \in {\bf{w}}} \sum\nolimits_i {{w_i}{{\left\| {{\Delta{{\bf{\hat T}}}}{{\bf{q}}_{q,i}} - {{\bf{q}}_{db,j}}} \right\|}_2}}  + \frac{{\mu \left( {1 - {w_i}} \right)}}{{\mu  + {w_i}}}{\xi ^2} \label{Eq: GNC-2}
}
\end{equation}
where (\ref{Eq: GNC-1}) is a weighted version of the outlier-free pose estimation problem (\ref{Eq: relative pose estimation}), which can be solved globally using SVD, and (\ref{Eq: GNC-2}) can also typically be solved in closed form.

\section{Experimental Setup}\label{Sec: Experimental Setup}
During experiments, the dimension of the joint-image-level classification token and patch-level local feature token is set to $D=384$, and DINO-ViTs-8\cite{DINO} is introduced as the foundation model of the proposed UniLGL.
We use the Adam optimizer with a learning rate of $2\times10^{-5}$ and a weight decay of $0$.

The resolution of BEV images is set to $r=0.4 m$, and the point cloud cropping range is set to $40 m$ ($[-40,40]$ for panoramic LiDAR, and $[0,80]$ for Fov-limited LiDAR), resulting in the BEV images being scaled to $H\times W= 201\times 201$.
Each experiment conducted in Section~\ref{Sec: Experimental Evaluations and Validations} is evaluated on a computer equipped with an Intel Core i9-13900KF and an NVIDIA GeForce RTX 4080.
For a fair comparison, all methods listed in Section~\ref{Sec: Comparison baseline} are retrained for 20 epochs following their default configurations on sequences \textit{ntu\_day\_01}, \textit{ntu\_night\_08}, \textit{Snail\_81R\_01}, and \textit{Garden\_db}.
The details of the data sequences are summarized in Section~\ref{Sec: Datasets for Evaluation}.
\subsection{Datasets} \label{Sec: Datasets for Evaluation}
UniLGL aims to enable uniform global localization for various types of LiDARs in diversified environments.
To demonstrate the effectiveness of the proposed method, experiments are conducted on three representative datasets, which encompass a variety of environments including campus areas, urban roadways, and highly repetitive unstructured scenes.
\subsubsection{MCD\cite{nguyen2024mcd}} 
The MCD dataset includes both FoV-limited (Livox Mid-70) and panoramic (Ouster OS1-128) LiDAR measurements collected in campus environments.
For evaluation, five high-speed sequences, including \textit{ntu\_day\_01},  \textit{ntu\_day\_02}, \textit{ntu\_day\_10}, \textit{ntu\_night\_08}, and \textit{ntu\_night\_13}, are selected, that feature numerous loop closures and span both daytime and nighttime conditions.
During the experiments, \textit{ntu\_night\_08} served as the database sequence.
To enable a fair comparison of benchmark methods under different LiDAR configurations, we use the Livox Mid-70 and the Ouster OS1-128 as sensor inputs and generate the corresponding data sequences, referred to as \textit{Mid\_NTU\_XX} and \textit{OS\_NTU\_XX}, respectively.
\subsubsection{Snail\cite{huai2024snail}}
The Snail dataset was collected using a roof-mounted panoramic LiDAR (Hesai Pandar XT32) and contains extensive urban driving data featuring dynamic objects and high-rise buildings.
For evaluation, three sequences, including \textit{20240116-2}, \textit{20240123-2}, and \textit{20240116-3}, are selected, covering a total distance of real urban driving scene over $25km$.
In the experiments, each of the above three sequences is divided based on travel distance. Specifically, the first $50\%$ of each trajectory is used as the database, and the remaining $50\%$ is used as the query set for evaluating the benchmark methods.
The resulting split sequences are named \textit{Snail\_81R\_01}, \textit{Snail\_81R\_02}, and \textit{Snail\_81R\_03}, respectively.
\subsubsection{Garden\cite{Mag-MM}}
The Garden dataset was collected with a Husky mobile robot platform mounted with a panoramic LiDAR (Ouster OS1-32) for robot global localization in unstructured and highly repetitive environments, which are characterized by dense vegetation and symmetric, repetitive paths.
To evaluate the long-term global localization performance of UniLGL on FoV-limited LiDARs, we further augmented the Garden dataset using a mobile robot platform equipped with a FoV-limited LiDAR (Livox Avia) and a panoramic LiDAR (Robosense Helios 32).
Five new sequences, namely \textit{Garden\_db}, \textit{Garden\_01}, \textit{Garden\_02}, \textit{Garden\_03}, and \textit{Garden\_04}, are augmented to the original Garden dataset, which are collected in the same region of the Garden dataset.
Notably, the original four Garden sequences, referred to as \textit{Garden\_LT\_01}, \textit{Garden\_LT\_02}, \textit{Garden\_LT\_03}, and \textit{Garden\_LT\_04}, provide long-term measurements across 8 months when compared with the 5 augmented data sequences.
During the experiments, \textit{Garden\_db} is used as the database, while \textit{Garden\_01} to \textit{Garden\_04} are employed to evaluate the localization performance of UniLGL on FoV-limited LiDARs. In addition, \textit{Garden\_LT\_01} to \textit{Garden\_LT\_04} are utilized to assess the long-term generalization ability of UniLGL.

\subsection{Comparison Baseline} \label{Sec: Comparison baseline}
UniLGL aims to achieve uniform global localization across spatial and material domains, as well as between FoV-limited and panoramic LiDAR observations.
To demonstrate the effectiveness of the proposed method, we present detailed quantitative analyses comparing UniLGL with SOTA LPR/LGL methods, including BEVPlace++\cite{BEVPlace++}, LoGG3D-Net\cite{LoGG3D-Net}, and RING++\cite{ring++}.
Moreover, to investigate the impact of introducing spatial and material uniformity as well as sensor-type uniformity into LGL, we conduct an ablation study on the proposed UniLGL by evaluating it under various configurations.
Details of the baselines are given as follows:   
\begin{itemize}
    \item \textit{BEV-Place++}\cite{BEVPlace++}: a LGL method based on spatial BEV image representation.
    The BEV-image-represented point cloud enables LPR and global pose recovery through a rotation-invariant image similarity detection network and image registration, respectively.
    \item \textit{LoGG3D-Net}\cite{LoGG3D-Net}: an end-to-end LPR method that accounts for spatial and material uniformity by learning global place recognition descriptors directly from raw 4D point clouds.
    To complement the limitations of LoGG3D-Net as a pure LPR method, SpectralGV\cite{vidanapathirana2023sgv} is integrated to provide 6-DoF metric localization capability.
    \item \textit{RING++}\cite{ring++}: a handcrafted LGL method with sensor-type uniformity, which constructs translation-invariant descriptors and orientation-invariant metrics over the BEV Radon sinogram by leveraging the spatial information of the point cloud.
    \item \textit{UniLGL}: the full algorithm proposed in this paper that comprehensively considers spatial and material uniformity together with sensor-type uniformity.
    \item \textit{UniLGL w/o Intensity}: utilizes only the spatial information of the point cloud for global localization by extracting LPR descriptors and local features from the spatial BEV image.
    \item \textit{UniLGL w/o Spatial}: utilizes only partial spatial and intensity information from the point cloud for global localization by extracting LPR descriptors and local features from the intensity BEV image while eliminating height information from the point cloud.
    \item \textit{UniLGL w/o Loc. Feat.}: achieves spatial and material uniformity using the features fusion network described in Section~\ref{Sec: Learning Spatial and Material Uniformity}, but only considers sensor-type uniformity at the global descriptor level, neglecting pixel-level viewpoint invariance supervised by the local feature loss $\mathcal{L}_l$ defined in (\ref{Eq: loss_local}).
    \item \textit{UniLGL w/o FM}: UniLGL trained from scratch without foundation model initialization.
\end{itemize}
It is worth noting that the aforementioned methods achieve sensor-type uniformity by supervising the viewpoint invariance defined in Hypothesis~\ref{Hyp: Viewpoint Invariance}.
In contrast, the original BEVPlace++\cite{BEVPlace++} and LoGG3D-Net\cite{LoGG3D-Net} are designed based on the translation equivariance hypothesis defined in (\ref{Eq: Translation Equivariance}).
To validate the effectiveness of the proposed viewpoint invariance supervision, we additionally train UniLGL (called \textit{UniLGL Dis. Sup.}) and the SOTA LGL method BEVPlace++ (called \textit{BEVPlace++ Dis. Sup.}) using the conventional translation equivariance hypothesis for comparison.
\begin{itemize}
    \item \textit{UniLGL Dis. Sup.}: UniLGL is supervised by the translation equivariance hypothesis.
    Following the training strategy used in BEVPlace++, for each query frame, its positive samples are the ones within $5m$ away from itself and its negative samples are the other frames.
    \item \textit{BEVPlace++ Dis. Sup.}\cite{BEVPlace++}: BEVPlace++ is supervised by the translation equivariance hypothesis, which is consistent with its original configuration.
\end{itemize}
\section{Experimental Evaluations and Validations}\label{Sec: Experimental Evaluations and Validations}
\subsection{Evaluation of Place Recognition}\label{Sec: Place Recognition}
For place recognition evaluation, the IoU/distance between two point clouds is used to determine whether a retrieved match is correct.
For methods supervised by the translation equivariance hypothesis (Hypothesis~\ref{Hyp: Translation Equivariance}), such as \textit{UniLGL Dis. Sup.} and \textit{BEVPlace++ Dis. Sup.}, point cloud pairs with distances below $5m$ are chosen as positive place recognition samples.
For other methods listed in Section~\ref{Sec: Comparison baseline}, point cloud pairs with $\text{IoU}>0.25$ are chosen as positive place recognition samples.
In this experiment, three metrics are leveraged to assess the performance of all methods listed in Section~\ref{Sec: Comparison baseline}.
\begin{itemize}
    \item \textit{Top-1 Recall}: For each query, we find its nearest descriptor and retrieve the Top-1 match from the database.
    According to the IoU/distance threshold, we determine whether the prediction is a True Positive (TP), False Positive (FP), or False Negative (FN). The Top-1 recall rate is defined as the ratio of TP overall positives:
    \begin{equation}
    \small{
        \text{Recall} = \frac{\text{TP}}{\text{TP} + \text{FN}}
        }
    \end{equation}
    \item \textit{Average Precision}: Precision is computed as the ratio of TP overall predicted positives:
    \begin{equation}
    \small{
        \text{Precision} = \frac{\text{TP}}{\text{TP} + \text{FP}}
        }
    \end{equation}
    By setting different descriptor distance thresholds, the corresponding precision and recall pair can be calculated. The average precision is the area under the Precision-Recall curve.
    \item \textit{Precision–recall curve}: A curve that plots the precision and recall of the retrieval results as the descriptor distance threshold changes.
\end{itemize}

\begin{table*}[!t] \centering
\centering
\caption{The Comparison of Recall ($\%$) at Top-1/Average Precision ($\%$).}
\renewcommand{\arraystretch}{1.2}
\label{tab: recall}
\subfigure[LPR using heterogeneous LiDARs]{
\label{tab: recall-MCD}
\begin{adjustbox}{max width=\textwidth}
\begin{threeparttable}
    \setlength{\tabcolsep}{3.0pt} 
    \begin{tabular}{l| l|| c c c c c c c c|| c c}
    \hline\hline
    {} &{Sequence} &{UniLGL} &\makecell{UniLGL\\w/o Intensity} &\makecell{UniLGL\\w/o Spatial} &\makecell{UniLGL\\w/o Loc. Feat.} &\makecell{UniLGL\\w/o FM} &{BEVPlace++} &{Logg3D-Net} &{RING++} &\makecell{UniLGL\\Dis. Sup.} &\makecell{BEVPlace++\\Dis. Sup.}	\\\hline
    \multirow{4}{*}{\rotatebox{90}{FoV-limited}}
     &Mid\_NTU\_02 &{\color{blue}\bf99.87}/{\color{blue}\bf88.88} &{\color{red}\bf99.73}/85.82 &98.72/{\color{red}\bf87.35} &{\bf99.20}/\bf86.69 &76.58/17.11 &88.56/60.56 &80.38/45.22 &41.63/8.33 &80.73/75.20 &59.94/40.34\\
     &Mid\_NTU\_10 &{\color{red}\bf95.75}/{\color{blue}\bf86.43} &{\color{blue}\bf95.82}/{\bf83.87} &{\bf95.70}/{\color{red}\bf83.95} &94.60/83.49 &90.53/26.13 &94.07/60.02 &91.59/45.04 &60.98/9.56 &91.78/79.44 &92.98/64.29\\
     &Mid\_NTU\_13 &{\color{blue}\bf98.71}/{\color{blue}\bf88.38} &{\color{red}\bf98.58}/85.39 &{\bf98.36}/{\color{red}\bf87.21} &96.71/\bf85.42 &72.68/12.76 &93.25/64.53 &76.37/40.62 &42.03/6.76 &81.24/74.14 &56.09/44.23\\\cline{2-12}
     &Average     &{\color{blue}\bf98.11}/{\color{blue}\bf87.90} &{\color{red}\bf98.04}/85.03 &{\bf97.59}/{\color{red}\bf86.17} &96.84/{\bf85.20} &79.93/18.67 &91.96/61.70 &82.78/43.63 &48.21/8.22 &84.58/76.26 &69.67/49.62\\\hline\hline
     
     \multirow{4}{*}{\rotatebox{90}{Panoramic}}
     &OS\_NTU\_02 &{\color{blue}\bf100.0}/89.93 &{\color{blue}\bf100.0}/86.19 &{\color{blue}\bf100.0}/87.57 &{\color{blue}\bf100.0}/\bf92.83 &{\color{red}\bf98.54}/85.71 &{\color{blue}\bf100.0}/76.56 &98.32/74.75 &77.72/24.52 &{\bf98.48}/{\color{blue}\bf94.37} &{98.33}/{\color{red}\bf94.09}\\
     &OS\_NTU\_10 &{\color{red}\bf99.94}/88.49 &99.75/88.95 &{\color{blue}\bf100.0}/88.39 &{\bf99.92}/\bf90.64 &99.78/84.55 &99.69/76.76 &99.10/62.91 &78.46/26.32 &{\color{blue}\bf100.0}/{\color{red}\bf97.23} &{\color{blue}\bf100.0}/{\color{blue}\bf97.31}\\
     &OS\_NTU\_13 &{\color{blue}\bf100.0}/\bf90.81 &{\color{blue}\bf100.0}/87.37 &{\color{blue}\bf100.0}/89.23 &{\color{blue}\bf100.0}/90.72 &98.30/85.78 &{\color{red}\bf99.78}/76.99 &88.99/62.77 &62.28/18.08 &{\bf99.71}/{\color{red}\bf92.65} &99.31/{\color{blue}\bf93.09}\\     \cline{2-12}
     &Average     &{\color{red}\bf99.98}/89.74 &{\bf99.92}/87.50 &{\color{blue}\bf100.0}/83.40 &99.98/{\bf91.40} &98.87/85.35 &99.82/76.77 &95.47/66.81 &72.82/22.97 &99.40/{\color{red}\bf94.75} &99.21/{\color{blue}\bf94.83}\\
    \hline\hline
    \end{tabular}
\end{threeparttable}
\end{adjustbox}
}
\subfigure[LPR in repetitive scenarios]{
\begin{adjustbox}{max width=\textwidth}
\begin{threeparttable}\label{tab: recall garden}
    \setlength{\tabcolsep}{4.0pt} 
    \begin{tabular}{l|| c c c c c c c c|| c c}
    \hline\hline
    {Sequence} &{UniLGL} &\makecell{UniLGL\\w/o Intensity} &\makecell{UniLGL\\w/o Spatial} &\makecell{UniLGL\\w/o Loc. Feat.} &\makecell{UniLGL\\w/o FM} &{BEVPlace++} &{Logg3D-Net} &{RING++} &\makecell{UniLGL\\Dis. Sup.} &\makecell{BEVPlace++\\Dis. Sup.}	\\\hline
    Garden\_01 &{\color{blue}\bf97.67}/{\color{blue}\bf83.92} &{\bf96.81}/\bf82.61 &{\color{red}\bf97.30}/{\color{red}\bf83.05} &95.15/79.77 &88.49/17.84 &95.24/59.76 &90.67/54.18 &86.59/15.60 &84.33/78.46 &84.21/35.58\\
    Garden\_02 &{\color{blue}\bf98.65}/{\color{blue}\bf84.87} &{\color{red}\bf98.52}/{\color{red}\bf83.12} &{\bf98.19}/\bf82.88 &93.87/78.09 &87.29/18.62 &94.64/63.19 &91.02/45.84 &86.89/16.08 &81.59/75.70 &81.20/37.16\\
    Garden\_03 &{\color{blue}\bf98.63}/{\color{blue}\bf85.45} &{\bf96.15}/{\color{red}\bf82.87} &{\color{red}\bf97.33}/\bf77.54 &91.57/70.55 &83.17/18.35 &91.14/56.48 &84.92/36.64 &72.77/15.44 &81.57/73.76 &82.93/35.27\\
    Garden\_04 &{\color{blue}\bf99.42}/{\color{blue}\bf86.61} &{\color{red}\bf99.18}/{\color{red}\bf86.35} &98.66/\bf83.25 &96.14/79.77 &89.50/17.66 &{\bf99.17}/59.93 &91.05/41.76 &81.05/19.53 &84.88/73.34 &87.98/32.39\\\hline
    Average     &{\color{blue}\bf98.59}/{\color{blue}\bf85.11} &{\bf97.67}/{\color{red}\bf83.74} &{\color{red}\bf97.87}/\bf81.68 &94.18/77.05 &87.11/18.12 &95.04/59.84 &89.42/44.61 &81.83/16.66 &83.09/75.32 &84.08/35.10\\\hline\hline
    \end{tabular}
\end{threeparttable}
\end{adjustbox}
}
\subfigure[Long-term LPR]{
\begin{adjustbox}{max width=\textwidth}
\begin{threeparttable}\label{tab: recall gardenLT}
\setlength{\tabcolsep}{4.0pt} 
    \begin{tabular}{l|| c c c c c c c c|| c c}
    \hline\hline
    {Sequence} &{UniLGL} &\makecell{UniLGL\\w/o Intensity} &\makecell{UniLGL\\w/o Spatial} &\makecell{UniLGL\\w/o Loc. Feat.} &\makecell{UniLGL\\w/o FM} &{BEVPlace++} &{Logg3D-Net} &{RING++} &\makecell{UniLGL\\Dis. Sup.} &\makecell{BEVPlace++\\Dis. Sup.}	\\\hline
    Garden\_LT\_01 &{\color{blue}\bf98.40}/{\color{blue}\bf87.89} &87.30/79.01 &{\bf87.68}/{\color{red}\bf82.31} &{75.35}/64.21 &51.42/47.16 &74.40/63.33 &48.96/39.26 &86.10/39.30 &{\color{red}\bf94.76}/\bf81.20 &76.18/35.97\\
    Garden\_LT\_02 &{\color{blue}\bf98.06}/{\color{blue}\bf87.85} &82.98/78.21 &{\bf86.54}/\bf79.57 &70.89/61.41 &53.21/43.25 &70.09/64.41 &53.23/36.09 &70.79/33.74 &{\color{red}\bf87.30}/{\color{red}\bf80.75} &73.33/33.17\\
    Garden\_LT\_03 &{\color{blue}\bf95.08}/{\color{blue}\bf83.31} &{\color{red}\bf94.44}/79.91 &{\bf90.19}/\bf80.34 &84.94/77.60 &65.94/51.37 &67.55/46.29 &53.70/43.79 &61.88/28.28 &89.20/{\color{red}\bf81.52} &78.73/47.56\\
    Garden\_LT\_04 &{\color{blue}\bf97.03}/{\color{blue}\bf86.69} &{\color{red}\bf94.60}/{\color{red}\bf86.45} &91.64/81.17 &85.36/73.77 &62.21/53.84 &69.24/55.00 &40.73/39.47 &73.55/31.12 &{\bf92.61}/\bf82.24 &84.80/61.57\\\hline
    Average     &{\color{blue}\bf97.14}/{\color{blue}\bf86.44} &{\bf89.85}/\bf80.90 &89.01/80.85 &79.14/69.25 &58.20/48.91 &70.32/57.26 &49.16/39.65 &73.08/33.11 &{\color{red}\bf90.97}/{\color{red}\bf81.43} &78.26/44.57\\
    \hline\hline
    \end{tabular}
\end{threeparttable}
\end{adjustbox}
}
\subfigure[LPR in large-scale urban driving scenarios]{
\begin{adjustbox}{max width=\textwidth}
\begin{threeparttable} \label{tab: recall urban driving}
    \setlength{\tabcolsep}{4.0pt} 
    \begin{tabular}{l|| c c c c c c c c|| c c}
    \hline\hline
    {Sequence} &{UniLGL} &\makecell{UniLGL\\w/o Intensity} &\makecell{UniLGL\\w/o Spatial} &\makecell{UniLGL\\w/o Loc. Feat.} &\makecell{UniLGL\\w/o FM} &{BEVPlace++} &{Logg3D-Net} &{RING++} &\makecell{UniLGL\\Dis. Sup.} &\makecell{BEVPlace++\\Dis. Sup.}	\\\hline
    Snail\_81R\_02 &95.62/\bf79.79 &95.03/76.55 &92.73/71.82 &{\bf96.15}/{\color{red}\bf88.76} &91.82/37.57 &92.08/59.95 &72.37/30.19 &36.09/6.03 &{\color{red}\bf98.67}/{\color{blue}\bf89.93} &{\color{blue}\bf99.50}/73.61\\
    Snail\_81R\_03 &{98.12}/{\bf88.19} &97.95/81.95 &{98.22}/87.82 &{\color{red}\bf98.53}/{\color{red}\bf89.36} &79.58/24.14 &97.60/64.07 &76.77/28.99 &35.56/8.48 &{\color{blue}\bf99.72}/{\color{blue}\bf91.26} &{\bf98.47}/61.83\\\hline
    Average     &96.87/{\bf83.99} &96.49/79.25 &95.48/79.82 &{\bf97.34}/{\color{red}\bf89.06} &85.70/30.86 &94.84/62.01 &74.57/29.59 &35.83/7.26 &{\color{blue}\bf99.20}/{\color{blue}\bf90.60} &{\color{red}\bf98.99}/67.72\\
    \hline\hline
    \end{tabular}
\end{threeparttable}
\end{adjustbox}
}
\begin{tablenotes}
    \footnotesize
    \parbox{0.97\textwidth}{
    \item The best result is highlighted in {\color{blue}\bf{Blue}}, the second-best result is highlighted in {\color{red}\bf{Red}}, and the third-best result is highlighted in {\color{black}\bf{Bold}}.
    \item For methods supervised by translation equivariance hypothesis (Hypothesis~\ref{Hyp: Translation Equivariance}), such as \textit{UniLGL Dis. Sup.} and \textit{BEVPlace++ Dis. Sup.}, point cloud pairs with distances below $5m$ are chosen as positive place recognition samples.
    For other methods listed in Section~\ref{Sec: Comparison baseline}, point cloud pairs with $\text{IoU}>0.25$ are chosen as positive place recognition samples.
    }
\end{tablenotes}
\vspace{-1em}
\end{table*}
\begin{figure}[!t]\centering
\includegraphics[width=0.9\linewidth]{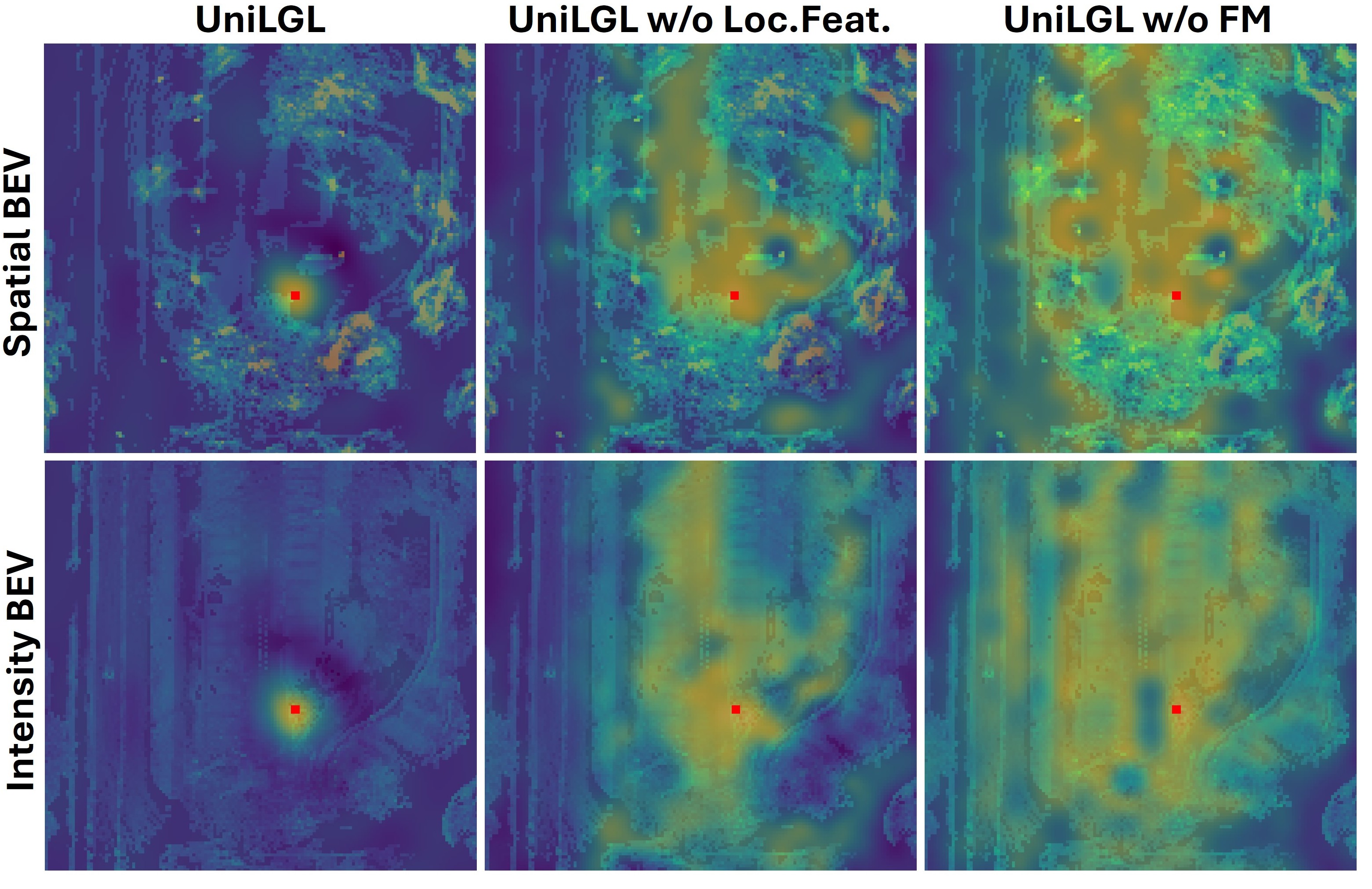}
\vspace{-1em}
\caption{Cosine similarity between local features and a reference pixel (highlighted in red).}\label{Fig: cosine}
\vspace{-1.5em}
\end{figure}
\begin{figure}[!t]\centering
\includegraphics[width=0.9\linewidth]{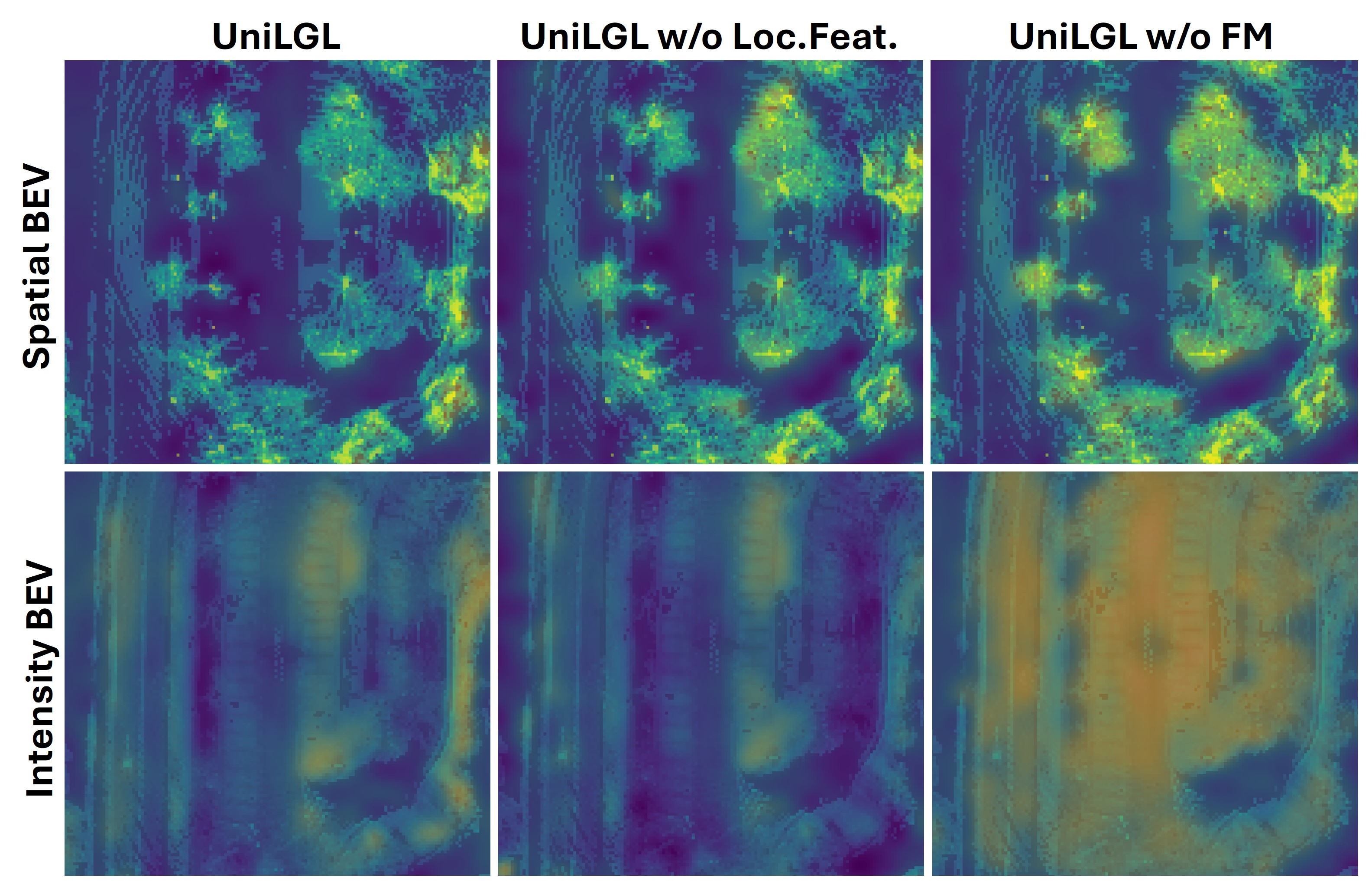}
\vspace{-1em}
\caption{Attention map visualization.}\label{Fig: attention}
\vspace{-1.5em}
\end{figure}

\begin{figure*}[!t]\centering
\includegraphics[width=\linewidth]{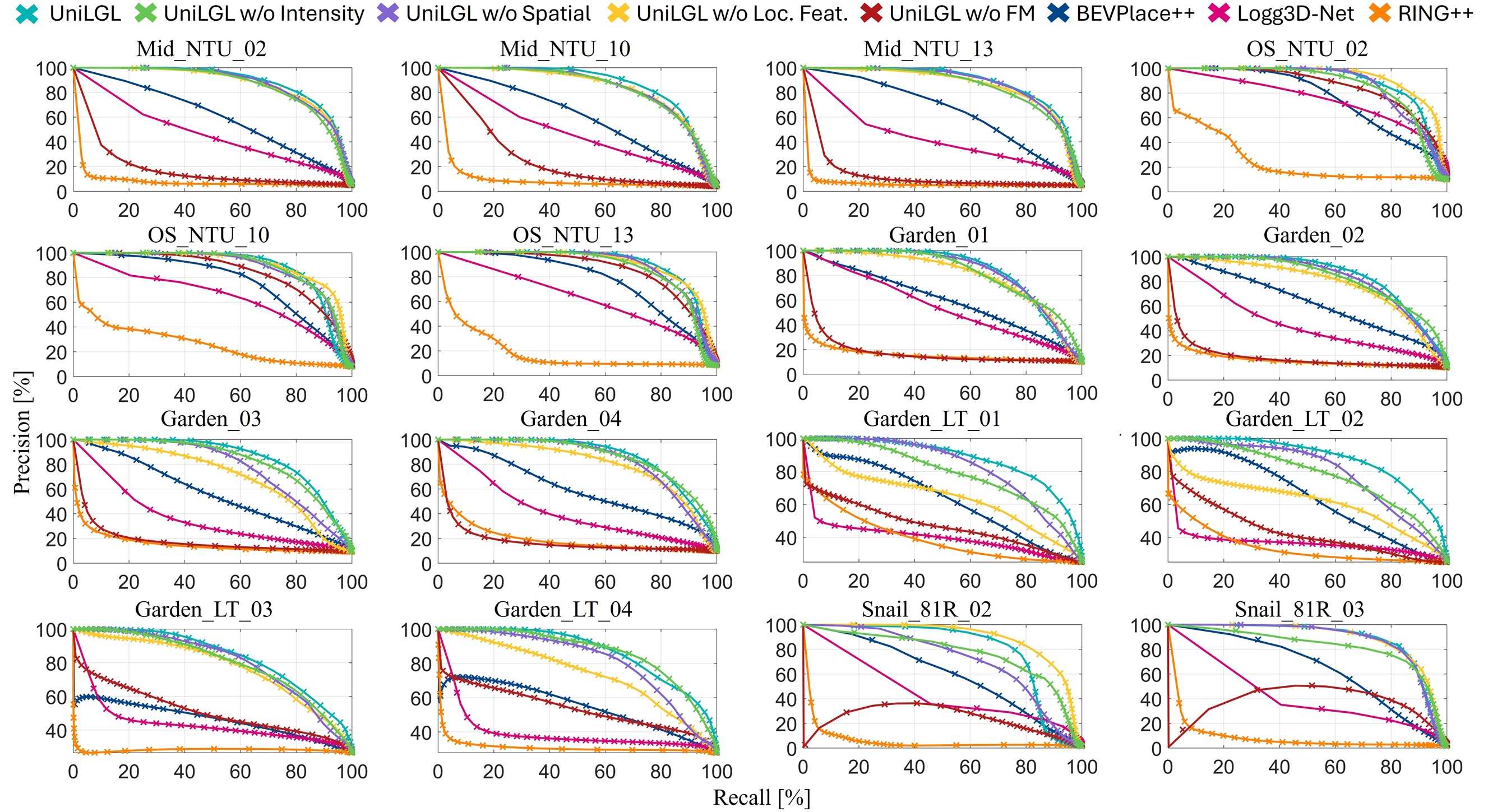}
\caption{Precision-recall curve of all benchmark method.}\label{Fig: PRCurve}
\end{figure*}
\subsubsection{Ablation Study}
The Top-1 recall and average precision of each method listed in \mbox{Section \ref{Sec: Comparison baseline}} is shown in \mbox{Table \ref{tab: recall}}.
\textbf{Image-level Sensor-type Uniformity:} To verify the sensor-type uniformity, Table~\ref{tab: recall-MCD} reports the LPR performance of UniLGL and its variants on FOV-limited and panoramic LiDARs.
From the results, the UniLGL achieves an average Top-1 recall of $98.11\%$ and $99.98\%$ when using FoV-limited LiDAR and panoramic LiDAR, respectively.
Thanks to the sensor-type uniformity supervision strategy based on the viewpoint invariance hypothesis proposed in Section~\ref{sec: Learning Sensor Type Uniformity}, UniLGL achieves consistent place recognition performance across heterogeneous LiDAR sensors.
In contrast, methods trained under the conventional translation equivariance hypothesis, such as \textit{UniLGL Dis. Sup.} and \textit{BEVPlace++ Dis. Sup.}, exhibit a $15$-$30\%$ drop in recall and a $20–45\%$ drop in precision on sequences collected using FoV-limited LiDARs compared to those using panoramic LiDARs.

\textbf{Patch-level Sensor-type Uniformity:}
To evaluate the impact of the local feature loss $\mathcal{L}_l$ on LPR performance, an ablation study is also conducted by comparing models with and without viewpoint-invariant local feature supervision.
Fig.~\ref{Fig: cosine} presents the cosine similarity visualizations corresponding to the example in Fig.~\ref{Fig: MotivationOfIntroduceIntensity}.
From the results, UniLGL exhibits more discriminative and better cross-image-consistent local features compared to \textit{UniLGL w/o Loc. Feat.}, demonstrating that introducing patch-level sensor-type uniformity preserves the locality of local features, thereby enhancing the LPR performance.
For LPR in large-scale urban driving scenarios, as shown in Table~\ref{tab: recall urban driving}, \textit{UniLGL w/o Loc. Feat.} outperforms UniLGL. 
This is attributed to the patch-level sensor-type uniformity which encourages UniLGL to focus on local details, including dynamic foreground objects, whose effect becomes particularly evident in the peak hour sequence (\textit{Snail\_81R\_02}).
For sequences with long-term time intervals, as shown in Table~\ref{tab: recall gardenLT}, UniLGL achieves an improvement of $18.00\%$ in Top-1 recall and $17.19\%$ in average precision when compare with \textit{UniLGL w/o Loc. Feat.}, demonstrating that the viewpoint-invariant supervision of patch-level local features effectively enhances the generalization capability of LPR.

\textbf{Spatial and Material Uniformity:}
To illustrate the effectiveness of spatial and material uniformity, we compare the place recognition performance of UniLGL with \textit{UniLGL w/o intensity} and \textit{UniLGL w/o spatial}.
As shown in Fig.~\ref{Fig: attention}, UniLGL focuses on complementary information over the spatial BEV and intensity BEV, respectively.
The objects with height information (e.g., trees) in the spatial BEV image and those containing distinctive material information (e.g., painted markers) in the intensity BEV image are effectively fused through the feature fusion network introduced in Section~\ref{Sec: Learning Spatial and Material Uniformity}, thereby enabling UniLGL to remain robust under geometric or intensity degradation scenarios shown in Fig.~\ref{Fig: MotivationOfIntroduceIntensity}.
This fusion significantly enhances the place recognition performance of UniLGL, enabling it to consistently outperform \textit{UniLGL w/o intensity} and \textit{UniLGL w/o spatial} on sequences collected by heterogeneous LiDARs across campus, repetitive garden, and large-scale urban driving scenarios.

\textbf{Foundation Model:} To facilitate understanding of the role of the foundation model, an ablation study is conducted by comparing the LPR performance of UniLGL with \textit{UniLGL w/o FM}.
From the results shown in Table~\ref{tab: recall}, UniLGL consistently outperforms \textit{UniLGL w/o FM} over all 16 sequences in both Top-1 recall and average precision.
As corroborated in \cite{ViT}, transformer-based architectures with a large number of parameters (e.g., ViT) underperform conventional CNNs (e.g., ResNets) when trained on small datasets, yet overtake them as the dataset scale grows.
Due to the insufficient amount of LiDAR training data, the attention in UniLGL w/o FM fails to converge to meaningful feature patterns, as illustrated in Fig.~\ref{Fig: attention}.
By initializing the network parameters with the foundation model DINO\cite{DINO}, which has been pretrained on a large-scale vision dataset, and subsequently fine-tuning it with a small amount of LiDAR data, UniLGL achieves high-performance LPR.
In several challenging scenarios, including repetitive scenes (Table~\ref{tab: recall garden}), long-term LPR (Table~\ref{tab: recall gardenLT}), and large-scale urban driving (Table~\ref{tab: recall urban driving}), UniLGL achieves $11.17$-$38.94\%$ and $37.53$-$66.99\%$ improvements in recall and precision, respectively, compared with \textit{UniLGL w/o FM}.

\begin{figure*}[!t]\centering
\includegraphics[width=0.9\linewidth]{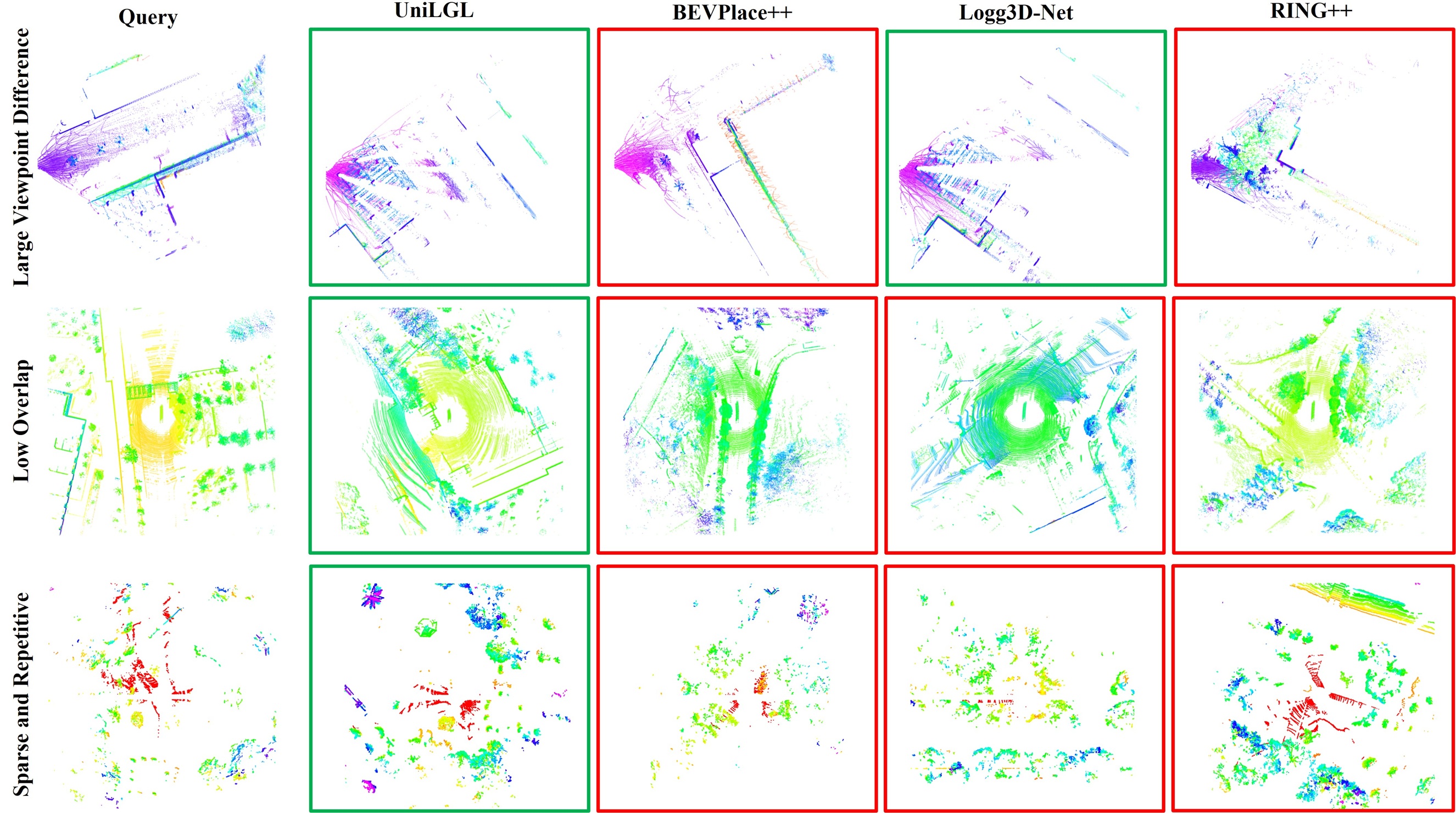}
\caption{Top-1 retrieved matches in challenging scenarios. The {\color{red}{$\square$}} represents the wrong retrieval result, and the {\color{green}{$\square$}} represents the correct retrieval result.}\label{Fig: PRinChanllengingScenarios}
\end{figure*}
\subsubsection{Benchmark with Baseline Methods}
As the results summarized in Table~\ref{tab: recall} and Fig.~\ref{Fig: PRCurve}, benefiting from the consideration of the uniformity introduced in Section~\ref{Sec: Learning Uniform Global Descriptor for LiDAR Place Recognition} and the introduction of VFM into the LPR network, UniLGL consistently outperforms SOTA learning-based\cite{BEVPlace++,LoGG3D-Net} and handcrafted\cite{ring++} LGL methods in all 16 sequences across campus, repetitive Garden, long-term retrieval, and large-scale urban driving scenarios.
BEVPlace++, Logg3D-Net, and RING++ exhibit their own distinct strengths.
RING++, a training-free LGL method which can easily generalize to new scenes without re-training or fine-tuning, shows better generalization ability over BEVPlace++ and LoGG3D-Net when processing long-term LPR, i.e., \textit{Garden\_LT} shown in Table~\ref{tab: recall gardenLT}.
However, the handcrafted descriptor lacks sufficient distinctiveness for large-scale scenes, leading RING++ to underperform the BEVPlace++ and Logg3D-Net on MCD and Snail datasets.
To perform LPR, BEVPlace++ and Logg3D-Net extract features from spatial BEV images and sparse point clouds using ResNet\cite{resnet} and Sparse U-Net\cite{U-net}, respectively.
However, due to the limited representational capacity of CNN-based architectures, both methods struggle in a variety of challenging scenarios, such as reversed loops, large viewpoint variations, loops with low-overlap regions, and sparse and repetitive environments, as illustrated in Fig.~\ref{Fig: PRinChanllengingScenarios}.
As discussed in the ablation study, the superior LPR performance of UniLGL can be attributed to multiple factors, including sensor-type uniformity, spatial and material uniformity, and the VFM-enabled highly representative LPR backbone.
As shown in Table~\ref{tab: recall-MCD}, UniLGL maintains comparable recall and precision when performing LPR with FoV-limited LiDAR compared to panoramic LiDAR, demonstrating its sensor-type uniformity.
In contrast, the SOTA LGL methods suffer a noticeable performance degradation, with a $7.86$–$24.61\%$ drop in recall and a $15.07$–$23.18\%$ decrease in precision.
For LPR performance in challenging scenarios summarized in Table~\ref{tab: recall garden}-\ref{tab: recall urban driving}, UniLGL achieves improvements of $2.03$–$61.04\%$ in recall and $21.98$–$76.73\%$ in precision, respectively, compared with SOTA LPR methods.

\subsection{Evaluation of Complete Global Localization} \label{sec: Accuracy Evaluations}
In this section, experiments are conducted to evaluate the accuracy of complete global localization, which estimates the global pose of query point cloud on $\text{SE}(3)$ against database references without prior knowledge of the initial pose.
For each query point cloud, UniLGL retrieves its Top-1 match from the database via place recognition and computes the global pose using the pose estimation algorithm derived in Section \ref{Sec: Relative Pose Estimation on Manifolds}.
Three metrics are introduced to evaluate the global localization performance of all methods listed in Section~\ref{Sec: Comparison baseline}.
\begin{itemize}
    \item \textit{Translation Error}: measures the Euclidean distance between estimated and ground truth translation vectors.
    \begin{equation}
    \small{
       e_t = {\left\| {{\bf{\hat t}} - {{\bf{t}}_{gt}}} \right\|_2} 
       }
    \end{equation}
    where ${\bf{\hat t}}\in \mathbb{R}^3$ and ${{\bf{t}}_{gt}}\in \mathbb{R}^3$ denote the estimated and ground truth translation vectors, respectively.
    \item \textit{Rotation Error}: measures the difference between the estimated and ground truth rotation.
    \begin{equation}
    \small{
        e_R = {\left\| {\text{Log}\left( {{{{\bf{\hat R}}}^\top}{{\bf{R}}_{gt}}} \right)} \right\|_2}
        }
    \end{equation}
    where $\text{Log}(\cdot)$ denotes the mapping from a rotation matrix to a rotation vector, and ${\bf{\hat R}}\in \text{SO}(3)$ and ${{\bf{R}}_{gt}}\in \text{SO}(3)$ denote the estimated and ground truth rotation matrix, respectively.
    \item \textit{Success Rate}: represents the fraction of scan pairs with translation error $e_t < 2m$ and rotation error $e_R < 5^\circ$. 
\end{itemize}
\begin{figure}[!t]\centering
\includegraphics[width=\linewidth]{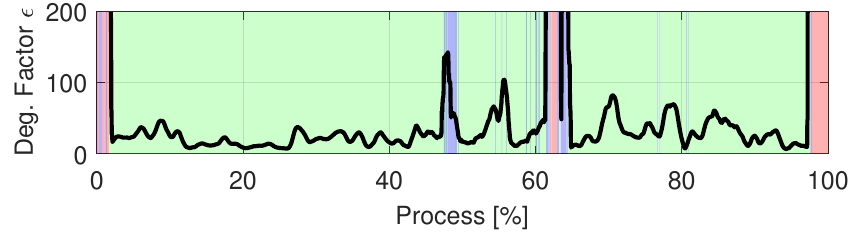}
\vspace{-1em}
\caption{Degeneration analysis of global pose estimation on the Mid\_NTU\_10.
The situations of LPR failure, LPR success but LGL failure, and LGL success are highlighted in \colorbox{red!20}{Red}, \colorbox{blue!20}{Blue}, and \colorbox{green!20}{Green}, respectively.
}\label{Fig: deg_plot}
\end{figure}
\subsubsection{Ablation Study}
\begin{table*}[!t] \centering
\centering
\caption{The Comparison of Success Rate ($\%$) /Rotation Error ($^\circ$) /Translation Error (Meters).}
\vspace{-1em}
\renewcommand{\arraystretch}{1.2}
\label{tab: GlobalLocalization}
\subfigure[LGL using heterogeneous LiDARs]{
\label{tab: GL MCD}
\begin{adjustbox}{max width=\textwidth}
\begin{threeparttable}
\vspace{-1em}
    \setlength{\tabcolsep}{3.2pt} 
    \begin{tabular}{l| l|| c c c c c c c c}
    \hline\hline
    {} &{Sequence} &{UniLGL} &\makecell{UniLGL\\w/o Intensity} &\makecell{UniLGL\\w/o Spatial} &\makecell{UniLGL\\w/o Loc. Feat.} &\makecell{UniLGL\\w/o FM} &{BEVPlace++} &{Logg3D-Net} &{RING++} \\\hline
    \multirow{4}{*}{\rotatebox{90}{FoV-limited}}
     &Mid\_NTU\_02 &{\color{blue}\bf88.38/\bf0.88}/{\color{red}\bf0.55} &{\color{red}\bf80.35}/1.21/0.61 &{\bf85.43}/{\color{red}\bf1.06}/0.71 &41.94/1.30/0.65 &35.54/1.68/0.71 &55.19/1.94/{\bf0.57}  &62.13/{\bf1.08}/{\color{blue}\bf0.38} &31.02/2.07/0.85\\
     &Mid\_NTU\_10 &{\color{red}\bf91.15}/{\color{blue}\bf0.59}/{\bf0.34} &{\bf90.64}/{\bf0.70}/0.41 &89.46/{\color{red}\bf0.63}/0.41 &82.95/0.84/0.41 &76.43/0.93/0.46 &88.70/1.10/{\color{red}\bf0.31} &{\color{blue}\bf93.75}/0.78/{\color{blue}\bf0.28} &57.08/1.45/0.78\\
     &Mid\_NTU\_13 &{\color{blue}\bf89.47}/{\color{red}\bf1.23}/0.53 &{\bf78.94}/1.41/0.64 &{\color{red}\bf87.80}/{\color{red}\bf1.23}/0.68 &44.57/{\color{blue}\bf1.18}/{\bf0.52} &37.92/{\bf1.29}/0.62 &55.31/1.77/{\color{red}\bf0.51} &58.15/1.33/{\color{blue}\bf0.41} &32.74/1.65/0.81\\\cline{2-10}
     &Average     &{\color{blue}\bf89.67}/{\color{blue}\bf0.90}/{\bf0.47} &{\bf83.31}/1.11/0.55 &{\color{red}\bf87.56}/{\color{red}\bf0.97}/0.60 &56.49/1.11/0.53 &49.96/1.30/0.60 &66.40/1.60/{\color{red}\bf0.46} &71.34/{\bf1.06}/{\color{blue}\bf0.36} &40.28/1.72/0.81\\\hline\hline
     
     \multirow{4}{*}{\rotatebox{90}{Panoramic}}
     &OS\_NTU\_02 &{\color{blue}\bf99.87}/{\color{blue}\bf0.39}/{\color{red}\bf0.29} &{\color{red}\bf99.74}/{\bf0.67}/{\color{blue}\bf0.27} &{\bf98.68}/{\color{red}\bf0.41}/0.43 &66.89/0.85/\bf0.42 &48.57/1.30/0.64 &96.15/1.97/0.42 &86.56/0.79/{\color{blue}\bf0.27} &75.51/1.44/0.50\\
     &OS\_NTU\_10 &{\color{blue}\bf99.69}/{\color{blue}\bf0.29}/\bf0.18 &{\color{blue}\bf99.69}/{\bf0.39}/0.18 &{\bf99.22}/{\color{red}\bf0.32}/0.25 &96.74/0.43/0.24 &82.33/0.68/0.37 &{\color{red}\bf99.38}/1.10/{\color{red}\bf0.17} &99.07/0.43/{\color{blue}\bf0.16} &75.57/1.44/0.50\\
     &OS\_NTU\_13 &{\color{blue}\bf100.0}/{\color{red}\bf0.96}/{\color{blue}\bf0.43} &{\color{red}\bf99.86}/{\bf1.08}/{\color{red}\bf0.49} &{\bf99.78}/{\color{blue}\bf0.95}/0.52 &65.94/1.40/0.64 &60.04/1.77/0.79 &96.33/1.86/0.52 &78.15/1.50/\bf0.50 &58.65/1.91/0.60\\\cline{2-10}
     &Average     &{\color{blue}\bf99.85}/{\color{blue}\bf0.55}/{\color{blue}\bf0.30} &{\color{red}\bf99.76}/{\bf0.71}/{\color{red}\bf0.31} &{\bf99.23}/{\color{red}\bf0.56}/0.40 &76.52/0.89/0.43 &63.65/1.25/0.60 &97.29/1.64/\bf0.37 &87.93/{0.91}/{\color{red}\bf0.31} &69.91/1.60/0.53\\
    \hline\hline
    \end{tabular}
\end{threeparttable}
\end{adjustbox}
}
\subfigure[LGL in repetitive scenarios]{
\begin{adjustbox}{max width=\textwidth}
\begin{threeparttable}\label{tab: GL garden}
\vspace{-1em}
    \setlength{\tabcolsep}{4.0pt} 
    \begin{tabular}{l|| c c c c c c c c}
    \hline\hline
    {Sequence} &{UniLGL} &\makecell{UniLGL\\w/o Intensity} &\makecell{UniLGL\\w/o Spatial} &\makecell{UniLGL\\w/o Loc. Feat.} &\makecell{UniLGL\\w/o FM} &{BEVPlace++} &{Logg3D-Net} &{RING++} 	\\\hline
    Garden\_01 &{\bf91.39}/{\color{blue}\bf0.53}/{\color{red}\bf0.26} &90.49/{0.66}/0.34 &88.96/{\color{blue}\bf0.53}/\bf0.31 &83.13/{\color{red}\bf0.59}/{\color{red}\bf0.26} &81.90/0.77/0.35 &{\color{red}\bf91.59}/{0.68}/{\color{blue}\bf0.18} &{\color{blue}\bf91.77}/{\bf0.63}/{\bf0.31} &83.33/1.12/0.67\\
    Garden\_02 &{\color{red}\bf90.88}/{\color{blue}\bf0.59}/{\color{red}\bf0.31} &{\bf90.36}/{\bf0.67}/0.34 &{90.20}/{\color{red}\bf0.64}/0.34 &76.77/0.77/\bf0.33 &76.34/0.87/0.39 &89.05/0.78/{\color{blue}\bf0.20} &{\color{blue}\bf91.47}/{0.76}/{0.34} &83.42/1.22/0.68\\
    Garden\_03 &{\color{blue}\bf90.76}/{\color{blue}\bf0.58}/{\color{red}\bf0.28} &86.54/{\bf0.70}/0.34 &{\color{red}\bf87.89}/{\color{red}\bf0.64}/0.36 &73.08/0.78/0.36 &72.95/0.94/0.43 &85.35/0.79/{\color{blue}\bf0.20} &{\bf87.58}/{0.71}/{\bf0.33} &67.12/1.24/0.73\\
    Garden\_04 &{\color{blue}\bf95.92}/{\color{red}\bf0.60}/{\color{red}\bf0.32} &{\bf95.14}/{\bf0.79}/0.36 &88.17/{\color{blue}\bf0.59}/\bf0.33 &79.28/0.73/0.29 &79.10/0.86/0.38 &90.00/{0.77}/{\color{blue}\bf0.20} &{\color{red}\bf95.79}/{0.71}/{\bf0.33} &74.23/1.23/0.74\\\hline
    Average     &{\color{blue}\bf92.24}/{\color{blue}\bf0.58}/{\color{red}\bf0.29} &{\bf90.63}/{0.71}/0.35 &88.81/{\color{red}\bf0.60}/0.34 &78.07/0.72/\bf0.31 &77.57/0.86/0.39 &89.00/0.76/{\color{blue}\bf0.20} &{\color{red}\bf91.65}/{\bf0.70}/{0.33} &77.03/1.20/0.71\\\hline\hline
    \end{tabular}
\end{threeparttable}
\end{adjustbox}
}
\subfigure[Long-term LGL]{
\begin{adjustbox}{max width=\textwidth}
\begin{threeparttable}\label{tab: GL gardenLT}
\vspace{-1em}
\setlength{\tabcolsep}{4.0pt} 
    \begin{tabular}{l|| c c c c c c c c}
    \hline\hline
    {Sequence} &{UniLGL} &\makecell{UniLGL\\w/o Intensity} &\makecell{UniLGL\\w/o Spatial} &\makecell{UniLGL\\w/o Loc. Feat.} &\makecell{UniLGL\\w/o FM} &{BEVPlace++} &{Logg3D-Net} &{RING++}\\\hline
    Garden\_LT\_01 &{\color{blue}\bf80.55}/{\color{blue}\bf1.21}/{0.58} &{\bf64.24}/{\color{red}\bf1.36}/{\bf0.57} &54.45/{\bf1.37}/0.77 &2.64/2.20/0.79 &1.10/2.20/0.76 &13.17/1.99/0.77 &19.75/1.80/{\color{red}\bf0.52} &{\color{red}\bf77.23}/1.65/{\color{blue}\bf0.48}\\
    Garden\_LT\_02 &{\color{blue}\bf74.97}/{\bf1.33}/0.73 &{\color{red}\bf61.54}/{\color{blue}\bf1.28}/{\color{red}\bf0.55} &50.15/1.43/0.79 &1.50/2.65/0.70 &1.56/2.72/0.73 &17.46/{\color{red}\bf1.29}/0.77 &20.20/1.91/{\bf0.56} &{\bf61.35}/1.62/{\color{blue}\bf0.47}\\
    Garden\_LT\_03 &{\color{blue}\bf75.81}/{\color{red}\bf1.30}/{\color{red}\bf0.56} &{\color{red}\bf71.79}/{\bf1.50}/\bf0.58 &{\bf55.36}/1.68/0.72 &6.22/2.31/0.75 &1.99/2.67/0.98 &18.07/{\color{blue}\bf1.23}/0.75 &26.40/1.81/{\color{red}\bf0.56} &53.21/1.58/{\color{blue}\bf0.44}\\
    Garden\_LT\_04 &{\color{blue}\bf80.59}/{\color{blue}\bf1.27}/{0.61} &{\color{red}\bf76.68}/{\bf1.39}/\bf0.56 &56.83/1.81/0.75 &6.47/1.79/0.92 &1.95/2.30/0.89 &18.25/{\color{red}\bf1.30}/0.79 &21.32/{\bf1.57}/{\color{red}\bf0.55} &{\bf61.61}/1.60/{\color{blue}\bf0.48}\\\hline
    Average     &{\color{blue}\bf77.98}/{\color{blue}\bf1.28}/0.62 &{\color{red}\bf68.56}/{\color{red}\bf1.38}/{\bf0.57} &54.20/1.57/0.76 &4.18/2.09/0.77 &1.61/2.46/0.83 &16.74/{\bf1.45}/0.77 &21.92/1.77/{\color{red}\bf0.55} &{\bf63.35}/1.61/{\color{blue}\bf0.47}\\
    \hline\hline
    \end{tabular}
\end{threeparttable}
\end{adjustbox}
}
\subfigure[LGL in large-scale urban driving scenarios]{
\begin{adjustbox}{max width=\textwidth}
\begin{threeparttable} \label{tab: GL urban driving}
\vspace{-1em}
    \setlength{\tabcolsep}{4.0pt} 
    \begin{tabular}{l|| c c c c c c c c}
    \hline\hline
    {Sequence} &{UniLGL} &\makecell{UniLGL\\w/o Intensity} &\makecell{UniLGL\\w/o Spatial} &\makecell{UniLGL\\w/o Loc. Feat.} &\makecell{UniLGL\\w/o FM} &{BEVPlace++} &{Logg3D-Net} &{RING++} \\\hline
    Snail\_81R\_02 &{\color{blue}\bf79.20}/{\color{blue}\bf0.75}/1.07 &{\color{red}\bf79.17}/{\color{red}\bf0.82}/1.07 &63.96/{\bf1.14}/\bf1.00 &15.32/1.89/{\color{blue}\bf0.89} &27.19/1.72/1.09 &{\bf73.06}/1.27/1.06 &31.51/2.07/{\color{red}\bf0.90} &29.88/1.40/{\bf0.94}\\
    Snail\_81R\_03 &{\color{blue}\bf82.83}/{\color{blue}\bf0.65}/{\color{blue}\bf0.50} &{\color{red}\bf82.52}/{\color{red}\bf0.79}/{\color{red}\bf0.56} &{\bf82.31}/{\bf0.81}/\bf0.60 &13.85/2.11/1.18 &33.55/1.75/0.87 &80.39/{1.31}/0.65 &40.57/2.03/0.64 &31.46/1.52/0.61\\\hline
    Average     &{\color{blue}\bf81.02}/{\color{blue}\bf0.70}/{\bf0.79} &{\color{red}\bf80.85}/{\color{red}\bf0.81}/0.82 &73.14/{\bf0.98}/{0.80} &14.59/2.00/1.04 &30.37/1.74/0.98 &{\bf76.73}/1.29/0.86 &36.04/2.05/{\color{blue}\bf0.77} &30.67/1.46/{\color{red}\bf0.78}\\
    \hline\hline
    \end{tabular}
\end{threeparttable}
\end{adjustbox}
}
\begin{tablenotes}
    \footnotesize
    \item The best result is highlighted in {\color{blue}\bf{Blue}}, the second-best result is highlighted in {\color{red}\bf{Red}}, and the third-best result is highlighted in {\color{black}\bf{Bold}}.
\end{tablenotes}
\end{table*}

The success rate and pose estimation precision of each method listed in \mbox{Section \ref{Sec: Comparison baseline}} is shown in \mbox{Table \ref{tab: GlobalLocalization}}.

\textbf{Patch-level Sensor-type Uniformity:}
To illustrate the effectiveness of introducing sensor-type uniformity at the local feature level, an ablation study is conducted by comparing models with and without viewpoint-invariant local feature supervision.
From the results shown in Table~\ref{tab: GL MCD}, the proposed UniLGL achieves an average success rate improvement of $33.18\%$ and $23.33\%$ over \textit{UniLGL w/o Loc. Feat.} when using FoV-limited LiDAR and panoramic LiDAR, respectively.
For UniLGL, when using a FoV-limited LiDAR, a $10.18\%$ drop is observed in the LGL success rate compared to using the panoramic LiDAR.
This primarily attribute to the reduced amount of observed information in FoV-limited LiDARs, which leads to registration degeneration.
As demonstrated in Fig.~\ref{Fig: deg_plot}, the degeneration factor is defined as the ratio between the largest and smallest eigenvalues obtained from SVD used to solve the outlier-free global pose estimation problem (\ref{Eq: GNC-2}).
Despite the LGL performance drop on FoV-limited LiDAR, UniLGL still consistently outperforms \textit{UniLGL w/o Loc. Feat.} across all 16 sequences, achieving an average improvement of $14.17\%$–$73.80\%$ in the global localization success rate.
This is attributed to the local features extracted by UniLGL exhibiting higher discriminability and better cross-image consistency compared to those of \textit{UniLGL w/o Loc. Feat.}, as illustrated in Fig.~\ref{Fig: cosine}.

\textbf{Spatial and Material Uniformity:}
To validate the effectiveness of introducing spatial and material uniformity, we compare UniLGL with two ablation variants, \textit{UniLGL w/o Intensity} and \textit{UniLGL w/o Spatial}.
As the results shown in Table~\ref{tab: GlobalLocalization}, UniLGL outperform \textit{UniLGL w/o Intensity} and \textit{UniLGL w/o Spatial} consistently in global localization success rate.
For challenging sequences with long time intervals, such as \textit{Garden\_LT\_01} to \textit{Garden\_LT\_04} illustrated in Table~\ref{tab: GL gardenLT}, UniLGL achieves more robust global localization performance, with average improvements of $9.42\%$ and $23.78\%$ in success rate over \textit{UniLGL w/o Intensity} and \textit{UniLGL w/o Spatial}, respectively.
A further observation is that \textit{UniLGL w/o Intensity} and \textit{UniLGL w/o Spatial} suffer noticeable declines in global localization success rate on the FoV-limited LiDAR sequences (Mid\_NTU\_XX) and the urban peak-hour scenario (Snail\_81R\_02), respectively.
By jointly leveraging the complementary cues from the spatial BEV image and the intensity BEV image, UniLGL achieves success rate improvements of $6.36\%$ under the limited observations of FoV-limited LiDARs and $15.24\%$ in the presence of dynamic objects, such as moving vehicles and pedestrians, during peak-hour urban driving, respectively.

\textbf{Foundation Model:}
To facilitate understanding of the role of the foundation model, an ablation study is conducted by comparing the LGL performance of UniLGL with \textit{UniLGL w/o FM}.
From the results shown in Table~\ref{tab: GlobalLocalization}, UniLGL consistently outperforms \textit{UniLGL w/o FM} over all the 16 sequences in both LGL success rate and translation error, rotation error.
Fig.~\ref{Fig: cosine} shows the cosine similarity maps between local features and a reference pixel (highlighted in red), which demonstrates that the introduction of the foundation model brings UniLGL better local feature discriminability and locality.
This enables UniLGL to achieve $14.67$-$76.37\%$ LGL success rate improvement in challenging scenarios across campus, repetitive garden, long-term retrieval, and large-scale urban driving.
\begin{figure*}[!t]\centering
\includegraphics[width=\linewidth]{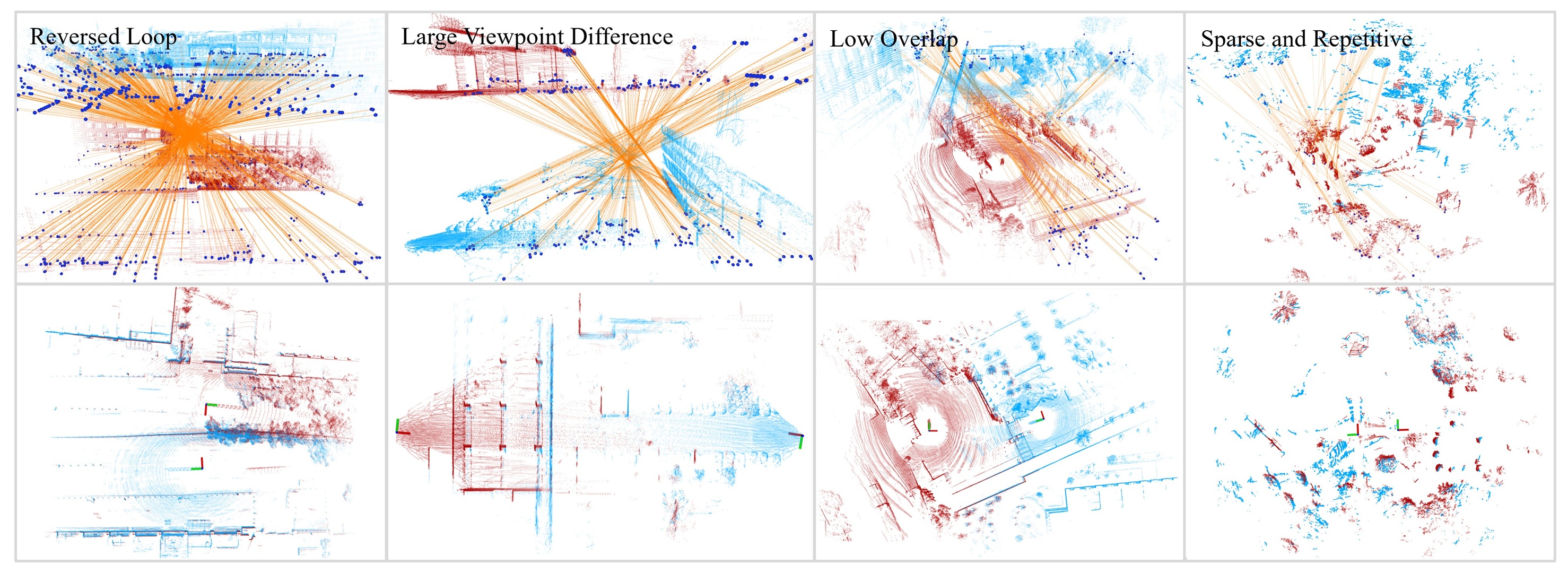}
\vspace{-1em}
\caption{Global localization in challenging scenarios. The top row shows the point clouds before alignment, where the orange lines indicate point correspondences obtained through local feature matching.
The bottom row shows the point clouds aligned by UniLGL without additional registration.
}\label{Fig: PointCloudAlignment}
\vspace{-1em}
\end{figure*}
\begin{figure}
\centering	
\subfigure[Translation Error $e_t$]{
    \includegraphics[width=\linewidth]{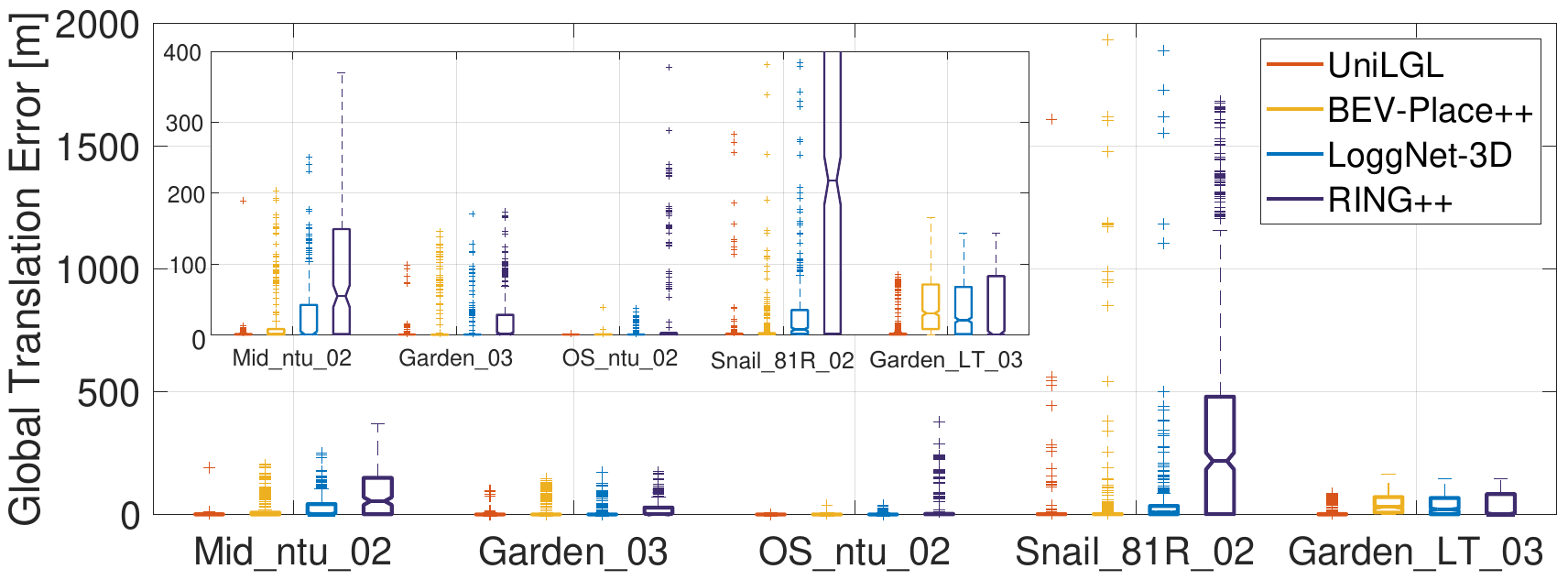}
	\vspace{-1em}
}
\subfigure[Rotation Error $e_R$]{
    \includegraphics[width=\linewidth]{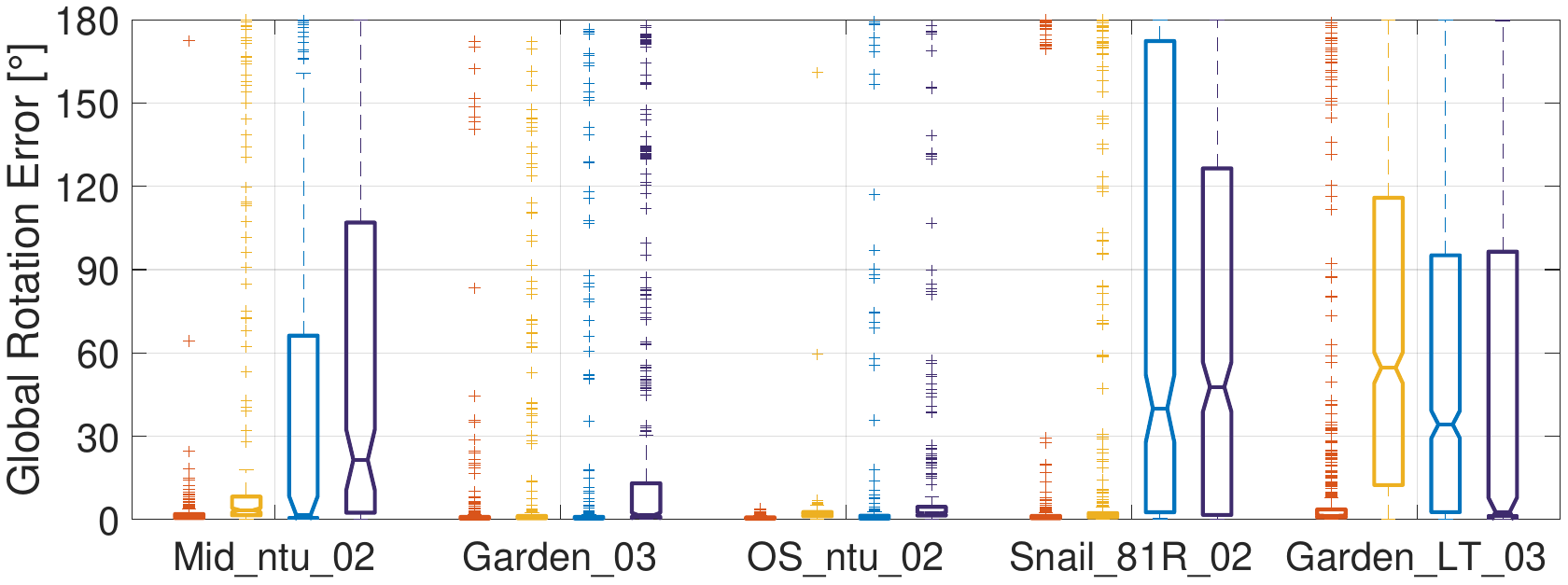}
}
\vspace{-1em}
\caption{Global localization error, including both successful and failed localization cases.}
\label{Fig: GlobalLocalizationError}
\vspace{-1.5em}
\end{figure}
\subsubsection{Benchmark with Baseline Methods}
As summarized in Table~\ref{tab: GL MCD}, UniLGL enables high-precision global localization without prior knowledge of the initial pose, improving the global localization success rate by $2.56\%$–$61.24\%$ and $14.63\%$–$50.35\%$ compared to SOTA BEV-based LGL methods, BEVPlace++ and RING++, on sequences collected by FoV-limited LiDAR and panoramic LiDAR, respectively.
It is worth noting that Logg3D-Net outperforms the proposed UniLGL on certain sequences, as Logg3D-Net is enhanced by its follow-up work SpectralGV, which re-ranks the Top-20 retrieval candidates based on local feature consistency.
UniLGL incorporates local feature consistency directly within the network, allowing it to achieve outstanding global localization performance using only the Top-1 retrieval.
As shown in Table~\ref{tab: GL gardenLT}, for long-term sequences such as \textit{Garden\_LT\_01} to \textit{Garden\_LT\_04}, the hand-crafted LGL method RING++ achieves significantly better global localization performance compared to learning-based methods BEVPlace++ and Logg3D-Net.
This is mainly attributed to the inherent generalization ability of handcrafted descriptors and features.
In contrast, the proposed UniLGL guided the VFM-enhanced feature fusion network introduced in Section~\ref{Sec: Learning Spatial and Material Uniformity} using the viewpoint invariance hypothesis (Hypothesis~\ref{Hyp: Viewpoint Invariance}), which encodes spatial, material, and sensor-type uniformity into the LGL process.
These designs enable UniLGL to achieve superior generalization performance compared to existing SOTA methods.
Specifically, on the \textit{Garden\_LT\_01} to \textit{Garden\_LT\_04}, UniLGL improves the global localization success rate by over $56.06\%$ compared to BEVPlace++ and Logg3D-Net, and by more than $14.63\%$ even compared to RING++.
Due to its higher global localization success rate compared to other methods, UniLGL considers a broader range of challenging scenarios, such as reverse loops, global localization under large rotations and translations, low-overlap regions, and sparse and repetitive long-term environments shown in Fig.~\ref{Fig: PointCloudAlignment}, which are excluded by other methods due to localization failure.
Despite considering these more difficult cases, UniLGL still maintains comparable pose estimation accuracy.
In Table \ref{tab: GlobalLocalization}, only the localization error for successful global localization is tabulated.
For a fair comparison, the localization error for positive place recognition, including both successful and failed global localization cases, is illustrated in Fig. \ref{Fig: GlobalLocalizationError}.
Compared to SOTA LGL methods, UniLGL produces fewer global localization outliers and effectively aligns pairs of point clouds in various challenging scenarios without additional registration.

\subsection{Evaluation of Running Time}
The average processing consumption of the proposed UniLGL and each benchmark global localization method when using heterogeneous LiDARs is summarized in \mbox{Table~\ref{tab: realtime}}.
UniLGL formulates the LPR problem as an image retrieval task, enabling efficient LPR through KNN search over the global descriptor space. This approach achieves consistent place recognition time consumption across varying-scale scenarios and heterogeneous LiDAR sensors.
From the results shown in Table~\ref{tab: realtime}, UniLGL outperforms SOTA LGL methods in LPR time consumption across all 16 sequences.
Compared to SOTA learning-based LGL methods, BEVPlace++ and Logg3D-Net, UniLGL adopts a Transformer as the backbone, which can be easily accelerated using Transformers acceleration toolkits (e.g., xFormers\cite{xFormers2022}) and extracts more distinctiveness global descriptors, leading to a $55\%$-$90\%$ reduction in LPR time consumption.
Compared with RING++, UniLGL replaces exhaustive search with global descriptor matching, which avoids the curse of dimensionality and improves computational efficiency by over $98\%$.

The time consumption of global pose estimation is decomposed into two parts, local feature matching (noted as `Match' in Table~\ref{tab: realtime}) and estimation (noted as `Solve' in Table~\ref{tab: realtime}).
RING++ achieves the best real-time performance in global pose estimation because its feature matching and rotation estimation are coupled with LPR, allowing the global pose to be estimated using a closed-form position-only solution.
However, this coupling also leads to a curse of dimensionality in LPR.
BEVPlace++ demonstrates more stable time consumption on local feature matching compared to UniLGL.
This is attributed to BEVPlace++ performing 3-DoF global localization using FAST keypoints\cite{rosten2006machine} extracted from BEV images.
In contrast, the proposed method is a fully 6-DoF global localization approach, which maps all local feature correspondences from BEV images to the point cloud.
Compared to the 6-DoF global localization method, Logg3D-Net, which performs re-ranking of the Top-20 LPR candidates using SpectralGV\cite{vidanapathirana2023sgv}, the proposed UniLGL conducts local feature matching only on the Top-1 correspondence, resulting in over $93\%$ reduction in computational cost.
For estimation, UniLGL employs the GNC-TLS optimization introduced in Section~\ref{Sec: Robust Global Pose Recovery}, which avoids the large number of iterations required by the RANSAC registration adopted in BEVPlace++ and LoGG3D-Net.
As a result, the GNC-TLS enables UniLGL to achieve over $98\%$ improvement in solving efficiency when compared with the 6-DoF RANSAC estimation used in LoGG3D-Net, and even a $24.65\%$–$80.96\%$ improvement when compared with the 3-DoF RANSAC estimation used in BEVPlace++.
UniLGL achieves an average total time consumption of $38.86 ms$ and $104.16 ms$ when using FoV-limited LiDAR and panoramic LiDAR, respectively.
As shown in Tables~\ref{tab: recall}-\ref{tab: realtime}, UniLGL delivers high-performance LPR and fully 6-DoF global localization while maintaining comparable real-time capability.

\begin{table*}[!t] \centering
\centering
\caption{Average Processing Consumption (Milliseconds).}
\setlength{\tabcolsep}{4.0pt} 
\renewcommand{\arraystretch}{1.2}
\label{tab: realtime}
\begin{adjustbox}{max width=\textwidth}
\begin{threeparttable}
    \begin{tabular}{l| l|| c c c c c c c c c c c c c c c c}
    \hline\hline
    \multirow{2}{*}{} &\multirow{2}{*}{Sequence} &\multicolumn{4}{c}{UniLGL} &\multicolumn{4}{c}{BEVPlace++} &\multicolumn{4}{c}{Logg3D-Net} &\multicolumn{4}{c}{RING++}\\
    \cline{3-18}
    &{} &LPR &Match &Solve &Total &LPR &Match &Solve &Total &LPR &Match &Solve &Total &LPR &Match &Solve &Total\\\hline
    \multirow{4}{*}{\rotatebox{90}{FoV-limited}}
     &Mid\_NTU\_02 &{\color{black}\bf6.96} &23.06 &6.71 &\bf36.73 &15.27 &13.74 &40.75 &69.76 &38.97 &1068.34 &1748.24 &2855.50 &351.26 &\bf0 &{\color{black}\bf1.38} &352.64\\
     &Mid\_NTU\_10 &{\color{black}\bf6.87} &25.88 &8.90 &\bf41.65 &15.27 &13.79 &40.10 &69.17 &39.70 &1062.64 &1305.98 &2408.32 &372.46 &\bf0 &{\color{black}\bf1.25} &373.71\\
     &Mid\_NTU\_13 &{\color{black}\bf6.98} &23.88 &7.35 &\bf38.21 &15.25 &13.77 &39.65 &68.67 &39.58 &1069.36 &1737.18 &2846.11 &355.01 &\bf0 &{\color{black}\bf1.25} &356.26\\
     \cline{2-18}
     &Average    &{\color{black}\bf6.94} &24.27 &7.65 &\bf38.86 &15.26 &13.77 &40.17 &69.20 &39.42 &1066.78 &1597.13 &2703.31 &359.33 &\bf0 &{\color{black}\bf1.29} &360.87\\\hline\hline
     \multirow{4}{*}{\rotatebox{90}{Panoramic}}
     &OS\_NTU\_02 &{\color{black}\bf7.33} &61.59 &26.45 &95.37 &15.30 &14.68 &41.55 &\bf71.54 &40.63 &1071.78 &1923.43 &3035.84 &838.60 &\bf0 &{\color{black}\bf1.24} &839.84\\
     &OS\_NTU\_10 &{\color{black}\bf7.35} &64.92 &29.31 &101.58 &15.34 &14.69 &42.40 &\bf72.44 &40.02 &1071.96 &1247.27 &2359.24 &901.14 &\bf0 &{\color{black}\bf1.22} &902.37\\
     &OS\_NTU\_13 &{\color{black}\bf7.35} &72.85 &35.34 &115.54 &15.27 &14.90 &42.71 &\bf72.88 &41.27 &1071.86 &2155.37 &3268.50 &966.50 &\bf0 &{\color{black}\bf1.24} &967.74\\
     \cline{2-18}
     &Average     &{\color{black}\bf7.34} &66.45 &30.37 &104.16 &23.37 &14.22 &40.31 &\bf79.18 &40.95 &972.06 &1812.01 &2825.02 &1030.55 &\bf0 &{\color{black}\bf1.47} &1032.04\\
    \hline\hline
    \end{tabular}
\begin{tablenotes}
    \footnotesize
    \item The best result is highlighted in {\color{black}\bf{Bold}}.
    Logg3D-Net\cite{LoGG3D-Net} is integrated with SpectralGV\cite{vidanapathirana2023sgv} to achieve 6-DoF metric localization.
\end{tablenotes}
\end{threeparttable}
\end{adjustbox}
\vspace{-1em}
\end{table*}

\section{Real-world Applications}
\begin{figure}
\centering	
\subfigure[Full-size Truck]{
    \includegraphics[width=0.43\linewidth]{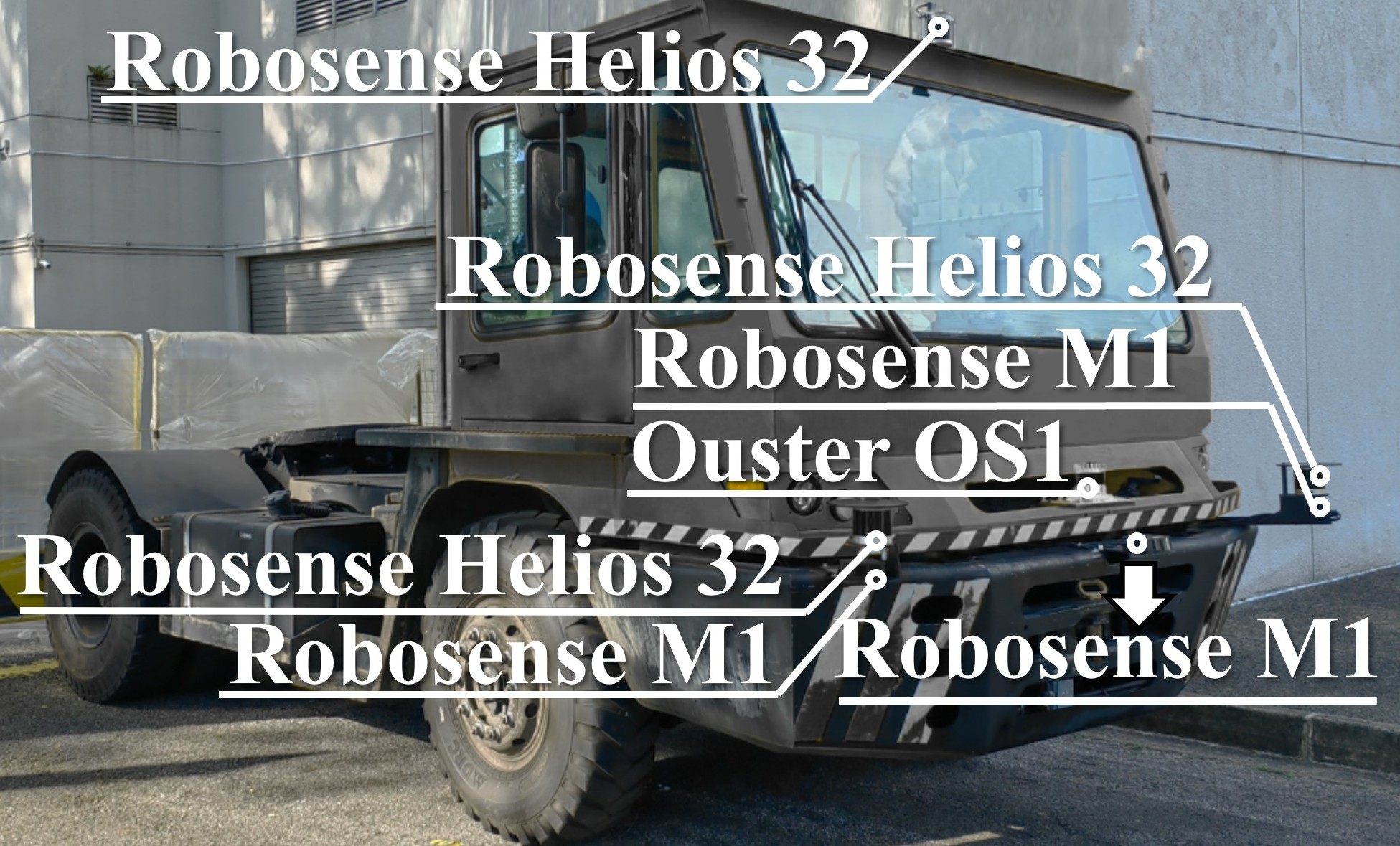}
    \label{Fig: Truck}		
}
\subfigure[MAV]{
    \includegraphics[width=0.5\linewidth]{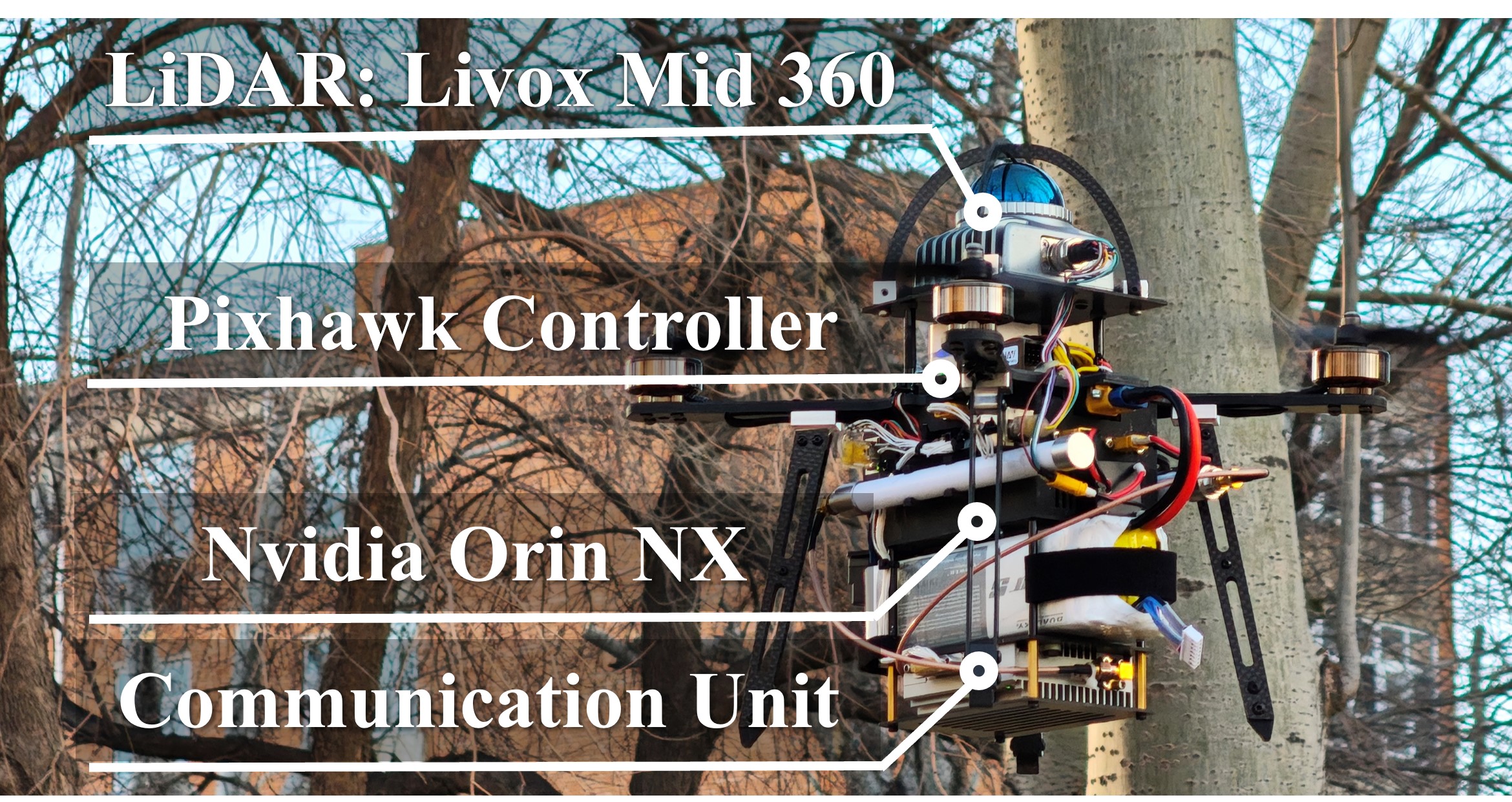} 
    \label{Fig: MAV}
}
\vspace{-0.5em}
\caption{Platforms used in real-world applications.}
\label{Fig: Platforms}
\vspace{-1.0em}
\end{figure}
\subsection{Application 1: LiDAR-only Localization for Self-driving Truck in Large-scale Port Scenario}
\begin{figure*}
\centering	
\subfigure[]{
    \includegraphics[width=0.4\linewidth]{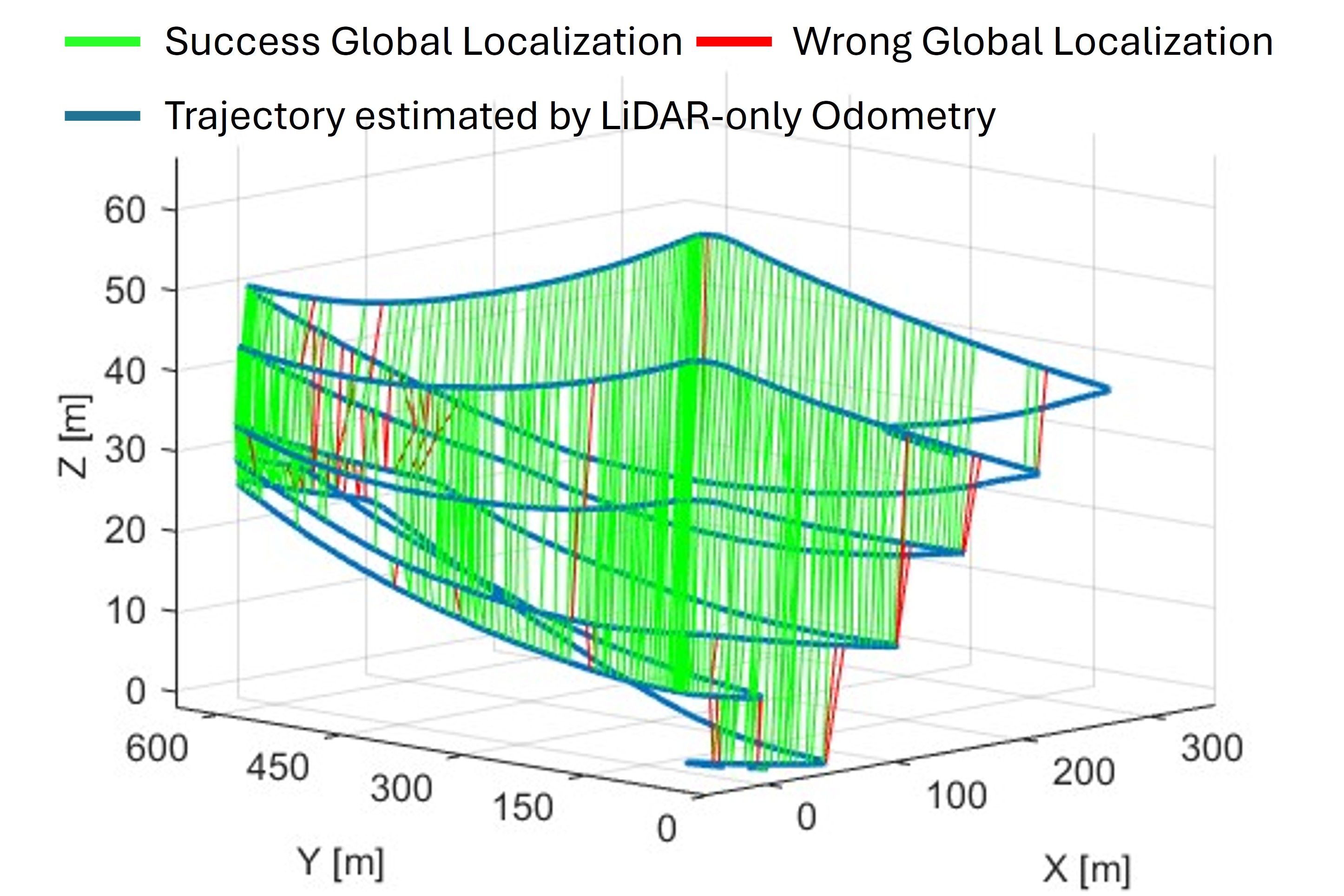}
    \label{Fig: PSATraj}
}
\subfigure[]{
    \includegraphics[width=0.55\linewidth]{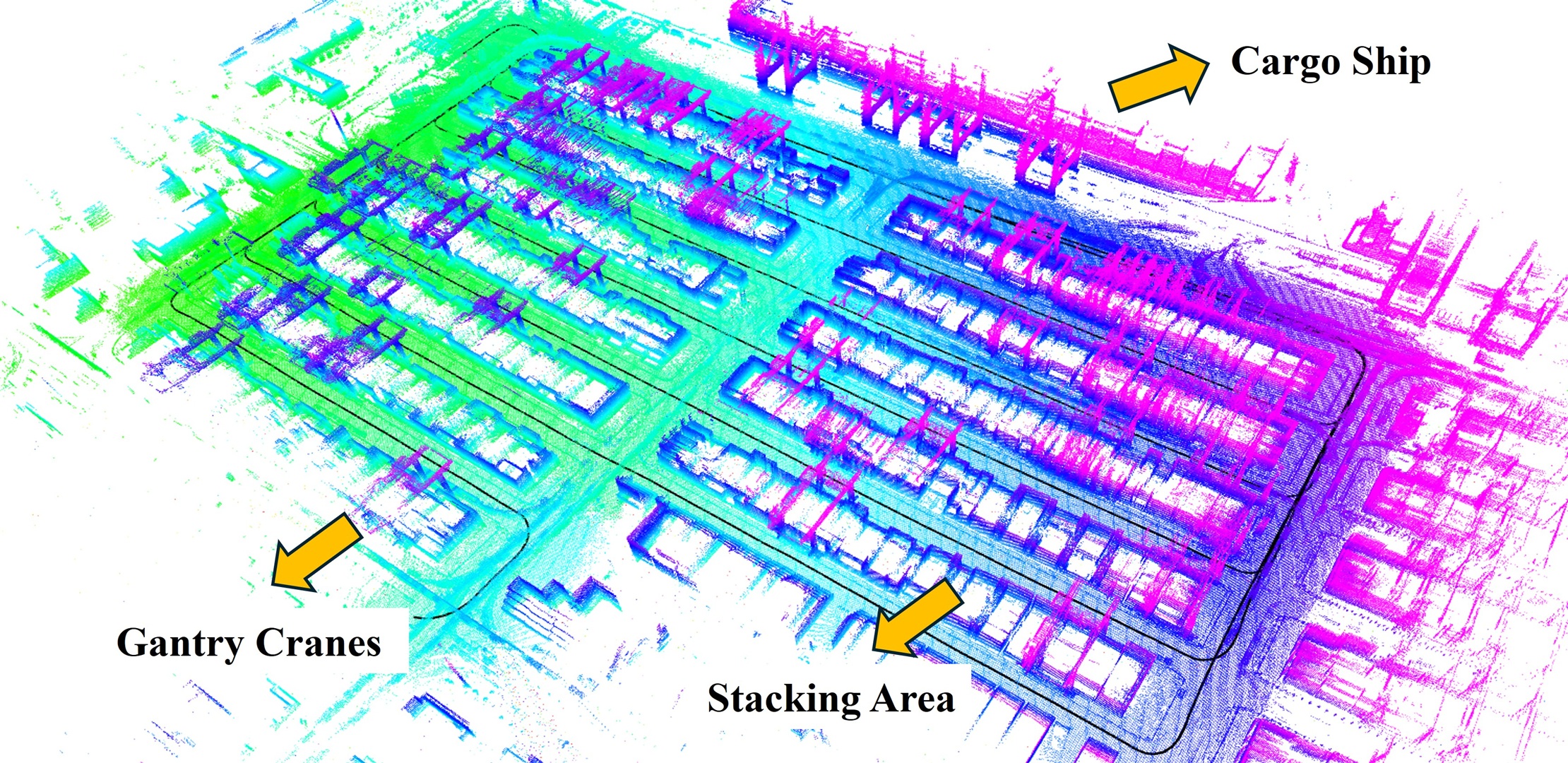} 
    \label{Fig: PSAMap}
}
\caption{Long-term localization and mapping in highly repetitive large-scale port scenario. (a) Global localization results of UniLGL. (b) The reconstruction result. UniLGL is adopted to detect loops and provide relative pose estimations, thereby eliminating the drift accumulated by LiDAR-only odometry.}
\vspace{-1em}
\end{figure*}
To attest to its practicality, the proposed UniLGL is integrated with a multi-LiDAR odometry (CTE-MLO\cite{cte-mlo}) to deliver high-precision and comprehensive localization and mapping for a full-size autonomous driving truck.
Driving tests spanning over $10 km$ are conducted across three highly similar yet distinct areas within a large-scale seaport.
As shown in Fig.~\ref{Fig: Truck}, the truck is equipped with 7 heterogeneous LiDARs, including 1 Ouster OS1-32, 3 Robosense Helios 32, and 3 Robosense M1 solid-state LiDAR.
During autonomous driving tests, CTE-MLO\cite{cte-mlo} is adopted to provide real-time state feedback to the control level of the truck by tightly coupling multi-LiDAR measurements.
However, for high-level planning tasks, such as port vehicle scheduling, drift-free localization and mapping is typically required.
To achieve high-precision long-term localization and mapping, we introduce UniLGL to perform loop closure detection and relative pose estimation, which is used to correct the long-term drift of multi-LiDAR odometry.
A factor graph is then employed to fuse the drift-prone odometry with the loop closure constraints provided by UniLGL, enabling globally consistent and accurate localization over extended periods.
The mathematical expression of factor graph optimization is shown as follows:
\begin{equation}
\small{
\begin{aligned}
{\mathbb{\hat T}} = \mathop {\arg \min }\limits_{\mathbb{T}}  &\sum\limits_{\left( {i,j} \right) \in {\mathbb{O}}} {\left\| \text{Log}\left({{\bf{T}}_i^{ - 1}{{\bf{T}}_j}\Delta {{{\bf{\hat T}}}_o}} \right)\right\|_2}  \\&+ \sum\limits_{\left( {i,j} \right) \in {\mathbb{L}}} {\left\| \text{Log}\left({{\bf{T}}_i^{ - 1}{{\bf{T}}_j}\Delta {{{\bf{\hat T}}}_l}} \right)\right\|_2} 
\end{aligned}\label{Eq: PGO}
}
\end{equation}
where $\mathbb{T}$ denotes the trajectory of the truck, and $\mathbf{T}_i, \mathbf{T}_j \in \mathbb{T}$ are the $i$-th and $j$-th poses along the trajectory, with $i > j$, $\mathbb{O}$ and $\mathbb{L}$ are the sets of index pairs corresponding to odometry constraints and loop closure constraints, 
and $\Delta\mathbf{\hat T}_o$ and $\Delta\mathbf{\hat T}_l$ represent the relative pose measurements between $\mathbf{T}_i$ and $\mathbf{T}_j$ obtained from CTE-MLO and UniLGL, respectively.
As shown in Fig.~\ref{Fig: PSATraj}, UniLGL provides a high success rate of global localization results in the highly repetitive ports environment, supplying high-quality loop closure constraints for the factor graph optimization problem defined in (\ref{Eq: PGO}). This effectively eliminates the drift of multi-LiDAR odometry and ensures reliable long-term localization and mapping performance in real-world autonomous truck driving scenarios.
The high-quality point cloud shown in Fig.~\ref{Fig: PSAMap} illustrates that the proposed method can provide
robust trajectory estimation to reconstruct a dense 3D, high-precision map in a large-scale port scenario when integrated with LiDAR-only odometry.
\subsection{Application 2: Multi-MAV Collaborative Exploration}
\begin{figure}[!t]\centering
\includegraphics[width=0.85\linewidth]{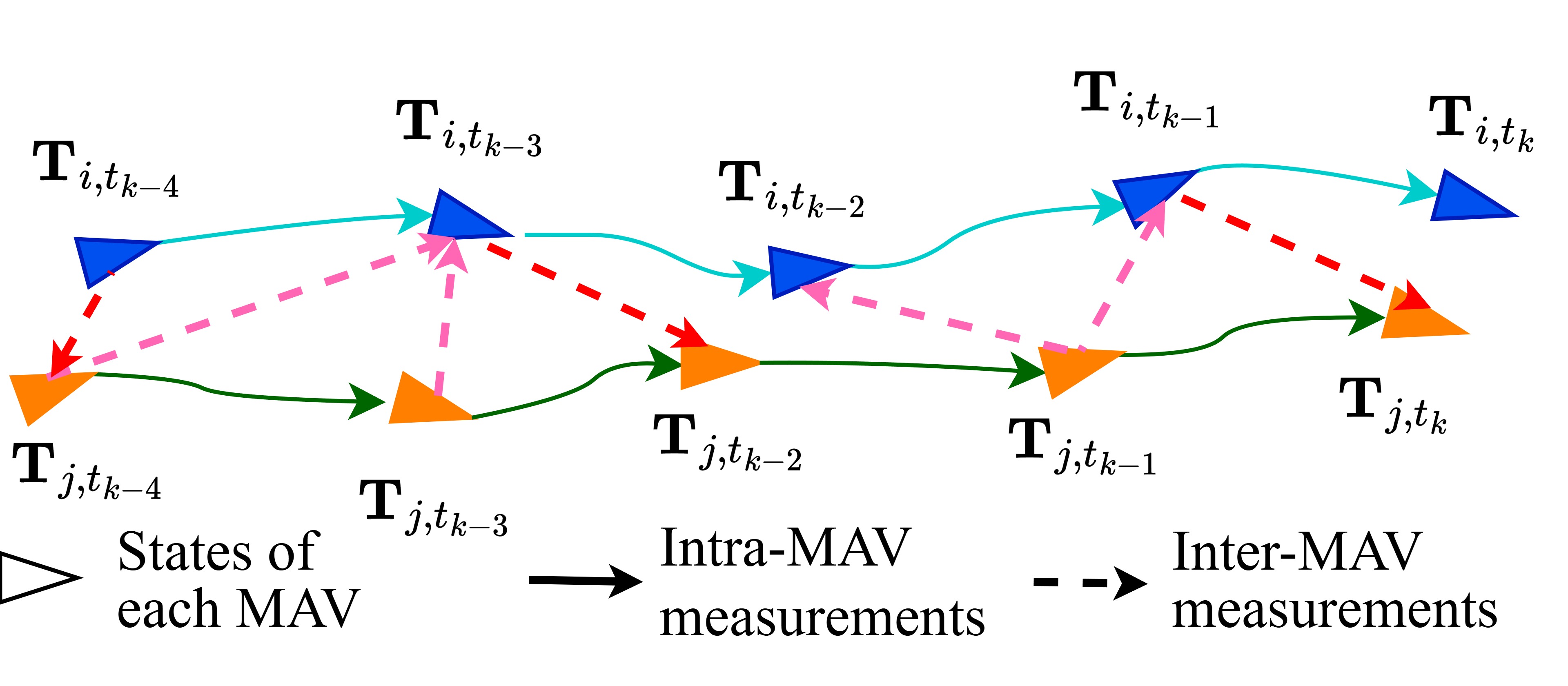}
\vspace{-0.5em}
\caption{Factor graph of the decentralized collaborative state estimator in two MAVs scenario (labeled with $i,j$).
}\label{Fig: FactorGraph}
\vspace{-1em}
\end{figure}
\begin{figure}[!t]\centering
\includegraphics[width=\linewidth]{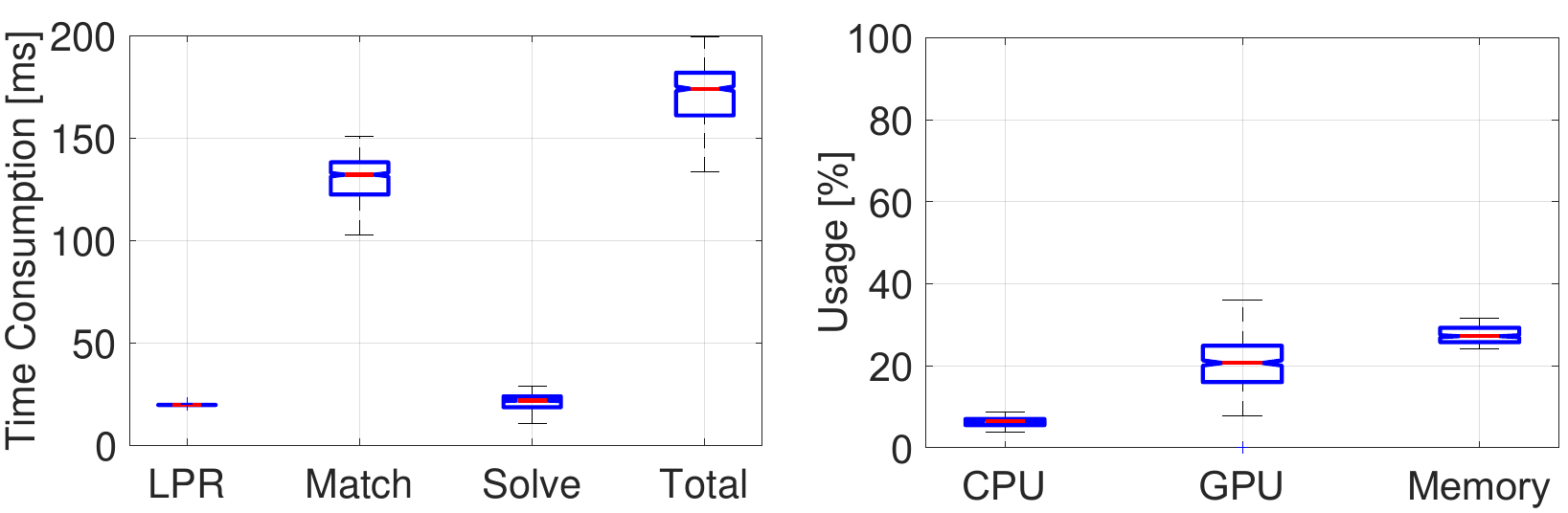}
\vspace{-1em}
\caption{Time consumption and deployment cost of UniLGL during the multi-MAV collaborative exploration experiment. The NVIDIA Orin NX adopts a unified memory architecture; the reported memory usage reflects the total memory shared by both the CPU and GPU.
\vspace{-1em}
}\label{Fig: TimeUsageUAV}
\end{figure}
\begin{figure*}[!t]\centering
\includegraphics[width=\linewidth]{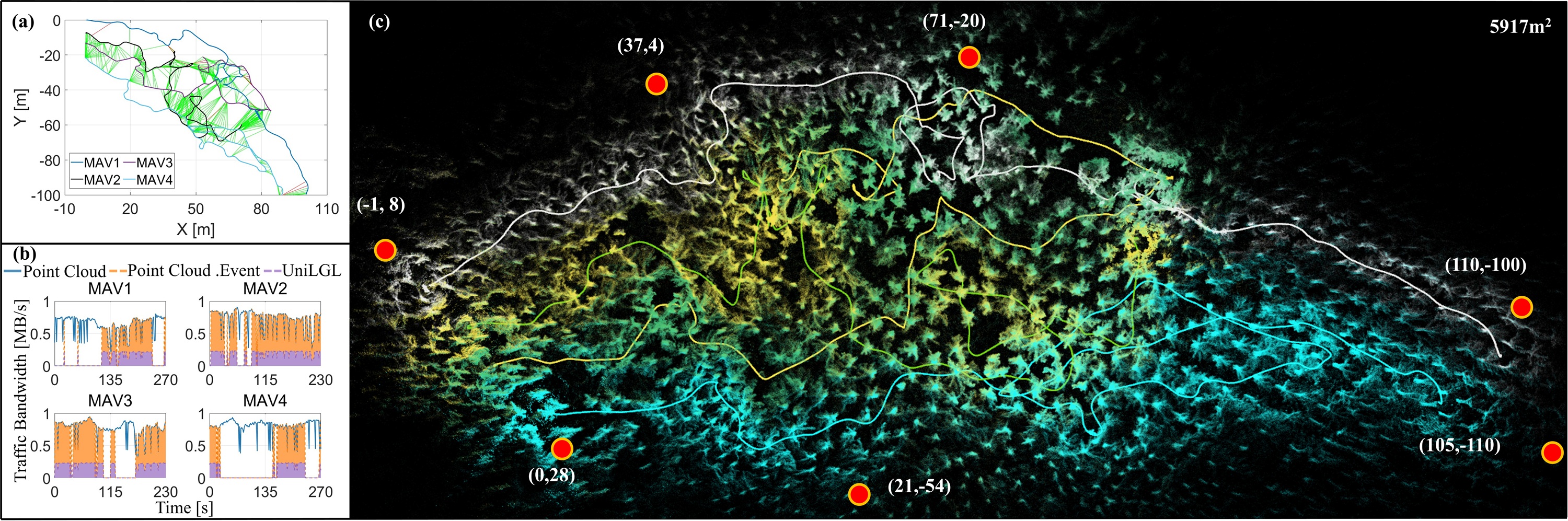}
\vspace{-0.5em}
\caption{Multi-MAV collaborative exploration. (a) Relative localization constraints among multiple MAVs provided by UniLGL. (b) The communication bandwidth requirement. (c) Point cloud map reconstructed through multi-MAV collaborative exploration. The red dots denote the exploration boundary.
}\label{Fig: UAVExploration}
\vspace{-1em}
\end{figure*}
To further validate its practicality on lightweight power-limited platforms, UniLGL is deployed on a multi-MAV system constructed by 4 identical MAVs to enable collaborative localization during an exploration task in a field scenario.
As shown in Fig. \ref{Fig: MAV}, each MAV platform is equipped with a Livox Mid 360 LiDAR, a Pixhawk4 flight controller, a low-power onboard computer (Nvidia Orin NX), and a VEWOE VMA10A mesh networking unit.
During collaborative exploration, we integrate the observations from UniLGL and CTE-MLO into a factor graph to enable decentralized collaborative state estimation for multiple MAVs.
As illustrated in Fig.~\ref{Fig: FactorGraph}, the relative pose estimations provided by UniLGL are incorporated as inter-MAV measurements, while the high frequency odometry information from CTE-MLO is utilized as intra-MAV measurements within the factor graph.
In Fig.~\ref{Fig: TimeUsageUAV}, we evaluate the time consumption and computational resource usage of deploying UniLGL on the NVIDIA Orin NX. 
UniLGL exhibits reasonable CPU ($6.47\%$ on average), GPU ($20.61\%$ on average), and memory ($27.50\%$ on average) usage, making it well-suited for low-power edge-computing platforms to provide inter-MAV constraints for multi-MAV systems.
The average processing time of UniLGL is $170.56 ms$, which enables more than $5$ frames per second relative state estimation.
As shown in Fig.~\ref{Fig: UAVExploration}{\color{blue}(a)}, UniLGL provides relative localization constraints among multiple MAVs when they observe similar point clouds (i.e., LPR succeeds).
By fusing the inter-MAV constraints provided by UniLGL with the high-frequency odometry information supplied by CTE-MLO through a factor-graph, real-time collaborative localization for multi-MAV systems can be achieved.

Notably, adopting UniLGL for inter-MAV constraints significantly reduces the communication bandwidth required in multi-agent systems.
This efficiency is reflected in two main aspects: 1) UniLGL enables collaborative localization through an event-triggered mechanism.
Specifically, a MAV transmits relative localization information to others only after a successful global descriptor matching, rather than continuously broadcasting point clouds as required by methods such as \cite{ebadi2020lamp,chang2022lamp,shen2022voxel}.
2) For relative pose estimation, UniLGL requires transmitting only a small number of point cloud keypoints associated with matched local features, thereby eliminating the need to transmit raw point clouds for pose refinement.
As shown in Fig.~\ref{Fig: UAVExploration}{\color{blue}(b)}, the UniLGL-based collaborative localization approach achieves substantially lower communication bandwidth consumption compared to methods that lack event-triggered mechanisms (noted as \textit{Point Cloud} in Fig.~\ref{Fig: UAVExploration}{\color{blue}(b)}) or require additional point cloud registration (noted as \textit{Point Cloud. Event} in Fig.~\ref{Fig: UAVExploration}{\color{blue}(b)}).
Owing to the superior efficiency of UniLGL in terms of computational and communication resource consumption, we integrate UniLGL-based collaborative localization with a decentralized exploration algorithm DPPM\cite{DPPM} and the MAV trajectory tracking control algorithm FxTDO-MPC\cite{10778610}, which enables a real-world collaborative exploration fully onboard.
As shown in Fig.~\ref{Fig: UAVExploration}{\color{blue}(c)}, the multi-MAV system completed high-precision scanning in a dense forest environment with an area of $5917 m^2$.

\section{Discussion and Limitations}
\begin{figure*}
\centering	
\subfigure[LPR fails with FoV-limited LiDAR due to the lack of information, while it succeeds with panoramic LiDAR.]{
    \includegraphics[width=0.95\linewidth]{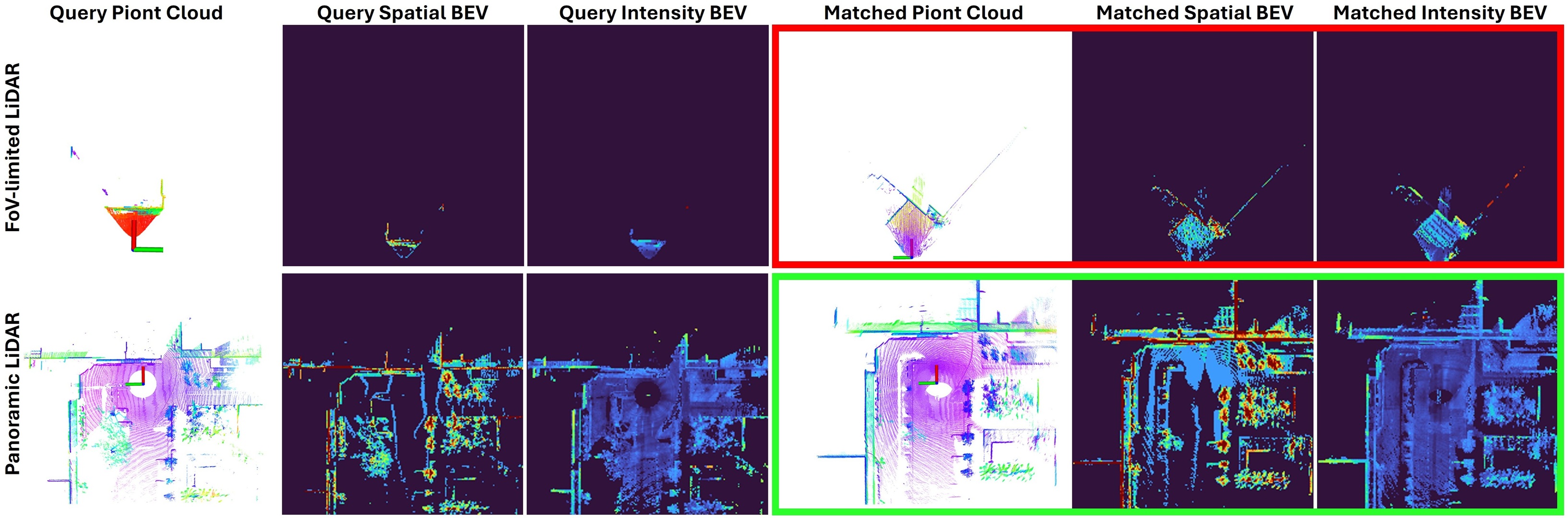}
    \label{Fig: FailureCase-noinfo}		
}
\subfigure[LPR fails in a degeneration scenario.]{
    \includegraphics[width=0.95\linewidth]{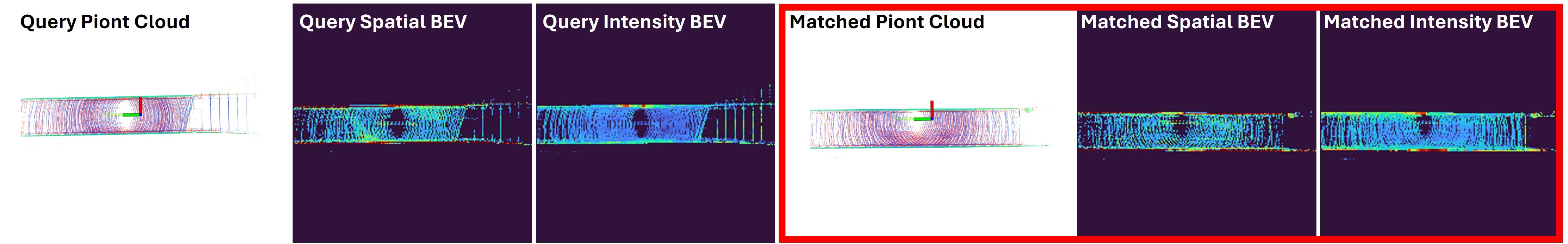} 
    \label{Fig: FailureCase-PRDegeneration}
}
\subfigure[LGL degeneration.]{
    \includegraphics[width=0.3705\linewidth]{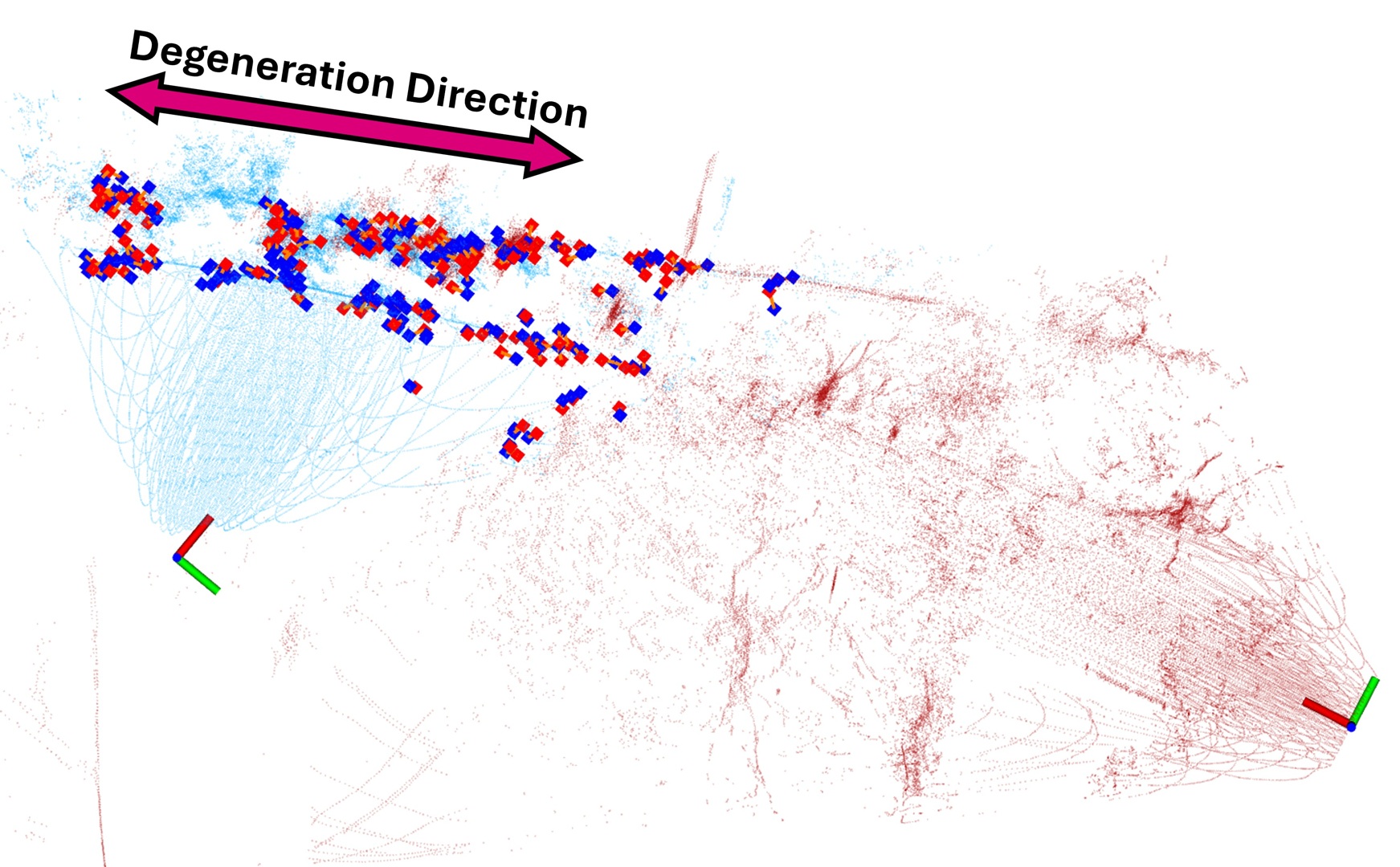} 
    \label{Fig: FailureCase-GLDegeneration}
}
\subfigure[LGL failure in the presence of a substantial number of dynamic objects.]{
    \includegraphics[width=0.5415\linewidth]{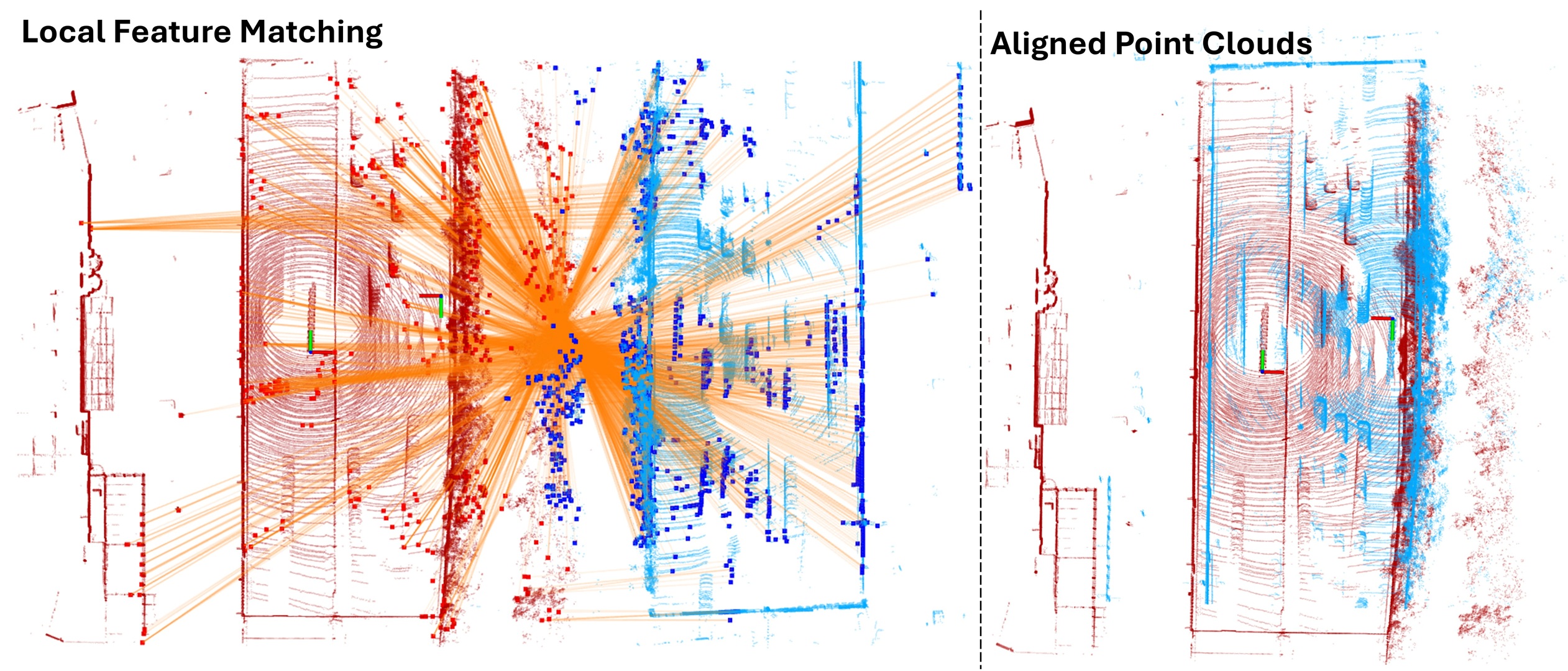} 
    \label{Fig: FailureCase-Dyna}
}
\caption{Representative failure cases. (a) and (b) illustrate corner cases that lead to LPR failure, while (c) and (d) show examples where LPR succeeds but LGL fails. In (c) and (d), the matched local features of the query and database point clouds are highlighted in blue and red, respectively, and the orange lines denote the local feature correspondences.} \label{Fig: failure cases}
\end{figure*}
Although UniLGL is highly robust against most challenges in LGL, there are a few corner cases shown in Fig.~\ref{Fig: failure cases} that it struggles to tackle effectively.
As shown in Fig.~\ref{Fig: FailureCase-noinfo}, UniLGL provides incorrect place retrieval when using FoV-limited LiDAR, as most of the LiDAR observations are occluded by nearby obstacles within the FoV,
leading to a slight performance drop compared with panoramic LiDAR (a $1.87\%$ decrease in recall and a $1.84\%$ decrease in average precision, as reported in Table~\ref{tab: recall-MCD}).
In Fig.~\ref{Fig: FailureCase-PRDegeneration}, a degeneration case is illustrated, where the query and database images retrieved by UniLGL depict nearly identical scenes, yet are geographically far apart.
The challenge lies in the fact that improving the network’s representation capacity or enhancing the discriminative ability of the global descriptor cannot effectively address issues caused by extreme information loss or scene degeneration.
Given that the LPR network introduced in Section~\ref{Sec:End-to-end BEV images fusion LPR network} supports image fusion, UniLGL can be easily extended to incorporate additional sensors.
A feasible solution is to introduce observations that are complementary to LiDAR observation (e.g., panorama image\cite{lu2025ringSharp}) into the LPR network.
For global pose estimation, as shown in Fig.~\ref{Fig: FailureCase-GLDegeneration}, UniLGL exhibits an accuracy degradation along the direction of degeneration, due to the lack of sufficient geometric constraints within the limited overlap of two FoV-limited LiDAR scans.
Although UniLGL fails to produce a correct global pose estimate in this case, it nevertheless serves as a representative example of its ability to perform LPR on low-overlap point clouds.
A similar accuracy drop can also be observed in Fig.~\ref{Fig: FailureCase-Dyna}, which is mainly caused by a substantial number of matching outliers introduced by dynamic objects, such as moving vehicles.
A straightforward solution is to incorporate data augmentation with and without dynamic objects during training, enabling the local features to learn to suppress dynamic foregrounds and focus on static landmarks that are beneficial for reliable global pose estimation.
\section{Conclusion}
In this article, a Uniform LiDAR-based Global Localization system, UniLGL, is developed to achieve cascaded place recognition and global pose estimation with the consideration of spatial and material uniformity as well as sensor-type uniformity.
To equip UniLGL with spatial and material uniformity, we represent the 4D point cloud information in a lossless manner using an image pair consisting of a spatial BEV image and an intensity BEV image, and design an end-to-end BEV fusion network for place recognition.
For sensor-type uniformity, a viewpoint invariance hypothesis is introduced to replace the conventional translation equivariance assumption commonly used in existing LPR networks\cite{BEVPlace++,PointNetVLAD,lcdnet,lu2025ringSharp,LoGG3D-Net,jung2025imlpr}, which hypothesis guides UniLGL to learn global descriptors and local features with consistency across geographically distant but co-visible areas.
Moreover, a VFM, DINO\cite{DINO}, is elegantly integrated into the proposed BEV fusion network to enhance the generalization capability, without requiring large amounts of LiDAR training data.
Thanks to the consistency of local features across co-visible areas, a global pose estimator is derived using GNC optimization\cite{8957085} to estimate the 6-DoF global pose based on point-level correspondences established through local feature matching between BEV images.
The extensive experimental results have demonstrated that the proposed UniLGL remains robust under extremely challenging conditions, such as long-term global localization and large viewpoint variations, across heterogeneous LiDAR configurations.
Furthermore, the proposed UniLGL has been deployed to support autonomy on diverse platforms, from full-size trucks to lightweight MAV, which enables high-precision truck localization and mapping in a port environment and multi-MAV collaborative exploration in a forest environment.
These real-world deployments affirm the extendability and applicability of UniLGL in industrial and field scenarios.
\appendices 
\section{Can Foundation Model Bring Generalization Ability?}
Foundation models in robotics aim to provide systems with broad generalization capabilities by leveraging large-scale pretraining on diverse data sources.
To facilitate understanding of the role of introducing foundation models into LGL, we evaluate the LGL performance of three variants: UniLGL, \textit{UniLGL w/o FM}, and UniLGL initialized with a foundation model but without fine-tuning (\textit{UniLGL w/o FT}).
To comprehensively evaluate the cross-model and cross-scene generalization capability of UniLGL, two benchmark datasets, MCD~\cite{nguyen2024mcd} and Garden~\cite{Mag-MM}, are utilized. Both datasets provide paired FoV-limited and panoramic LiDAR measurements, serving to assess the cross-model generalization performance.
Moreover, their data are acquired from entirely distinct environments, thereby facilitating the evaluation of the cross-scene generalization capability.
\subsection{Cross-model Generalization Ability}
\begin{figure}[!t]\centering
\includegraphics[width=\linewidth]{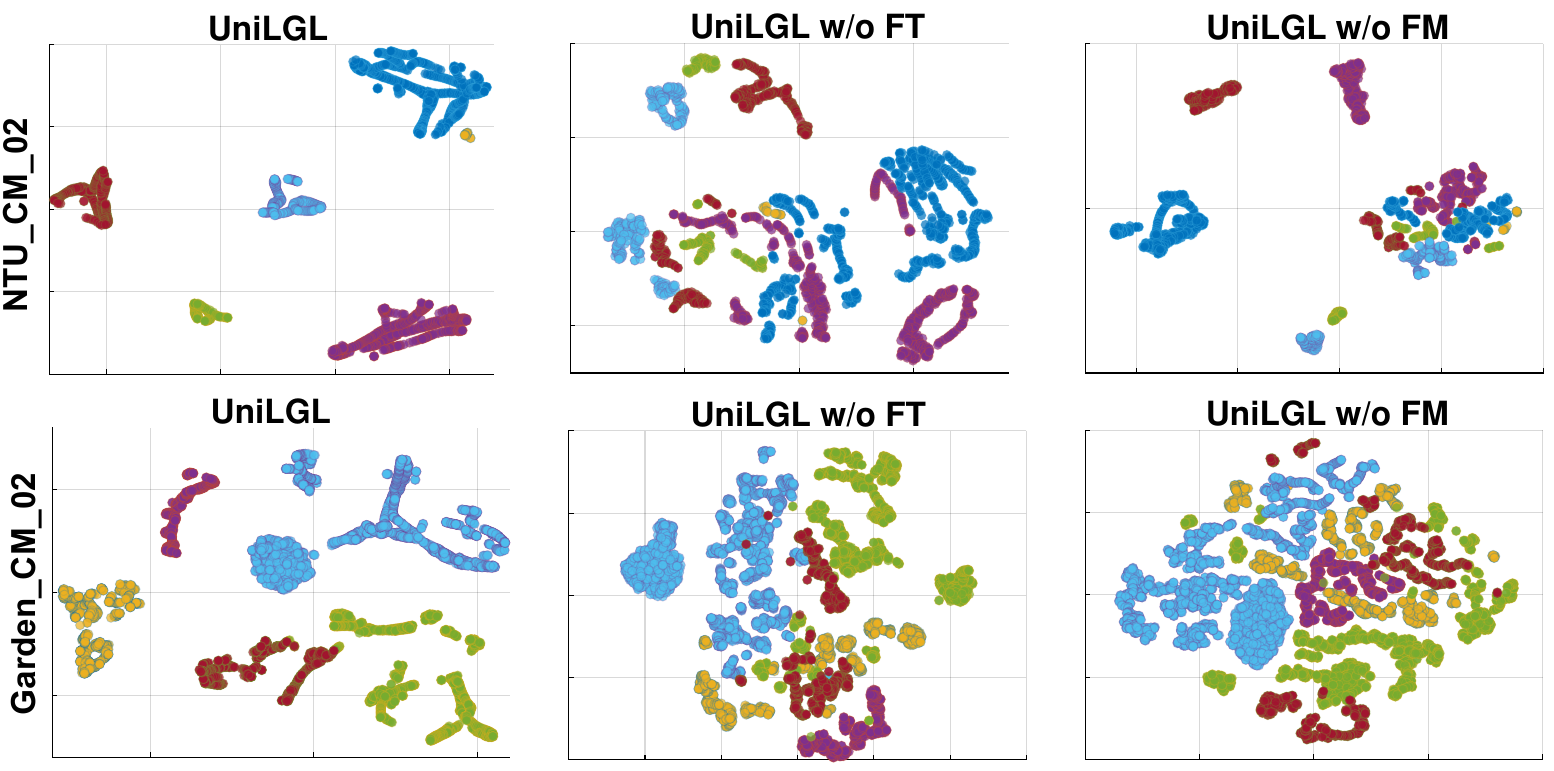}
\vspace{-1em}
\caption{t-SNE visualization of global descriptor encoded by UniLGL, UniLGL w/o FT, and UniLGL w/o FM. For each sequence, we select six distinct locations to visualize the discriminability of the global descriptors.}\label{Fig: tsne-appendix}
\vspace{-1em}
\end{figure}

During the experiments, the FoV-limited LiDAR scans are used as queries, while the panoramic LiDAR scans serve as the database.
The associated data sequences are referred to as NTU\_CM\_XX and Garden\_CM\_XX, respectively.
It is worth noting that, to demonstrate the \textit{zero-shot} generalization ability, no cross-modal training is performed.
As shown in the t-Distributed Stochastic Neighbor Embedding (t-SNE)\cite{tsne} visualization in Fig.~\ref{Fig: tsne-appendix}, introducing a foundation model imparts a certain level of generalization ability to the LPR network, enabling \textit{UniLGL w/o FT} to achieve better clustering performance than \textit{UniLGL w/o FM}, without fine-tuning.
By initializing the network with DINO\cite{DINO} and fine-tuning with only a small amount of homogeneous LiDAR data, UniLGL learns highly discriminative global descriptors in the heterogeneous LPR task.
In addition to the above qualitative analysis, we present quantitative results of place recognition and global localization performance in Table~\ref{tab: zero-shot recall}.
The results show that, thanks to the introduction of the foundation model, UniLGL achieves outstanding \textit{zero-shot cross-modal generalization ability}.
When compared to \textit{UniLGL w/o FT} and \textit{UniLGL w/o FM}, UniLGL achieves a $59.64\%$-$82.24\%$ improvement in recall and over $79\%$ increase in LGL successful rate.
\begin{table}[!t] \centering
\centering
\caption{Zero-Shot Cross-model LPR (Recall ($\%$) at Top-1)/LGL (Success Rate ($\%$)) Performance.}
\setlength{\tabcolsep}{3.3pt} 
\label{tab: zero-shot recall}
\begin{threeparttable}
    \begin{tabular}{l|| c c c}
    \hline\hline
    {Sequence} &{UniLGL} &\makecell{UniLGL w/o FT} &\makecell{UniLGL w/o FM}	\\\hline
     NTU\_CM\_02 &\bf97.75/86.75 &37.95/0.07 &0.60/0.01 \\
     NTU\_CM\_10 &\bf92.95/78.19 &32.61/0.13 &3.64/0.01\\
     NTU\_CM\_13 &\bf96.46/86.13 &41.02/0.35 &8.59/0.00\\
     Garden\_CM\_01 &\bf93.19/75.77 &26.02/0.25 &3.33/0.00\\
     Garden\_CM\_02 &\bf91.86/76.10 &30.54/0.34 &17.54/0.01\\
     Garden\_CM\_03 &\bf94.37/76.07 &31.32/0.41 &19.71/0.00 \\
     Garden\_CM\_04 &\bf87.37/76.94 &37.06/0.36 &24.86/0.01\\\hline
     Average     &\bf93.42/79.42 &33.78/0.27 &11.18/0.01\\\hline\hline
    \end{tabular}
\begin{tablenotes}
    \item The best result is highlighted in {\color{black}\bf{Bold}}.
\end{tablenotes}
\end{threeparttable}
\vspace{-2em}
\end{table}
\subsection{Cross-scene Generalization Capability}
To evaluate the cross-scene generalization capability, UniLGL and \textit{UniLGL w/o FM} are retrained solely on the MCD dataset for 5 epochs, using \textit{ntu\_day\_01} and \textit{ntu\_night\_08} as the training sequences.
Table~\ref{tab: cross-scene} presents the performance of UniLGL, \textit{UniLGL w/o FT}, and \textit{UniLGL w/o FM} across two distinct environments — the large-scale campus scenes in the MCD dataset and the repetitive artificial vegetational scenes in the Garden dataset.
From the results, \textit{UniLGL w/o FT} consistently outperforms \textit{UniLGL w/o FM}, and achieves a comparable performance with UniLGL on the Garden dataset.
This indicates that the VFM enables the network to obtain preliminary place recognition ability even without LGL-specific fine-tuning.
However, on the MCD dataset, \textit{UniLGL w/o FT} exhibits a marked performance degradation. This is attributed to the presence of numerous low-overlap point cloud pairs induced by large viewpoint differences, which represents an LGL-specific condition that a task-agnostic pre-trained VFM is hard to handle adequately without fine-tuning.
UniLGL fine-tunes the DINO\cite{DINO} using a small amount of LiDAR data under LGL-specific supervision, enabling it to achieve consistent performance across distinct environments and thereby demonstrating \textit{cross-scene generalization capability}.
As the quantitative results shown in Table~\ref{tab: cross-scene}, UniLGL achieves a $7.39$-$17.05\%$ improvement in recall and a $13.57$-$28.29\%$ increase in LGL successful rate.
\begin{table}[!t] \centering
\centering
\caption{Cross-scene LPR (Recall ($\%$) at Top-1)/LGL (Success Rate ($\%$)) Performance.}
\setlength{\tabcolsep}{3.3pt} 
\label{tab: cross-scene}
\begin{threeparttable}
    \begin{tabular}{l|| c c c}
    \hline\hline
    {Sequence} &{UniLGL} &\makecell{UniLGL w/o FT} &\makecell{UniLGL w/o FM}	\\\hline
     Mid\_NTU\_02 &\bf98.50/75.30 &81.48/28.81 &70.97/23.99 \\
     Mid\_NTU\_10 &\bf95.44/88.98 &92.11/75.88 &87.80/70.17\\
     Mid\_NTU\_13 &\bf96.85/71.05 &70.10/39.58 &66.95/33.54\\
     Garden\_01 &\bf91.19/85.53 &90.58/84.62 &78.51/67.83\\
     Garden\_02 &\bf91.21/84.76 &90.08/83.24 &76.88/62.70\\
     Garden\_03 &\bf86.75/78.37 &86.23/77.24 &73.90/52.22 \\
     Garden\_04 &\bf90.80/83.96 &88.39/83.47 &76.36/59.51 \\\hline
     Average    &\bf92.96/81.14 &85.57/67.57 &75.91/52.85\\\hline\hline
    \end{tabular}
\begin{tablenotes}
    \item The best result is highlighted in {\color{black}\bf{Bold}}.
\end{tablenotes}
\end{threeparttable}
\vspace{-2em}
\end{table}

\bibliographystyle{IEEEtran}
\bibliography{IEEEabrv,main}
\end{document}